\documentclass[10pt,twocolumn,letterpaper]{article}

\usepackage[pagenumbers]{cvpr} 
\usepackage[accsupp]{axessibility} 
\usepackage{pifont}
\usepackage{amsmath}
\usepackage{booktabs}
\usepackage{multirow}
\usepackage{colortbl}
\usepackage{tabularx}
\usepackage{tcolorbox}
\usepackage{listings}
\usepackage{newtxtext,newtxmath}

\definecolor{cvprblue}{rgb}{0.21,0.49,0.74}
\usepackage[pagebackref,breaklinks,colorlinks,allcolors=cvprblue]{hyperref}

\def\confName{CVPR}
\def\confYear{2026}

\title{FineCog-Nav: Integrating Fine-grained Cognitive Modules for Zero-shot Multimodal UAV Navigation}

\author{Dian Shao$^{1}$\thanks{Corresponding authors.}\ ,\ Zhengzheng Xu$^{1}$,\ Peiyang Wang$^{2}$,\ Like Liu$^{1}$,\ Yule Wang$^{1}$,\ Jieqi Shi$^{2*}$,\ Jing Huo$^{2}$\\
$^1$Northwestern Polytechnical University~~~~~
$^2$Nanjing University\\
{\tt\small shaodian@nwpu.edu.cn, isjieqi@nju.edu.cn}
}

\begin{document}
\maketitle
\addtocontents{toc}{\protect\setcounter{tocdepth}{0}}

\begin{abstract}
UAV vision-language navigation (VLN) requires an agent to navigate complex 3D environments from an egocentric perspective while following ambiguous multi-step instructions over long horizons.
Existing zero-shot methods remain limited, as they often rely on large base models, generic prompts, and loosely coordinated modules.
In this work, we propose \textbf{FineCog-Nav}, a top-down framework inspired by human cognition that organizes navigation into fine-grained modules for language processing, perception, attention, memory, imagination, reasoning, and decision-making.
Each module is driven by a moderate-sized foundation model with role-specific prompts and structured input-output protocols, enabling effective collaboration and improved interpretability.
To support fine-grained evaluation, we construct AerialVLN-Fine, a curated benchmark of 300 trajectories derived from AerialVLN, with sentence-level instruction-trajectory alignment and refined instructions containing explicit visual endpoints and landmark references.
Experiments show that FineCog-Nav consistently outperforms zero-shot baselines in instruction adherence, long-horizon planning, and generalization to unseen environments.
These results suggest the effectiveness of fine-grained cognitive modularization for zero-shot aerial navigation.
Project page: \url{https://smartdianlab.github.io/projects-FineCogNav}.
\end{abstract}
\section{Introduction}
\label{sec:intro}
Vision-Language Navigation for UAVs poses fundamental challenges in embodied AI. Unlike ground-based VLN, UAV navigation requires following ambiguous multi-step instructions and planning over long horizons in complex 3D environments from an egocentric view~\cite{AerialVLN}. These settings make robust navigation difficult due to continuous 3D motion, limited global awareness, and weak landmark distinctiveness~\cite{openfly,openuav}.

Although current zero-shot VLN frameworks based on large foundation models (LFMs) achieve strong results in ground navigation, they still face substantial challenges in UAV scenarios.
These methods often depend heavily on large LLMs or VLMs~\cite{Zhou2023NavGPTER, zhang2025citynavagent}. Their performance can degrade substantially with smaller models; for example, the success rate reported in~\cite{zhang2025citynavagent} drops from 28.3 to 1.7 when GPT-4V is replaced with LLaVA-7B.
Moreover, most existing methods rely on generic prompts and weak coordination across modules, limiting the interaction between perception, reasoning, and decision-making. 
As a result, they often lack components that are important for complex UAV navigation, such as hierarchical planning~\cite{STMR}, dynamic subgoal extraction~\cite{chen20232A2Nav}, and memory mechanisms~\cite{discussnav}.
Fundamentally, they overlook the integrated, collaborative nature of cognition required for true navigation intelligence.

\begin{figure}[t!]
    \centering
    \includegraphics[width=\linewidth]{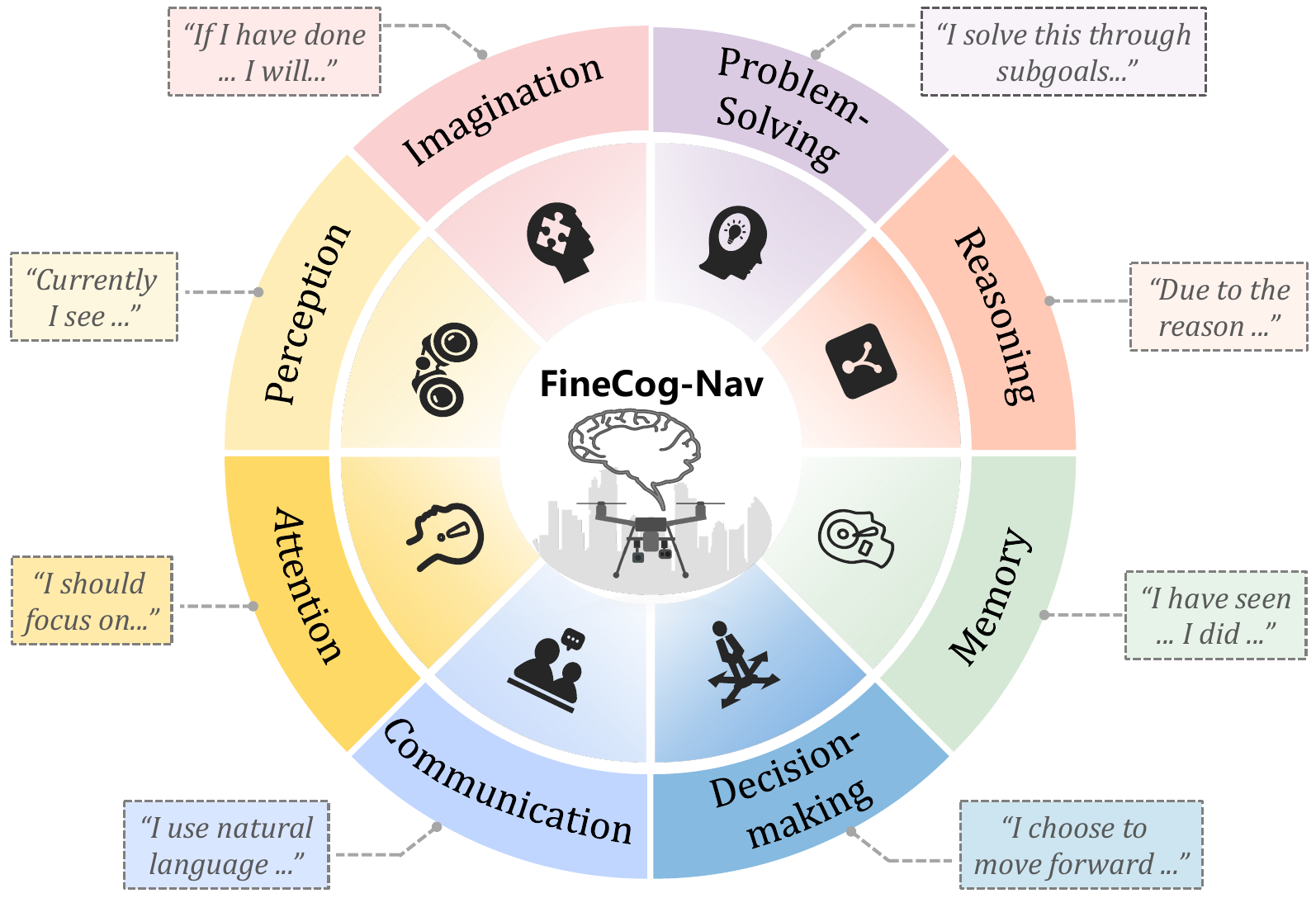}
    \caption{We propose FineCog-Nav, a framework designed to simulate diverse human cognitive functions.} 
    \label{fig:teaser}
\end{figure}

In this work, we introduce \textbf{FineCog-Nav}, a zero-shot framework that integrates \textbf{Fine}-grained, human-\textbf{Cog}nition–inspired modules for robust UAV \textbf{Nav}igation. 
Unlike prior multi-agent architectures, FineCog-Nav is organized around cognitive functions rather than agent identities.
As shown in \cref{fig:teaser}, FineCog-Nav explicitly models the core, interdependent cognitive functions (\ie language processing, perception, attention, memory, imagination, reasoning, and decision-making) through the following dedicated modules:
an \underline{\textit{Instruction Parser}} that decomposes complex instructions into manageable sentences, with a \underline{\textit{Subgoal Extractor}} further identifying subgoals;
a \underline{\textit{Perception Module}} that describes the current scene in first-person view, guided by the \underline{\textit{Attention Module}} that highlights salient landmarks and critical visual cues;
an \underline{\textit{Imagination Module}} that simulates potential environmental states after subgoal completion;
a \underline{\textit{Memory Module}} that maintains hierarchical memory across step, subgoal, and instruction levels;
a \underline{\textit{Subgoal Judger}} that leverages imagination, memory, and other information to evaluate subgoal completion;
and a \underline{\textit{Decision-Making Module}} that selects the optimal navigational action at each step and explains its reasoning.
Each module is powered by a moderate-sized foundation model with carefully designed prompts that specify cognitive role, general task, UAV-related guidance, structured I/O protocols, etc.
Through effective inter-module collaboration, FineCog-Nav enables a more interpretable and human-like navigation process, without relying on large-scale models or extensive training data. To the best of our knowledge, FineCog-Nav is the first modular UAV VLN framework explicitly modeling cognitive interdependence at this granularity, moving beyond monolithic or loosely coupled components.

To facilitate fine-grained evaluation, we also construct \textbf{AerialVLN-Fine}, a curated benchmark of 300 high-quality trajectories derived from AerialVLN~\cite{AerialVLN}. Each trajectory is paired with sentence-level annotations that align instructions with trajectory segments. Ambiguous instructions are further refined with explicit visual endpoints and landmark references. This benchmark supports fine-grained evaluation beyond traditional trajectory-level metrics. In summary, our main contributions are as follows:
\begin{itemize}
    \item We propose \textbf{FineCog-Nav}, a cognition-inspired zero-shot framework for UAV VLN. It enables interpretable, human-like navigation via collaborative, cognitively structured modules and precisely specified prompts, surpassing mere individual module enhancement.
    \item We construct \textbf{AerialVLN-Fine}, a curated benchmark with sentence-level alignment and refined instructions for fine-grained UAV navigation evaluation.
    \item We conduct extensive experiments and analyses showing that FineCog-Nav consistently outperforms strong baselines across diverse model sizes and evaluation settings.
\end{itemize}
\section{Related Work}
\textbf{Vision-and-Language Navigation for UAVs.}
Ground-based VLN has evolved from attention-based seq-to-seq models \cite{anderson2018vision,Fried2018SpeakerFollowerMF} to pretraining- and map-based methods \cite{Hao2020TowardsLA,Chen2022ThinkGA,Wang2023GridMMGM}, and more recently to zero-shot LLM/VLM agents \cite{Zhou2023NavGPTER,long2024instructnavzeroshotgenericinstruction,Dorbala2022CLIPNavUC}. Recent work has also moved toward continuous-control settings \cite{Wu2024VoroNavVZ,TopV-Nav,esc,L3MVN,Zhang2024NaVidVV} for more realistic embodied navigation.
Compared to ground-based VLN, UAV VLN faces unique challenges. 
Early work (e.g., LANI \cite{Misra2018MappingIT}, ANDH \cite{Fan2022AerialVN}) relied on top-down maps, while AerialVLN \cite{AerialVLN} introduced realistic FPV and continuous flight settings. 
Later methods enhanced grounding and efficiency through map integration \cite{STMR, navagent, SkyVLN, zeng2025janusvln}, hierarchical planning \cite{Zhang2025GroundedVN, zhang2025citynavagent, Wei2025StreamVLNSV}, and VLM/LLM-driven decision-making \cite{openfly, Cai2025FlightGPTTG, zhou2024navgpt2unleashingnavigationalreasoning}.
However, zero-shot UAV VLN remains underexplored. Recent methods, including STMR \cite{STMR}, NavAgent \cite{navagent}, VLM-RRT \cite{Ye2025VLMRRTVL}, and See, Point, Fly (SPF) \cite{hu2025see}, improve grounding or leverage pretrained models for aerial decision making, but largely focus on specific components rather than a unified framework. Prior work has explored individual capabilities such as memory \cite{GeoNav} and instruction parsing \cite{navagent,Cai2025FlightGPTTG}, as well as hybrid settings including dual-UAV collaboration \cite{Wu2025AeroDuoAD} and cross-entity generalization \cite{navfom,zhang2024uni,gao2025octonav}. Still, UAV VLN lacks a modular and interpretable zero-shot framework for long-horizon instruction following, motivating FineCog-Nav.

\vspace{5pt}
\noindent \textbf{VLN Datasets.}
VLN research began with indoor tasks~\cite{anderson2018vision,chen2021history} and later extended to continuous control in Habitat \cite{savva2019habitat}. Subsequent datasets~\cite{Ku2020RoomAcrossRoomMV,jain2019stay,chen2021history} increased complexity and multilingual support, but remained focused on short-range ground navigation. 
In contrast, UAV VLN requires long-range planning in outdoor scenes. 
Early aerial datasets like ANDH \cite{Fan2022AerialVN} emphasized interactive QA but lacked realism in control. AerialVLN \cite{AerialVLN} addressed this by utilizing first-person views, physics realism, and long-horizon paths, establishing a strong benchmark. Other works \cite{Lee2024CityNavLA,openuav,openfly,aeroverse,wang2025uav,xiao2025uav} further increased scale, multimodal integration, and domain realism. Specialized datasets such as LogisticsVLN \cite{Zhang2025LogisticsVLNVN} reflect emerging application needs like last-mile delivery.
Nonetheless, many large-scale datasets still suffer from noisy, auto-generated instructions. To ensure fine-grained evaluation, some recent works build cleaner, task-specific datasets: VLFly \cite{Zhang2025GroundedVN} for indoor UAVs, UAV-VLN \cite{Saxena2025UAVVLNEV} for simplified outdoor setups. However, such datasets often trade generality for control, demanding high-quality benchmarks with clear instructions in UAV VLN research.
\begin{figure*}[t]
    \centering
    \includegraphics[width=1.0\linewidth]{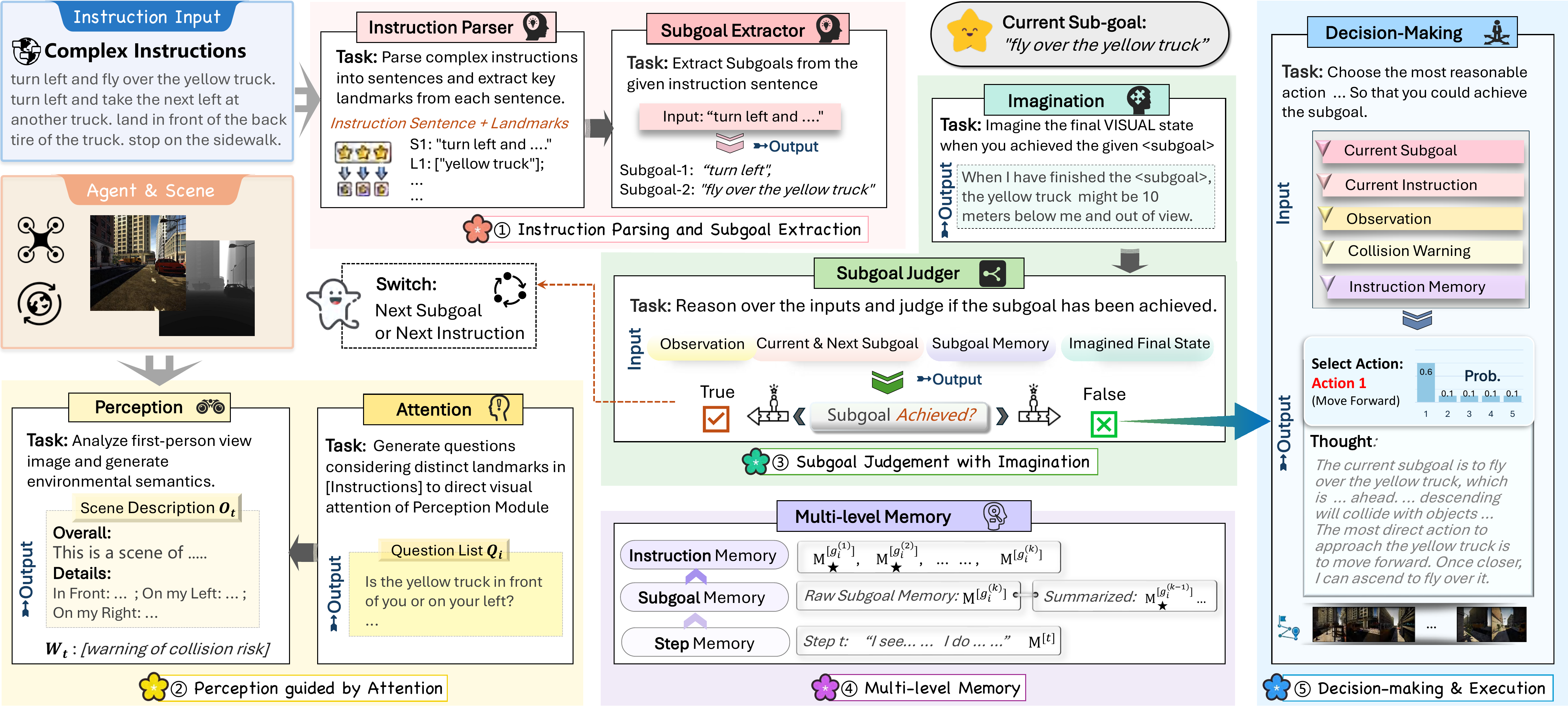}
    \caption{\textbf{Overview of FineCog-Nav}, a zero-shot UAV VLN framework using cognitively inspired LLM/VLM-based modules to explicitly model cognitive interdependence. Given a complex natural language instruction, it involves the following steps: \ding{182} Instruction Parsing and Subgoal Extraction; \ding{183} Perception guided by Attention; \ding{184} Subgoal Judgment with Imagination; \ding{185} Multi-level Memory Management; and \ding{186} Decision-Making and Action Execution. In the figure, we mainly illustrate the interactions among modules within each step; for the overall workflow and interactions between steps, please refer to the detailed description in the Method section. Best viewed in color.}
    \label{fig:pipeline}
\end{figure*}

\section{Methodology}
\subsection{Task Formulation}
In UAV VLN, an agent (drone) receives a natural language instruction $I$ and navigates in a partially observable 3D environment $\mathcal{E}$ from a first-person view. The goal is to navigate to a destination such that the Euclidean distance between its final position $p_{stop}$ and the target destination $p_{dest}$ is below a threshold $\delta$:
$ \|  p_{stop} - p_{dest} \|  \leq \delta $.
The set of executable actions is: 
$\mathcal{A}=\{$\textit{TaskFinish}, 
$\textit{MoveForward}(d)$, 
$\textit{TurnLeft}(\theta)$, 
$\textit{TurnRight}(\theta)$, 
$\textit{Ascend}(h)$, 
$\textit{Descend}(h)$, 
$\textit{MoveLeft}(d)$, 
and $\textit{MoveRight}(d) \}$,  where $\theta = 15^{\circ}$, $h=2$ units and $d=5$ units~\cite{AerialVLN}.
The task terminates when $\textit{TaskFinish}$ is selected.

\subsection{The FineCog-Nav Framework}
FineCog-Nav is a zero-shot VLN framework tailored for UAVs in complex 3D environments. It comprises a set of cognition-inspired modules, each implemented with moderate-scale foundation models, to systematically model key cognitive abilities such as language processing, perception, attention, memory, imagination, reasoning, and decision-making~\cite{sternberg2006cognitive}, all in a functional rather than biologically faithful sense.
Unlike monolithic architectures, it features explicit cognitive interdependence: each module's output informs the next, forming a closed processing loop.
This design improves interpretability through explicit intermediate states and enables generalizable navigation without extensive pretraining or task-specific fine-tuning. The detailed workflow and module functionalities are illustrated in \cref{fig:pipeline} and described below (full prompts see Suppl.).

\vspace{2pt}
\noindent \textbf{\ding{182} Instruction Parsing and Subgoal Extraction.}  
Given a complex natural language navigation instruction $I$, FineCog-Nav first employs the \textit{Instruction Parser} $\mathcal{S}$ to segment ${I}$ into sequential sentences, each paired with its associated landmarks $L_i$:
\begin{align}
     \mathcal{S} (I) = \{ ({I}_1, L_1), \, ({I}_2, L_2) , ... \, ,({I}_N, L_N) \}. 
\end{align}
The \textit{Subgoal Extractor} $\mathcal{E}$ further dynamically generates an ordered list of executable subgoals when fetching a new $I_i$, conditioned on both the parsed instruction and current environmental observations:
\begin{align}
     \{ g_i^{(k)} \}_{k=1}^{K} = \mathcal{E} (I_i, O_t),
\end{align}
where $\{ g_i^{(k)} \}$ prioritizes execution order over syntactic structure.
Such a \textit{hierarchical problem-solving} strategy not only reduces the complexity of planning but also promotes adaptive and interpretable task progression (see \cref{fig:qualitative} for illustration). This step initiates the cognitive flow by passing structured goals and landmarks to subsequent modules.

\vspace{2pt}
\noindent \textbf{\ding{183} Perception guided by Attention.}  
The \textit{Perception Module} $\mathcal{P}$ generates semantic scene descriptions $O_t$ based on first-person observations $V_t^{(RGB)}$. To ensure focus on task-relevant information and mitigate distractions from irrelevant details, the \textit{Attention Module} identifies key landmarks $\{L_i, L_{i+1}\}$ from both the current and upcoming instructions, and formulates targeted queries $\{Q_i\}$ regarding their spatial status and attributes:
\begin{align}
   \{Q_i\} &= \text{Attention}(\{L_i,\, L_{i+1}\}), \\
   O_t &= \mathcal{P}(V_t^{(RGB)},\, \{Q_i\}).
\end{align}
Separately, a lightweight rule-based safety module estimates collision risk from the depth observation $V_t^{(D)}$ using geometric heuristics and outputs a warning signal $W_t$ for candidate actions. These semantic observations and safety cues are then used by the subgoal judger, memory module, and decision module.

\vspace{3pt}
\noindent \textbf{\ding{184} Subgoal Judgement with Imagination.} 
The \textit{Subgoal Judger} $\mathcal{J}$ determines whether the current subgoal $g_i^{(k)}$ has been achieved by integrating multi-source cues: current visual observation $O_t$, subgoal memory $\mathrm{M}^{[g_i^{(k)}]}$, and an imagined reference state $\mathrm{R}^{[g_i^{(k)}]}$ generated by the \textit{Imagination Module}. Here, the imagined reference is a brief, landmark-centric description of the expected scene at subgoal completion, conditioned on relevant landmarks and neighboring subgoals, rather than an open-ended scene generation. This constrained design helps mitigate hallucination by using imagination as an auxiliary cue. The judger infers subgoal completion as follows:
\begin{align}
\mathcal{J}\big(O_t,\, g_i^{(k)},\, \mathrm{M}^{[g_i^{(k)}]},\, \mathrm{R}^{[g_i^{(k)}]}\big) \to \{\text{True},\, \text{False}\}.
\end{align}
Upon completion of the current subgoal, the system transitions to the next one. Once all subgoals in $I_i$ are accomplished, the \textit{Subgoal Extractor} is invoked to retrieve the next instruction $I_{i+1}$. Otherwise, the system continues toward fulfilling the current objective. 
By regulating the overall reasoning flow and dynamically affecting memory organization and planning, the judgment results facilitate a coherent and adaptive cognitive workflow.

\vspace{4pt}
\noindent \textbf{\ding{185} Multi-level Memory Management.} Rather than a specific stage, the hierarchical \textit{Memory Module} serves as a foundational component, maintaining context throughout navigation at three levels:
\noindent\ding{227} \textit{Step Memory} $\mathrm{M}^{[t]}$ stores observation and action at each time step $t$, capturing “I see..., I do...”;
\noindent\ding{227} \textit{Subgoal Memory} $\mathrm{M}^{[g_i^{(k)}]}$ accumulates the step memories associated with subgoal $g_i^{(k)}$. When the subgoal is completed, these records are summarized into a compact subgoal memory $\mathrm{M}^{[g_i^{(k)}]}_\star$ by an LLM, inspired by human memory consolidation;
\noindent\ding{227} \textit{Instruction Memory} $\mathrm{M}^{[I_i]}$ aggregates the completed subgoal summaries for instruction $I_i$:
\begin{align}
    \mathrm{M}^{[I_i]} = \{\; \mathrm{M}^{[g_i^{(1)}]}_\star,\, \mathrm{M}^{[g_i^{(2)}]}_\star,\, \ldots, \; \mathrm{M}^{[g_i^{(k)}]}_\star\}.
\end{align}
This multi-level memory preserves long-horizon context while filtering local noise, which helps stabilize decision-making over extended trajectories. Ablation Results in \cref{tab:ablation} show replacing this hierarchy with flat history (as in NavGPT~\cite{Zhou2023NavGPTER}) substantially degrades performance, highlighting the effectiveness of our memory design.

\vspace{2pt}
\noindent \textbf{\ding{186} Decision-Making and Action Execution.} 
The \textit{Decision-Making Module} $\mathcal{D}$ selects the optimal action $a_t$ from the valid action set $\mathcal{A}$. This selection is achieved by jointly integrating the current subgoal $g_i^{(k)}$, environmental observation $O_t$, collision warning $W_t$, current instruction $I_i$, and \textit{hierarchical} instruction memory $\mathrm{M}^{[I_i]}$. 
Specifically, these inputs are structured into a prompt for the LLM, which returns relative action-preference scores $p$ over the candidate actions and selects the top-ranked action $a_t$. Meanwhile, it generates an interpretable rationale $\tau$ that explicitly considers observations, memory, and subgoal objectives:
\begin{align}
    \{p, a_t,\, \tau \} = \mathcal{D}\big(g_i^{(k)},\, I_i \, ,O_t,\, W_t,\, \mathrm{M}^{[I_i]}\big).
\end{align}
This process enables transparent, human-like decision-making and is repeated until all instructions are completed or the navigation task ends.  Selected actions influence subsequent perception and are stored in memory for future decision cycles, thus maintaining a continuous cognitive loop.

\textbf{Overall,} FineCog-Nav frames UAV navigation as a sequence of fine-grained, cognition-inspired modules, each designed to address key challenges in real-world VLN. Through hierarchical task decomposition, attentive perception, imagination-driven reasoning, multi-level memory, and explainable decision-making, our framework achieves interpretable and adaptive navigation, even in complex, partially observable environments. The continuous exchange of structured information among modules not only preserves temporal and contextual coherence, but also closes the perception–reasoning–action loop, supporting robust and transparent behavior.
\begin{figure*}[t]
    \centering
    \includegraphics[width=1\linewidth]{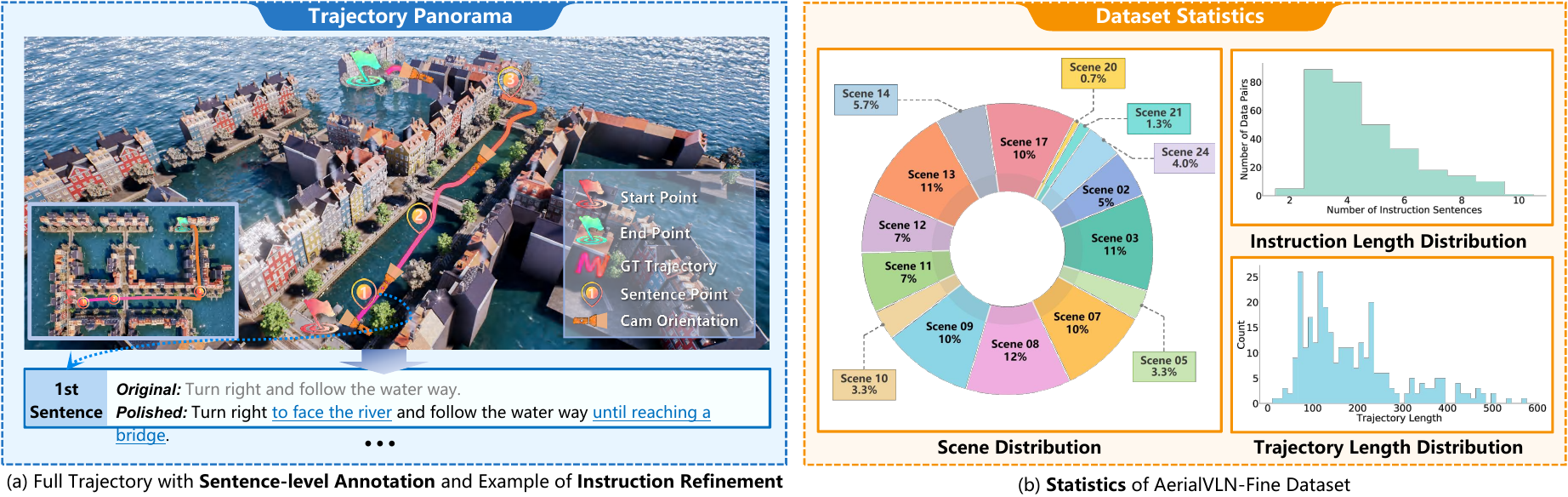}
    \caption{Overview of the AerialVLN-Fine dataset. \textit{Left:} (a) Example of fine-grained annotation in AerialVLN-Fine, showing sentence-level alignment between instructions and trajectory segments, as well as refinement of instruction sentences. \textit{Right:} (b) Visualizations of scene, instruction, and trajectory length distributions, highlighting the dataset’s diversity and complexity.}
    \label{fig:dataset}
\end{figure*}

\section{The AerialVLN-Fine Dataset}
\vspace{1pt}
\noindent \textbf{Data Source and Motivation.}
While UAV VLN has recently seen several large-scale datasets~\cite{openfly,openuav,aeroverse}, most lack thorough peer validation. Among them, AerialVLN~\cite{AerialVLN} is the most widely recognized public benchmark. However, our analysis reveals notable data quality issues, including trajectory-instruction misalignment, ambiguous instructions, collision anomalies, and indistinct/invisible landmarks (see Suppl.). These issues limit the dataset’s ability to diagnose model performance, especially in zero-shot settings that require cleaner data. 
To address this, we introduce \textbf{AerialVLN-Fine}, a carefully curated benchmark with fine-grained sentence-level alignment and refined instructions across 15 AerialVLN scenes, enabling precise diagnosis of model capabilities and supporting robust, effective evaluation for future research.

\vspace{3pt}
\noindent \textbf{The Annotation Process.}
To ensure the quality and diversity of AerialVLN-Fine, we adopted a systematic pipeline combining automated checks with rigorous manual verification (details in Suppl.), involving:
\ding{42} \textit{Manual Review and Selection:} Each instruction-trajectory pair was manually checked using four quality criteria, and  $\sim$20 diverse, high-quality samples were retained per scene.
\ding{42} \textit{Instruction Segmentation and Refinement:} Complex instructions were split into instruction sentences, and further refined for explicit actions, clear endpoints, and unambiguous/visible landmarks.
\ding{42} \textit{Trajectory Segmentation and Alignment:} Trajectories were segmented and precisely aligned with each instruction sentence (see \cref{fig:dataset}).
\ding{42} \textit{Rigorous Quality Control:} All annotations underwent multiple rounds of manual review for accuracy and consistency.

\vspace{3pt}
\noindent \textbf{Dataset Statistics.}\label{sec:dataset}
AerialVLN-Fine comprises 300 high-quality instruction-trajectory pairs, sourced from AerialVLN-S-Val splits. 
Each trajectory averages 189 meters and 76 actions, with 4.6 fine-grained semantic sentences. Instruction refinement increases the average instruction length from 59 to 103 words, yielding richer descriptions. The dataset contains 1,383 sentences, each spanning 41 meters and 17 actions on average. 
We also classify trajectories into five capability dimensions for capability-level evaluation (details see Suppl.).
Dataset statistics and annotation examples are illustrated in \cref{fig:dataset},  highlighting its suitability for evaluation and sentence-level analysis. 
Moreover, its moderate scale enables comprehensive evaluation and analysis without high computational cost. The same annotation pipeline can also be extended to other UAV VLN datasets or larger stratified subsets in future work.

\section{Experiments}
\begin{table*}[ht]
    \caption{\textbf{Performance comparison with single-module BaseModel} across various LLMs on AerialVLN-Fine, all using the same VLM (Qwen2.5-VL-32B). FineCog-Nav consistently outperforms BaseModel on all metrics.}
    \label{tab:base_result}
    \centering
    \small
    \begin{tabular}{l|l|ccccccc} \hline 
    \textbf{Model} & \textbf{Method} 
    & SR$^{2D}$/\%$\uparrow$ & SR$^{3D}$/\%$\uparrow$ & OSR/\%$\uparrow$ & NE/m$\downarrow$
    & nDTW/\%$\uparrow$ & PL/m & Steps \\ 
    \hline
    \rowcolor{gray!10}
    \multicolumn{9}{l}{\textit{Closed-source LLM}}\\
    \multirow{2}{*}{Gemini 2.5 Flash-Lite}     
        & BaseModel  & 1.33 & 1.00 & 3.67 & 240.04 & 8.84 & 329.78 & 129.70 \\
        & \cellcolor{cyan!10}\textbf{FineCog-Nav} & \cellcolor{cyan!10}\textbf{4.00} & \cellcolor{cyan!10}\textbf{2.67} & \cellcolor{cyan!10}\textbf{6.33} & \cellcolor{cyan!10}\textbf{120.70} & \cellcolor{cyan!10}\textbf{15.66} & \cellcolor{cyan!10}146.24 & \cellcolor{cyan!10}57.52 \\ \cline{2-9} 
    \multirow{2}{*}{GPT-4o-mini}  
        & BaseModel  & 0.33 & 0.33 & 2.00 & 325.98 & 8.74 & 350.43 & 103.10 \\
        & \cellcolor{cyan!10}\textbf{FineCog-Nav} & \cellcolor{cyan!10}\textbf{4.00} & \cellcolor{cyan!10}\textbf{2.33} & \cellcolor{cyan!10}\textbf{3.67} & \cellcolor{cyan!10}\textbf{100.37} & \cellcolor{cyan!10}\textbf{20.45} & \cellcolor{cyan!10}56.98 & \cellcolor{cyan!10}31.73 \\  \hline
    \rowcolor{gray!10}
    \multicolumn{9}{l}{\textit{LLM Model Size: 8B - 15B}}\\
    \multirow{2}{*}{InternLM3-8B} 
        & BaseModel  & 0.67 & 0.33 & 3.33 & 128.13 & 14.01 & 86.30 & 34.65 \\
        & \cellcolor{cyan!10}\textbf{FineCog-Nav} & \cellcolor{cyan!10}\textbf{2.67} & \cellcolor{cyan!10}\textbf{2.33} & \cellcolor{cyan!10}\textbf{6.67} & \cellcolor{cyan!10}\textbf{120.72} & \cellcolor{cyan!10}\textbf{14.91} & \cellcolor{cyan!10}139.72 & \cellcolor{cyan!10}42.22 \\ \cline{2-9}
    \multirow{2}{*}{ChatGLM-4-9B}     
        & BaseModel  & 1.00 & 0.33 & 1.33 & 124.03 & 13.27 & 50.19 & 20.36 \\
        & \cellcolor{cyan!10}\textbf{FineCog-Nav} & \cellcolor{cyan!10}\textbf{3.00} & \cellcolor{cyan!10}\textbf{2.67} & \cellcolor{cyan!10}\textbf{2.67} & \cellcolor{cyan!10}\textbf{97.05} & \cellcolor{cyan!10}\textbf{19.26} & \cellcolor{cyan!10}30.49 & \cellcolor{cyan!10}19.39 \\ \hline
    \rowcolor{gray!10}
    \multicolumn{9}{l}{\textit{LLM Model Size: 20B - 32B}}\\
    \multirow{2}{*}{InternLM2.5-20B} 
        & BaseModel  & 0.33 & 0.33 & 1.67 & 152.34 & 9.94 & 141.27 & 74.16 \\
        & \cellcolor{cyan!10}\textbf{FineCog-Nav} & \cellcolor{cyan!10}\textbf{2.00} & \cellcolor{cyan!10}\textbf{2.00} & \cellcolor{cyan!10}\textbf{4.00} & \cellcolor{cyan!10}\textbf{103.94} & \cellcolor{cyan!10}\textbf{19.35} & \cellcolor{cyan!10}61.05 & \cellcolor{cyan!10}28.49 \\ \cline{2-9}
    \multirow{2}{*}{ChatGLM-4-32B}   
        & BaseModel  & 2.33 & 2.00 & 5.00 & 180.66 & 10.59 & 235.03 & 104.20 \\
        & \cellcolor{cyan!10}\textbf{FineCog-Nav} & \cellcolor{cyan!10}\textbf{3.33} & \cellcolor{cyan!10}\textbf{2.33} & \cellcolor{cyan!10}\textbf{5.33} & \cellcolor{cyan!10}\textbf{94.18} & \cellcolor{cyan!10}\textbf{21.25} & \cellcolor{cyan!10}45.91 & \cellcolor{cyan!10}20.18 \\ \cline{2-9}
    \multirow{2}{*}{Qwen3-32B}       
        & BaseModel  & 2.67 & 3.00 & 6.33 & 142.72 & 17.07 & 114.56 & 39.12 \\
        & \cellcolor{cyan!10}\textbf{FineCog-Nav} & \cellcolor{cyan!10}\textbf{5.00} & \cellcolor{cyan!10}\textbf{4.00} & \cellcolor{cyan!10}\textbf{7.00} & \cellcolor{cyan!10}\textbf{95.31} & \cellcolor{cyan!10}\textbf{20.31} & \cellcolor{cyan!10}60.36 & \cellcolor{cyan!10}26.05 \\ \hline
    \rowcolor{gray!10}
    \multicolumn{9}{l}{\textit{LLM Model Size: 70B - 72B}}\\
    \multirow{2}{*}{Llama3.3-70B}  
        & BaseModel  & 3.00 & 2.67 & 5.67 & 263.10 & 9.67 & 329.36 & 111.48 \\
        & \cellcolor{cyan!10}\textbf{FineCog-Nav} & \cellcolor{cyan!10}\textbf{6.67} & \cellcolor{cyan!10}\textbf{6.00} & \cellcolor{cyan!10}\textbf{9.67} & \cellcolor{cyan!10}\textbf{98.84} & \cellcolor{cyan!10}\textbf{20.17} & \cellcolor{cyan!10}84.96 & \cellcolor{cyan!10}38.91 \\ \cline{2-9}
    \multirow{2}{*}{Qwen2.5-72B}    
        & BaseModel  & 2.67 & 2.67 & 6.33 & 270.10 & 10.69 & 298.24 & 116.91 \\
        & \cellcolor{cyan!10}\textbf{FineCog-Nav} & \cellcolor{cyan!10}\textbf{8.00} & \cellcolor{cyan!10}\textbf{6.00} & \cellcolor{cyan!10}\textbf{9.00} & \cellcolor{cyan!10}\textbf{91.43} & \cellcolor{cyan!10}\textbf{22.48} & \cellcolor{cyan!10}67.96 & \cellcolor{cyan!10}33.90 \\ \hline
    \end{tabular}
    \vspace{2pt}
\end{table*}
\begin{table*}[ht] 
    \caption{\textbf{Comparison with framework baselines} on AerialVLN-Fine (300 samples) and AerialVLN-S-Val (864 samples). All methods employ a modular design and identical base model combinations. FineCog-Nav consistently achieves substantially better results across all metrics and datasets, demonstrating the effectiveness and robustness of its fine-grained modules and collaborative framework.}
    \label{tab:comparison_combined}
    \centering
    \small
    \setlength{\tabcolsep}{1mm}
    \begin{tabular}{c | c | l ccccccc}
    \toprule
    \textbf{Model Combination} & \textbf{Dataset} & \textbf{Method} & SR$^{2D}$/\%$\uparrow$ & SR$^{3D}$/\%$\uparrow$ & OSR/\%$\uparrow$ & nDTW/\%$\uparrow$ & NE/m$\downarrow$ & PL/m & Steps \\
    \midrule
    \multirow[c]{6}{3cm}{\centering Qwen2.5-72B \\ + Qwen2.5-VL-32B} 
    & \multirow[c]{3}{*}{AerialVLN-Fine} 
        & NavGPT & 0.33 & 0.00 & 0.67 & 15.90 & 110.94 & 21.44 & 9.33 \\
    & 
        & DiscussNav & 2.67 & 2.67 & 3.33 & 19.63 & 98.36 & 39.51 & 17.76 \\
    & 
        & \cellcolor{cyan!10} \textbf{FineCog-Nav} & \cellcolor{cyan!10}\textbf{8.00} & \cellcolor{cyan!10}\textbf{6.00} & \cellcolor{cyan!10}\textbf{9.00} & \cellcolor{cyan!10}\textbf{22.48} & \cellcolor{cyan!10}\textbf{91.43} & \cellcolor{cyan!10}67.96 & \cellcolor{cyan!10}33.90 \\ \cline{2-10}
    
    & \multirow[c]{3}{*}{AerialVLN-S} 
        & NavGPT & 0.12 & 0.12 & 0.46 & 8.29 & 135.65 & 44.92 & 13.02 \\
    & 
        & DiscussNav & 0.46 & 0.46 & 0.93 & 8.33 & 158.46 & 81.06 & 22.46 \\
    & 
        & \cellcolor{cyan!10} \textbf{FineCog-Nav} & \cellcolor{cyan!10}\textbf{1.97} & \cellcolor{cyan!10}\textbf{1.50} & \cellcolor{cyan!10}\textbf{1.85} & \cellcolor{cyan!10}\textbf{11.47} & \cellcolor{cyan!10}\textbf{130.32} & \cellcolor{cyan!10}75.28 & \cellcolor{cyan!10}34.52 \\ 
    \bottomrule
    \end{tabular}
\end{table*}
\vspace{-1pt}
\subsection{Experimental Setup}
\noindent \textbf{Datasets.} Primary evaluations are conducted on AerialVLN-Fine, which comprises 300 high-quality instruction-trajectory pairs with sentence-level annotation from diverse scenes, making it particularly suitable for zero-shot performance assessment.
We also report results on the original AerialVLN-S-Val dataset~\cite{AerialVLN}, which consists of 864 instruction-trajectory pairs. 
Experiments on this larger and noisier dataset serve to demonstrate the robustness and effectiveness of our method under a broader range of conditions, and further strengthen the credibility of our findings in more challenging, real-world scenarios.

\noindent \textbf{Evaluation Metrics.} We use standard VLN evaluation metrics, adapted to the 3D UAV setting in AerialVLN. While the official metrics are based on 2D Euclidean distances, we extend them to 3D space using $(x, y, z)$ coordinates for greater accuracy.  The core metrics include:
\ding{172} \textit{Navigation Error} (NE): Average 3D distance from the UAV’s final position to the goal (meters).
\ding{173} \textit{Success Rate} (SR$^{2D}$/SR$^{3D}$): Percentage of trajectories ending within 20m of the goal in 2D/3D space.
\ding{174}\textit{Oracle Success Rate} (OSR): Percentage of trajectories passing within 20m of the goal at any point.
\ding{175} \textit{Normalized Dynamic Time Warping} (nDTW): Trajectory fidelity to the reference path (higher is better)~\cite{nDTW}.
We also report average Path Length (PL) and steps to reflect the agent’s efficiency and exploration tendency. 

\noindent \textbf{LLM Selection.}
To systematically assess the adaptability of FineCog-Nav and baseline methods with different language models, we adopt a fixed Vision-Language Model (VLM) and vary the Large Language Models (LLMs), ensuring consistent perception.
Considering deployment on edge devices with limited GPU memory, we adopt the Qwen2.5-VL-32B as the VLM and restrict backbones to at most 72B parameters.
Specifically, for LLM selection, we evaluate a range of small- to moderate-sized models from diverse families, covering various parameter scales and architectures. Among \textit{closed-source} models, we include GPT-4o-mini and the recently introduced Gemini 2.5 Flash-Lite. For \textit{open-source} models, our evaluation spans three size tiers: 8B–15B, 20B–32B, and 70B–72B. 
All models are accessed through official APIs.

\begin{table*}[htbp]
    \caption{\textbf{Ablation Analysis of FineCog-Nav Modules on AerialVLN-Fine.}
    We ablate the Attention (Attn.) and Imagination (Imag.) modules, remove the subgoal mechanism, and replace hierarchical Memory (\textbf{H}) with plain history (\textbf{P}), respectively.}
    \small
    \label{tab:ablation}
    \centering
    \begin{tabular}{cccc | cccc | ccc}
    \toprule
    \multicolumn{4}{c}{\textbf{Module}} & \multicolumn{4}{c}{\textbf{Full-Traj}} & \multicolumn{3}{c}{\textbf{Sentence-level}} \\
    Attn. & Imag. & Subgoal & Mem. &
    SR/\%$\uparrow$ & OSR/\%$\uparrow$ & NE/m$\downarrow$ & nDTW/\%$\uparrow$ & 
    S-SR/\%$\uparrow$ & S-OSR/\%$\uparrow$ & S-NE/m$\downarrow$ \\
    \midrule
    \ding{56} & \ding{51} & \ding{51} & \textbf{H} & 3.00 & 7.00 & 104.38 & 19.55 
    & 19.92 & 26.46 & 66.21 \\
    \ding{51} & \ding{56} & \ding{51} & \textbf{H} & 3.67 & 5.00 & 101.27 & 19.67
    & 20.48 & 24.21 & 64.85 \\
    \ding{51} & \ding{51} & \ding{56} & \textbf{H} & 2.00 & 4.33 & 102.32 & 19.13 
    & 19.42 & 24.28 & 65.54 \\
    \ding{51} & \ding{51} & \ding{51} & \textbf{P} & 0.67 & 2.67 & 97.76 & 19.69 
    & 19.42 & 23.50 & 62.90 \\
    \rowcolor{yellow!10}
    \ding{56} & \ding{56} & \ding{51} & \textbf{H} & 3.67 & 8.67 & 106.59 & 18.14 
    & 19.35 & \textbf{27.30} & 70.16 \\
    \rowcolor{yellow!10}
    \ding{51} & \ding{56} & \ding{56} & \textbf{H} & 4.00 & 8.67 & 98.75 & 19.78 
    & 16.68 & 26.46 & 65.36 \\
    \rowcolor{cyan!10}
    \ding{51} & \ding{51} & \ding{51} & \textbf{H}  & \textbf{6.00} & \textbf{9.00} & \textbf{91.43} & \textbf{22.48} 
    & \textbf{22.03} & 27.23 & \textbf{59.02} \\
    \bottomrule
    \end{tabular}
\end{table*}
\begingroup
\setlength{\tabcolsep}{1mm}
\begin{table}[t]
    \caption{\textbf{Sentence-Level Analysis} of FineCog-Nav on AerialVLN-Fine with different base LLMs.}
    \label{tab:sentence_result}
    \centering
    {\small
    \begin{tabular}{l ccc}
    \toprule
    \textbf{Base LLM} & S-SR/\%$\uparrow$ & S-nDTW/\%$\uparrow$ & S-NE/m$\downarrow$ \\
    \midrule
    \rowcolor{gray!10} Gemini2.5-Flash-Lite & 16.40 & 26.33 & 79.00 \\
    \rowcolor{gray!10} GPT-4o-mini & 19.56 & 28.58 & 65.96 \\
    \midrule
    \rowcolor{cyan!5} InternLM3-8B & 15.62 & 28.19 & 72.01 \\
    \rowcolor{cyan!5} ChatGLM-4-9B  & 19.92 & 31.42 & \underline{62.18} \\
    \midrule 
    \rowcolor{cyan!5} InternLM2.5-20B  & 19.00 & \underline{34.11} & 67.67 \\
    \rowcolor{cyan!5} ChatGLM-4-32B   & 15.95 & 23.20 & 62.67 \\
    \rowcolor{cyan!5} Qwen3-32B & 18.76 & 30.04 & 62.42 \\
    \midrule
    \rowcolor{cyan!5} Llama3.3-70B & \underline{21.32} & 33.89 & 64.82 \\
    \rowcolor{cyan!5} Qwen2.5-72B  & \textbf{22.03} & \textbf{35.37} & \textbf{59.02} \\
    \bottomrule
    \end{tabular}
    \vspace{-2pt}
    }
\end{table}
\endgroup
\noindent \textbf{Baselines, Adaptations, and Parity Control.} 
To comprehensively evaluate FineCog-Nav, our main experiments involve two settings:
\ding{248} \textit{Single-module BaseModel}: 
Each base LLM is paired with prompts adapted from the action maker in \cite{STMR}, covering basic instruction understanding, memory, and planning (see Suppl. for full prompts). This enables comparison between our multi-cognitive-function framework (with carefully designed prompts) and conventional single-module approaches.
\ding{248} \textit{Framework baselines}: 
We compare FineCog-Nav with representative prompt-based VLN frameworks, NavGPT and DiscussNav, both of which utilize LLM coordination. In NavGPT, multiple modules are designed to realize basic navigation functions such as perception, history, and planning, while DiscussNav employs a multi-agent discussion mechanism for navigation tasks. 
This comparison tests whether FineCog-Nav benefits from its structured modular design beyond simply using multiple modules or agents.
To ensure fair comparison and isolate the effect of framework design, we enforce strict parity across all baselines:
\ding{249} \textit{Model Backbone:} All LLM components use the same model family and size.
\ding{249} \textit{Perceptual Input:} All methods receive identical egocentric visual inputs, use the same VLM, and are provided with uniform, rule-based collision cues.
\ding{249} \textit{Action Space:} All frameworks operate within the same continuous UAV command space, with consistent mapping for any discrete actions.

\subsection{Main Results}
\noindent  \ding{95} \textbf{Comparison with Single-module BaseModel.}
We evaluate several base LLMs with basic prompts, forming single-module baseline referred to as BaseModel, as described above. For a fair and meaningful comparison, all experiments are conducted on the AerialVLN-Fine dataset, which provides a cleaner and more discriminative benchmark. Results are summarized in \cref{tab:base_result}. We can see that:
\ding{172} FineCog-Nav consistently outperforms the BaseModel across all metrics. Specifically,
\textit{in terms of outcomes}, our SR and OSR are significantly higher, and the average distance to the goal (NE) is much shorter.
\ding{173} For FineCog-Nav, larger LLMs significantly improve navigation performance, demonstrating the framework's strong scalability. In contrast, simply increasing the size of LLMs in the BaseModel does not consistently yield better results.
\ding{174}  These results demonstrate that a well-designed modular framework with effective integration can further enhance performance, more fully activating the inherent capabilities of large models.

\vspace{2pt}
\noindent \ding{95} \textbf{Comparison with Framework Baselines.}
We first evaluate on the constructed AerialVLN-Fine dataset, followed by scalable experiments on the larger and noisier AerialVLN-S-Val subset. Results are shown in \cref{tab:comparison_combined}.
Our key findings are as follows:
\ding{172} Across both datasets, FineCog-Nav consistently outperforms all baselines across all metrics, validating the efficacy of our \textit{cognitive, collaborative framework}. Notably, NavGPT rarely explores the environment, while DiscussNav relies on multiple rounds of dialogue, resulting in lower efficiency.
\ding{173} Despite the increased challenges posed by AerialVLN-S-Val, our method maintains a clear advantage, demonstrating robustness on larger and noisier datasets.
\ding{174} Even with modular frameworks, zero-shot UAV VLN remains challenging when using moderate-sized models. However, this choice prioritizes future edge deployment,  as very large models like GPT-4, though performant, are impractical for such scenarios.
It is important to note that the modest success rates primarily owe to the intrinsic difficulty of UAV VLN, the zero-shot setting, and the limitations of current lightweight models. For reference, even fully trained SOTA VLN systems~\cite{navfom, openuav} report low SRs (e.g., 6.3\%/4.2\%) in complex, unseen scenarios. Nevertheless, these results do not diminish the importance of developing effective frameworks, irrespective of the specific SR attained.

\subsection{Ablations and Analysis}\label{sec:ablation}

\noindent \ding{96} \textbf{Ablation Studies on Modules.}
We conduct comprehensive ablation studies on the core cognitive modules of FineCog-Nav, as shown in \cref{tab:ablation}. We can see: 
\ding{172} \textit{single-module ablations} (rows 1–4) show that removing any module leads to performance drops, with the largest decline occurring when hierarchical memory is replaced by plain history, underscoring its critical role;
\ding{173} \textit{Joint ablations} (rows 5–6) outperform some single-module cases. This indicates that tightly coupled modules may collapse when only one is removed, whereas removing both avoids such cascading failures, consistent with our theory that effective navigation relies on coordinated interdependent processes.
\ding{174} The best results are achieved when all modules are enabled, highlighting the effectiveness of our complete framework.

\begin{figure*}[t!]
    \centering
    \includegraphics[width=\linewidth]{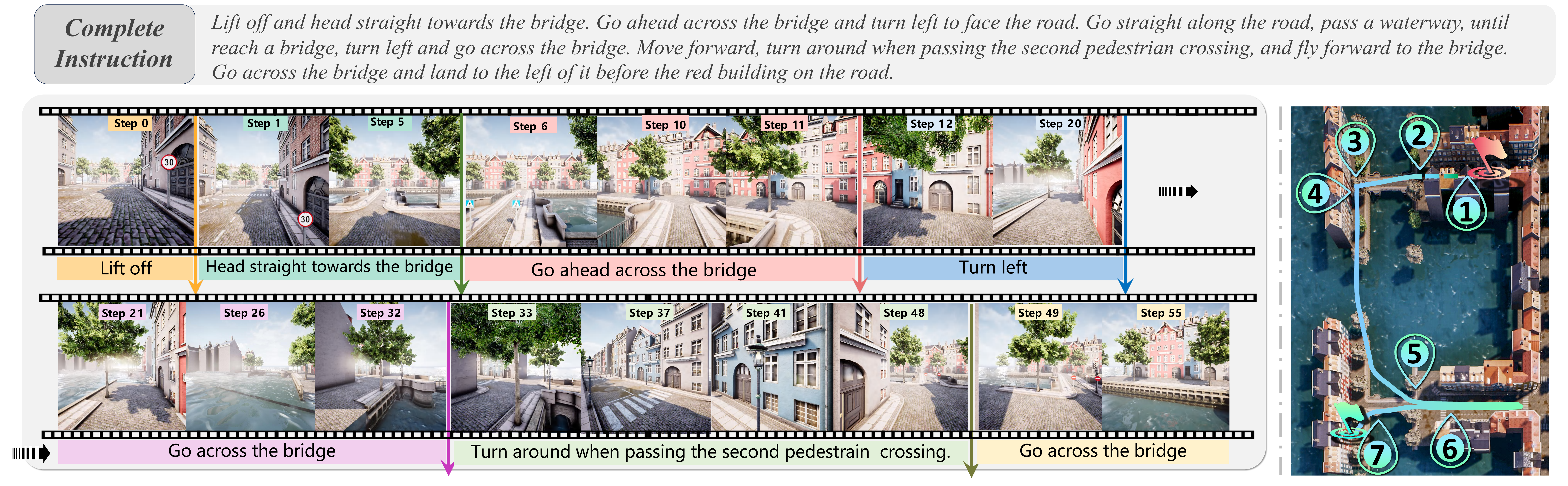}
    \caption{Qualitative example of FineCog-Nav. Left: Stepwise reasoning with sub-goals. Right: Bird’s-eye view of the trajectory, with each number marking a sub-goal’s endpoint.} 
    \label{fig:qualitative}
\end{figure*}
\vspace{2pt}
\noindent \ding{96} \textbf{Qualitative Visualization.} 
\cref{fig:qualitative} shows representative trajectories from FineCog-Nav on AerialVLN-Fine, demonstrating interpretable, stepwise reasoning. The agent consistently grounds linguistic cues to visual landmarks and maintains coherent subgoal transitions, successfully reaching the destination.
We also provide full navigation videos of FineCog-Nav and baselines in the Suppl.  
Additional visualizations of failure cases are provided in the Suppl., showing that even when the goal is not reached, our method continues to follow instructions and seek landmarks.

\noindent \ding{96} \textbf{Sentence-level Analysis.}
Recall that AerialVLN-Fine provides sentence-level annotations for fine-grained evaluation. 
Since our method can explicitly indicate sentence completion, we report sentence-level evaluation results of FineCog-Nav in \cref{tab:sentence_result}.
We observe that as the base LLM size increases, performance steadily improves. Although there is a gap between the smallest and 72B models, it remains moderate, reflecting the robustness of our framework. 
Notably, small sentence-level differences can accumulate over long trajectories, leading to larger gaps in full-trajectory metrics, especially in long-horizon 3D navigation, highlighting the need for fine-grained evaluation.

\noindent \ding{96} \textbf{Human Study.}
\begin{figure}[t]
    \centering
    \includegraphics[width=1\linewidth]{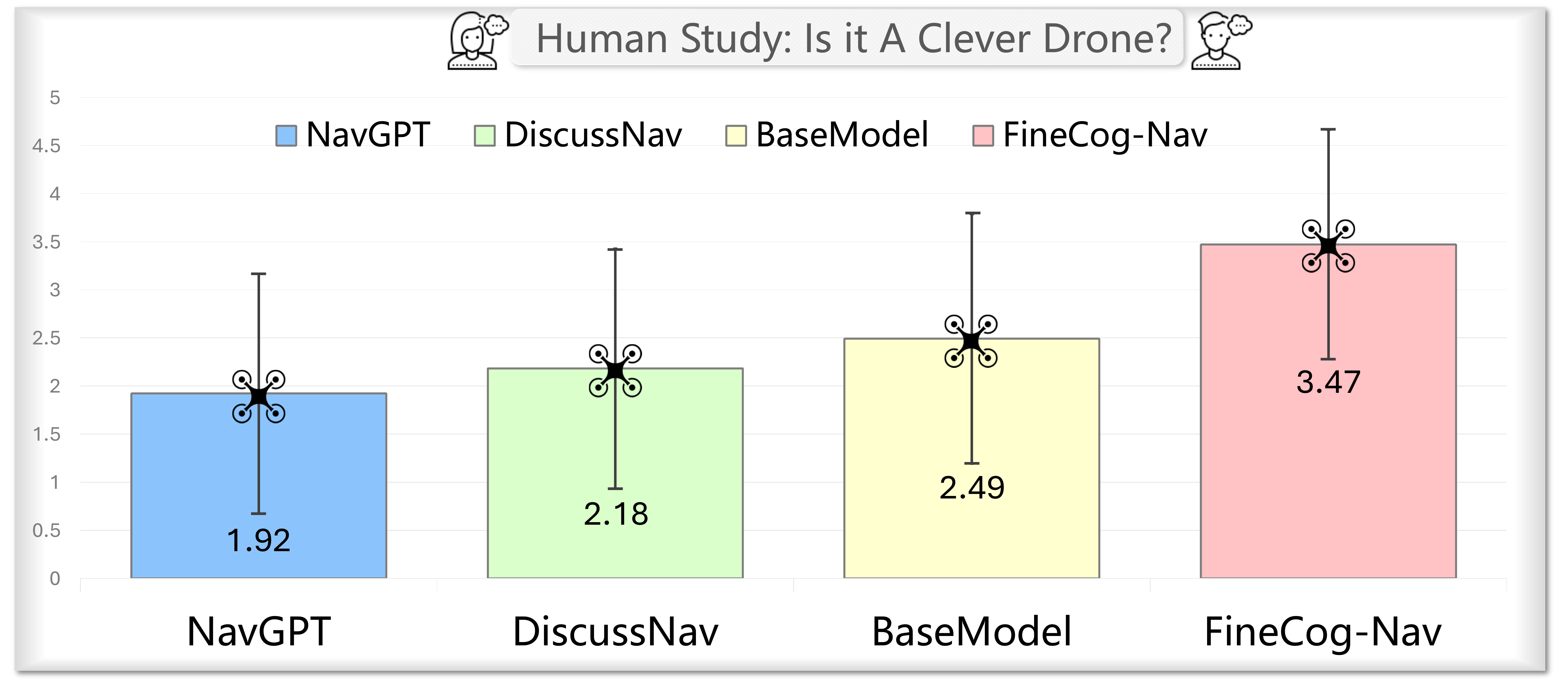}
    \caption{\textbf{Visualization of Human Study Results.} FineCog-Nav is consistently favored over baselines, even in failure cases, for better instruction following and more human-like exploration.
}
    \label{fig:human_study}
\end{figure}
Standard metrics are often uninformative for challenging long-horizon UAV VLN tasks, where performance is uniformly poor across methods. We therefore conducted a human study to evaluate overall navigation quality and instruction following, including failure cases. Specifically, we randomly selected 10 instructions and presented anonymized first-person navigation videos from our method and baselines. Approximately 70 participants rated each video from 1 (worst) to 5 (best), with at most one video per comparison group assigned a score of 5. All methods used the same foundation models for fairness. As shown in \cref{fig:human_study}, our method was consistently preferred by annotators, likely due to better instruction following and more human-like exploration (see Suppl. for details).

\noindent \ding{96} \textbf{Real-World Deployment and Future Work.}
To complement simulation results, we deploy FineCog-Nav on a RoboMaster TT UAV and conduct a preliminary real-world flight test. Given the instruction \textit{``Find a chair for me, then land on the table''}, the agent performs stepwise exploration, selects intermediate subgoals, and reaches the target after 17 steps.
This experiment provides an initial feasibility check beyond simulation. Handling real-world disturbances such as sensor noise and control error remains future work.

\begin{figure}[t]
    \centering
    \includegraphics[width=1.0\linewidth]{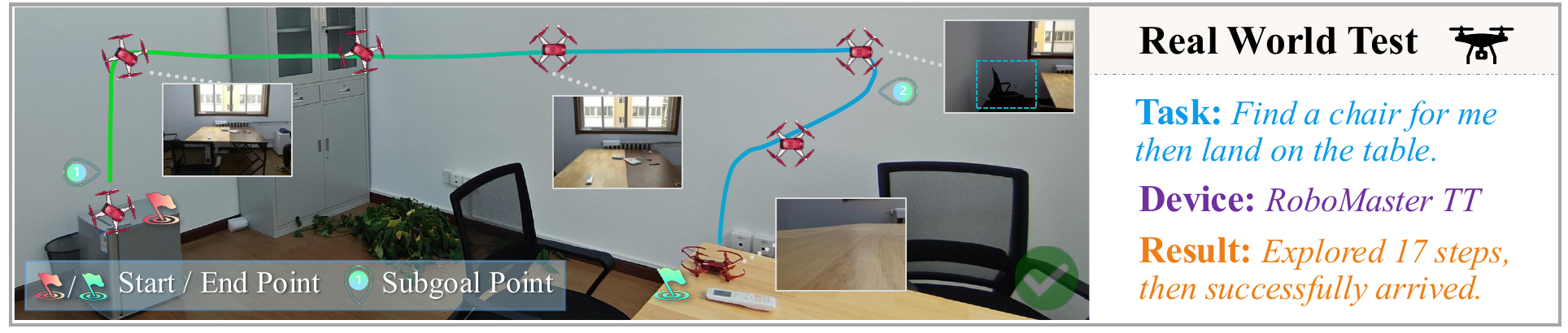}
    \caption{\textbf{Preliminary real-world deployment of FineCog-Nav}. For the given instruction, the agent reaches the target region after 17 steps. Start, subgoal, and end points are shown.}
    \label{fig:real_world}
\end{figure}

\section{Conclusion}
In this paper, we address the challenging task of UAV VLN, where existing zero-shot frameworks based on LFMs fall short. We present FineCog-Nav, the first zero-shot framework that explicitly models core, interdependent cognitive functions at a fine granularity, enabling robust and interpretable UAV navigation without reliance on large-scale models. To facilitate fine-grained evaluation, we introduce AerialVLN-Fine, a curated dataset with sentence-level alignment and refined instructions. Extensive experiments and human studies demonstrate that FineCog-Nav outperforms diverse baselines, establishing a new direction for cognitively inspired, modular architectures in embodied AI.

\section*{Acknowledgement}
This work was funded by the National Natural Science Foundation of China (NSFC) under Grant 62306239 and Grant 62506153, and also supported by the Sanqin Talents Introduction Plan of Shaanxi Province.

{
    \small
    \bibliographystyle{ieeenat_fullname}
    \bibliography{main}
}

\clearpage

\maketitlesupplementary
\addtocontents{toc}{\protect\setcounter{tocdepth}{2}}
\tableofcontents
\section{Dataset Details }
\begin{figure*}[h]
    \centering
    \includegraphics[width=1.0\linewidth]{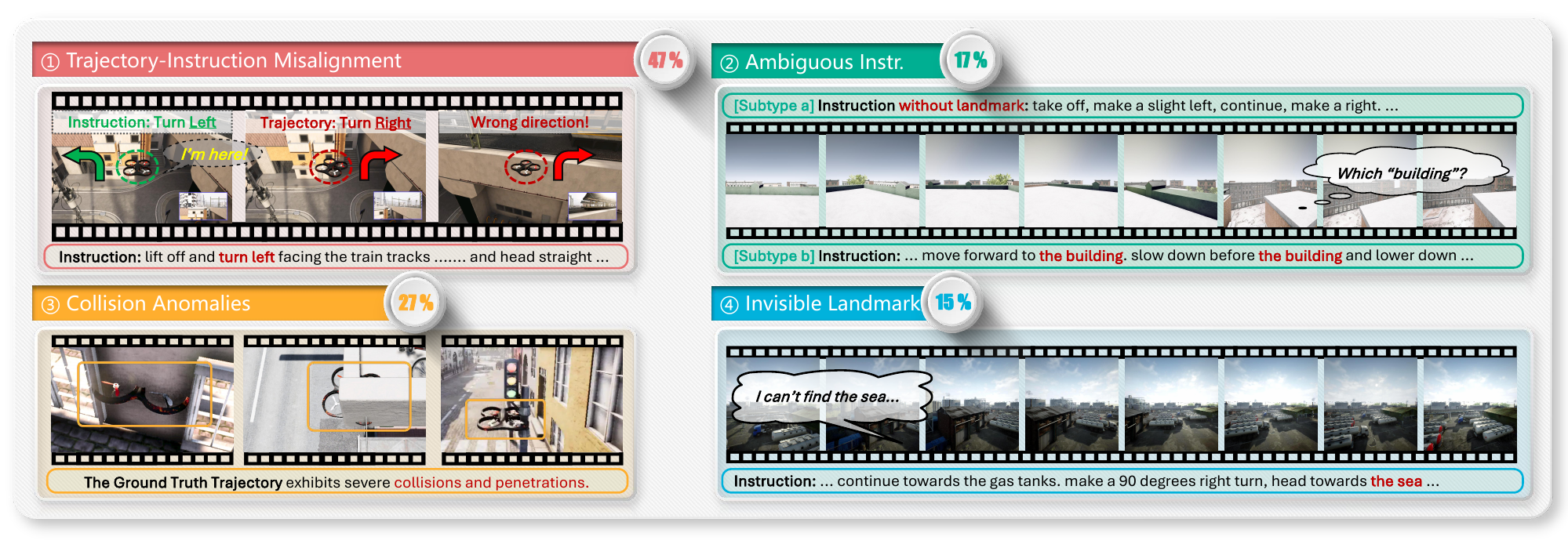}
    \caption{
   Through manual analysis of 200 randomly sampled instruction-trajectory pairs, we quantify four prevalent issues: \ding{40} Trajectory-Instruction Misalignment (47\%); \ding{40} Collision Anomalies (27\%); \ding{40} Ambiguous Instructions (17\%), and \ding{40} Invisible Landmarks (15\%). }
    \label{fig:dataset_issues}
\end{figure*}

\subsection{Analysis of Issues in AerialVLN Dataset}
While AerialVLN serves as a challenging and milestone benchmark for UAV VLN, we have observed a substantial proportion of flawed data samples during experimental analysis. To systematically analyze these issues, we randomly sampled 200 trajectory-instruction pairs from the AerialVLN-S subset and conducted a careful manual inspection. The issues are categorized into four main types, as illustrated in \cref{fig:dataset_issues}:

\ding{224} \textbf{Trajectory-Instruction Misalignment} is the most prevalent issue, affecting nearly half of the samples. Such as the instruction specifying a left turn while the corresponding ground-truth trajectory turns right;

\ding{224} \textbf{Collision Anomalies} rank second, comprising physically implausible trajectories that involve minor clipping (e.g., through fences/lampposts) or major clipping into walls/ground;
     
\ding{224} \textbf{Ambiguous Instructions} manifest as two subtypes: a) Instructions lacking spatial or landmark references, leaving agents uncertain about environmental interactions post-action; b) Vague landmark descriptions (e.g., generic terms like ``building'' or ``tree'') that lack distinctive visual correspondence in outdoor aerial scenes.

\ding{224} \textbf{Invisible Landmarks} mean that landmarks mentioned in instructions as references are visually absent or unobservable in corresponding viewpoints, severely hindering navigation grounding.

\begin{figure*}[h]
    \centering
    \includegraphics[width=1.0\linewidth]{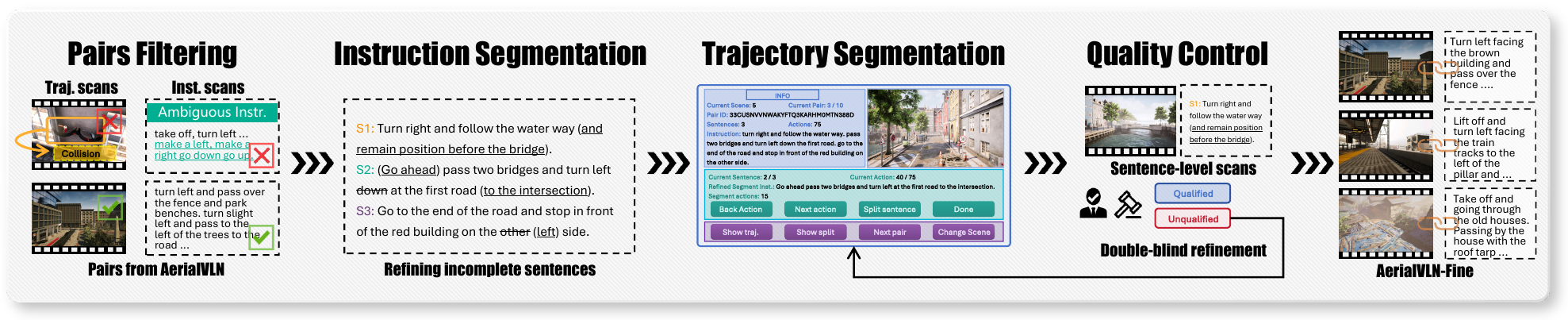}
    \caption{
   The construction process of AerialVLN-Fine, including pairs filtering, instruction segmentation, trajectory segmentation, and double-blind quality control to ensure high-quality, sentence-level annotations. }
    \label{fig:dataset_construction}
\end{figure*}
\begin{table}[h]
    \caption{Detailed statistics of AerialVLN-Fine by scene.}
    \label{tab:tab_supp_dataset_stat_by_scene}
    \centering
    \setlength{\tabcolsep}{6pt}
    \begin{tabular}{ c | c  c   c  c }
    \toprule
    \textbf{Scene ID} & \textbf{\# Traj.} & \textbf{Path Len} & \textbf{\# Action} & \textbf{\# Sent.} \\
    \midrule
    2  & 16 & 279.19 & 92.13 & 5.19 \\ 
    3  & 32 & 121.91 & 72.47 & 4.56 \\
    5  & 10 & 315.80 & 128.20 & 6.20 \\
    7  & 30 & 210.83 & 71.70 & 4.70 \\
    8  & 35 & 103.26 & 47.40 & 4.29 \\
    9  & 30 & 255.50 & 80.03 & 4.20 \\
    10 & 10 & 334.60 & 115.50 & 5.20 \\
    11 & 20 & 413.10 & 125.45 & 4.65 \\
    12 & 20 & 167.65 & 67.05 & 4.95 \\
    13 & 32 & 98.69  & 62.16 & 4.25 \\
    14 & 17 & 137.59 & 70.06 & 4.53 \\
    17 & 30 & 96.17  & 58.87 & 4.07 \\
    20 & 2  & 425.50 & 143.00 & 7.50 \\
    21 & 4  & 265.50 & 80.00 & 3.50 \\
    24 & 12 & 138.67 & 82.67 & 5.58 \\
    \bottomrule    
    \end{tabular}
\end{table}
\begin{table*}[t] 
\caption{Qualitative Analysis of Sentence-Level Instructions}
\label{tab:tab_supp_dataset_qualitative_analysis}
\small
\setlength{\tabcolsep}{4pt} 
\renewcommand{\arraystretch}{1.1} 
\begin{tabularx}{\textwidth}{
    >{\hsize=0.8\hsize\raggedright\arraybackslash}X 
    >{\hsize=1.2\hsize\raggedright\arraybackslash}X 
    >{\hsize=1.4\hsize\raggedright\arraybackslash}X 
    >{\hsize=0.4\hsize\raggedleft\arraybackslash}X 
    >{\hsize=0.5\hsize\raggedleft\arraybackslash}X 
}
\toprule
\textbf{Capability Dimension} & \textbf{Description} & \textbf{Example} & \textbf{Count} & \textbf{Proportion/\%} \\
\midrule
Spatial Relations & Understanding and processing relative positional relationships & Land to the right of the van that looks like the mystery machine from Scooby-Doo. & 1387 & 99.36 \\

Multi-Target Path Planning & Handling path planning between multiple locations & At the end of the road go straight across \textbf{the canal} then turn left facing \textbf{the road with bridges} on the left and head straight. & 736 & 52.72 \\

Trajectory Constraints & Following specific trajectory shapes or restrictions & Lower flight height and fly in a slow complete circle and maintain position. Maintain 5m altitude near the red warehouse. & 564 & 40.40 \\

Temporal Relations & Handling time-based task sequences & \textbf{Turn left after lifting off} and face the dog mural then head straight. & 413 & 29.58 \\

Ordinal/Cardinal Recognition & Understanding and processing ordinal or cardinal information in instructions & Turn around at \textbf{the fourth white building} in your path. & 207 & 14.83 \\

\bottomrule
\end{tabularx}
\end{table*}

\subsection{Detailed Construction Process}
To ensure the high quality and representativeness of AerialVLN-Fine, we designed a systematic filtering and re-annotation pipeline that combines automated checks with manual verification, as shown in \cref{fig:dataset_construction}. This process improves data purity, completeness, and annotation consistency. The main steps are as follows:

 \noindent \ding{182} \textbf{Manual Review and Selection:} Each instruction-trajectory pair is manually examined against four key quality criteria. Erroneous or incomplete samples are removed, and for each scene, an average of 20 diverse and high-quality trajectories are retained to ensure both diversity and balance.
    
\noindent \ding{183} \textbf{Instruction Segmentation and Refinement:} Complex instructions are semantically decomposed and numbered. Each sentence is checked for clear actions and termination conditions. Ambiguous or missing actions and landmarks are clarified or added based on the context and trajectory. Visual anchors along the trajectory are explicitly described to guide the navigation. Key principles include:
    \begin{itemize}
        \item Each action is paired with a clear stopping condition, typically linked to a visible landmark (e.g., ``proceed to the truck'', ``turn left to face the graffiti'').
        \item Landmark descriptions are made unique and unambiguous, using type, color, text, or surrounding context as needed.
        \item Landmarks are ensured to be visible; if not, a visible anchor is added.
        \item Instructions are written in the first person and kept concise, covering actions such as move forward, turn left/right, ascend, or descend.
    \end{itemize}
    
\noindent \ding{184} \textbf{Trajectory Segmentation and Alignment:} Trajectories are finely segmented according to the termination conditions of each sub-instruction, aligning each with its corresponding waypoint and action sequence. This results in more granular instruction-to-trajectory mappings.
    
\noindent \ding{185}  \textbf{Rigorous Quality Control:} After annotation, multiple rounds of manual review are performed based on the quality criteria to further eliminate errors.

Through this process, AerialVLN-Fine achieves high accuracy, consistency, randomness, and trajectory realism, covering a broad range of scenarios. This provides a solid foundation for fair algorithm comparison and capability assessment, supporting in-depth research and innovation in UAV Vision-and-Language Navigation.

\begin{figure*}[h]
    \centering
    \includegraphics[width=1.0\linewidth]{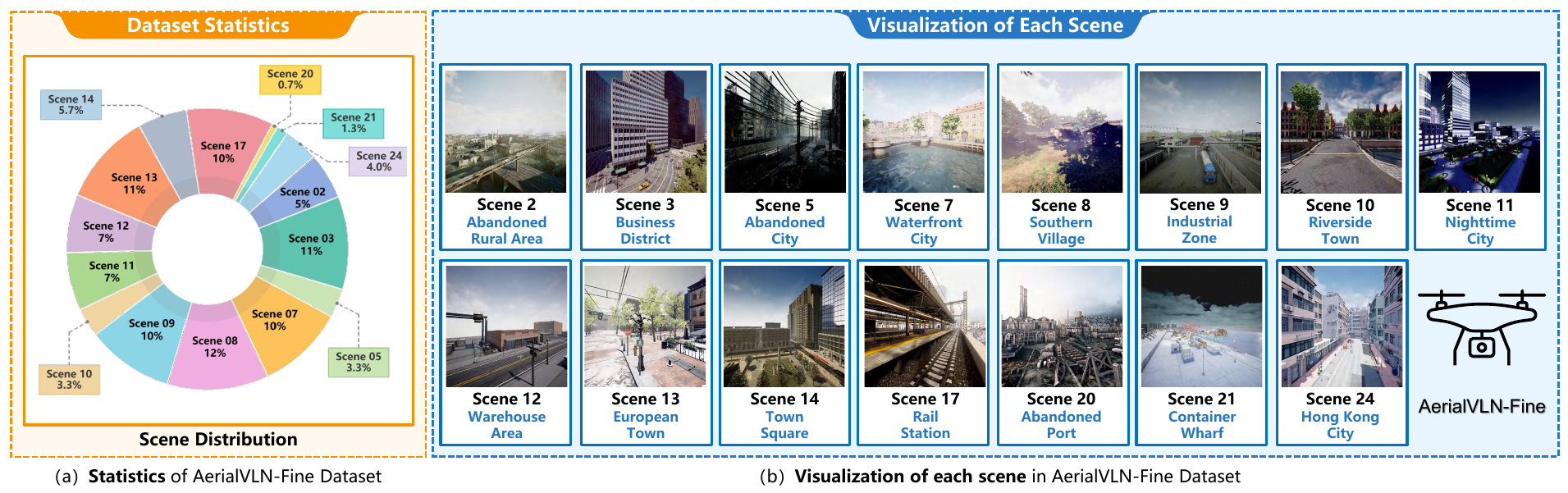}
    \caption{Distribution and Demonstration of scenes in AerialVLN-Fine. 15 scenes cover day, night, city, rural areas, and contain perceptual challenges such as similar landmarks and visual obstructions.}
    \label{fig:dataset_overview}
\end{figure*}

\subsection{Dataset Statistics} \label{dataset_statics}
In \cref{tab:tab_supp_dataset_stat_by_scene}, we display detailed statistics of the AerialVLN-Fine dataset for each scene, including the number of trajectories, average path length, average number of actions, and average number of sentences per trajectory. 
We can see that the AerialVLN-Fine dataset exhibits several notable characteristics:

\ding{172} First, it encompasses a diverse set of 15 scenes, providing a wide range of environments for evaluation;

\ding{173} Second, the navigation trajectories are considerably \textbf{longer} and more \textbf{complex} compared to those in existing indoor VLN benchmarks. For example, while the R2R dataset features an average path length of approximately 10 meters with around 5 actions per trajectory, and RxR reports an average path length of 15 meters and 8 actions, the latest OctoNav-Bench includes trajectories averaging 8.2 meters in length (with the longest reaching 42 meters). 
In stark contrast, trajectories in AerialVLN-Fine require a substantially greater number of actions and cover much longer distances on average—specifically, 187 meters in path length and 76 actions per trajectory, underscoring its significantly higher navigation complexity.

\ding{174} Additionally, there is substantial variation in both path length and number of actions across different scenes, with some environments presenting significantly more challenging navigation tasks. Furthermore, the average number of sentences per trajectory generally ranges from 4 to 7, which can be interpreted as the number of sub-tasks or instructions required to complete each navigation episode. 

In \cref{fig:dataset_overview}, we show the distribution and demonstration of scenes in AerialVLN-Fine. 
The dataset exhibits rich environmental diversity and realistic perceptual challenges, which can be summarized as follows:

\ding{95} \textbf{Urban areas}: Scenes such as \textit{Abandoned City} (Scene~5), \textit{Nighttime City} (Scene~11), \textit{Rail Station} (Scene~17), and \textit{Hong Kong City} (Scene~24) feature dense high-rise buildings, complex street layouts, and varying levels of urban decay. These settings test navigation robustness in cluttered metropolitan areas.

\ding{95} \textbf{Rural and suburban areas}: \textit{Abandoned Rural Area} (Scene~2), \textit{Southern Village} (Scene~8), \textit{Industrial Zone} (Scene~9) and \textit{Warehouse Area} (Scene~12) represent low-density settlements with sparse landmarks, irregular structures, and natural terrain, posing challenges in localization due to limited visual cues.

\ding{96} \textbf{Low-light and adverse weather conditions}: \textit{Nighttime City} (Scene~11) and \textit{Container Wharf} (Scene~21) are nighttime environments, while several scenes—including \textit{Abandoned City} (Scene~5) and \textit{Abandoned Port} (Scene~20)—contain fog or overcast conditions. These scenarios severely degrade color fidelity and landmark visibility.

\ding{96} \textbf{Visual occlusion and ambiguous landmarks}: Dense foliage in \textit{Business District} (Scene~3) and \textit{Town Square} (Scene~14), combined with collapsed structures in \textit{Abandoned Rural Area} (Scene~2) and \textit{Abandoned City} (Scene~5), leads to severe view obstruction. Moreover, repetitive architectures in urban and industrial zones cause landmark ambiguity and confusion.
 
In addition, \cref{tab:tab_supp_dataset_qualitative_analysis} presents a qualitative analysis of the capability dimensions reflected in sentence-level instructions within the AerialVLN-Fine dataset. The analysis covers five key capability dimensions: spatial relations, multi-target path planning, trajectory constraints, temporal relations, and ordinal/cardinal recognition. The results show that spatial relations account for the vast majority of instructions (99.36\%), indicating that most navigation tasks require the agent to understand and reason about relative positional relationships. Multi-target path planning is the second most common capability, appearing in 52.72\% of the instructions, reflecting the frequent need to navigate through multiple intermediate locations. Trajectory constraints are also prevalent, occurring in 40.40\% of the instructions, which suggests common requirements for specific flight patterns or altitude restrictions. Temporal relations are present in 29.58\% of cases, highlighting the importance of time-ordered actions, while ordinal/cardinal recognition appears in 14.83\% of instructions, indicating a moderate need for understanding numerical cues.
These findings highlight the diversity and complexity of navigation instructions in AerialVLN-Fine, reflecting a wide range of real-world challenges that test an agent’s ability to comprehend and execute various types of navigation tasks. We present a capability-aware ablation study in \cref{capability_ablation}, where we systematically evaluate how each cognitive module contributes to these distinct reasoning demands.

\section{More Analysis}

\begin{figure*}[h]
    \centering
    \includegraphics[width=1.0\linewidth]{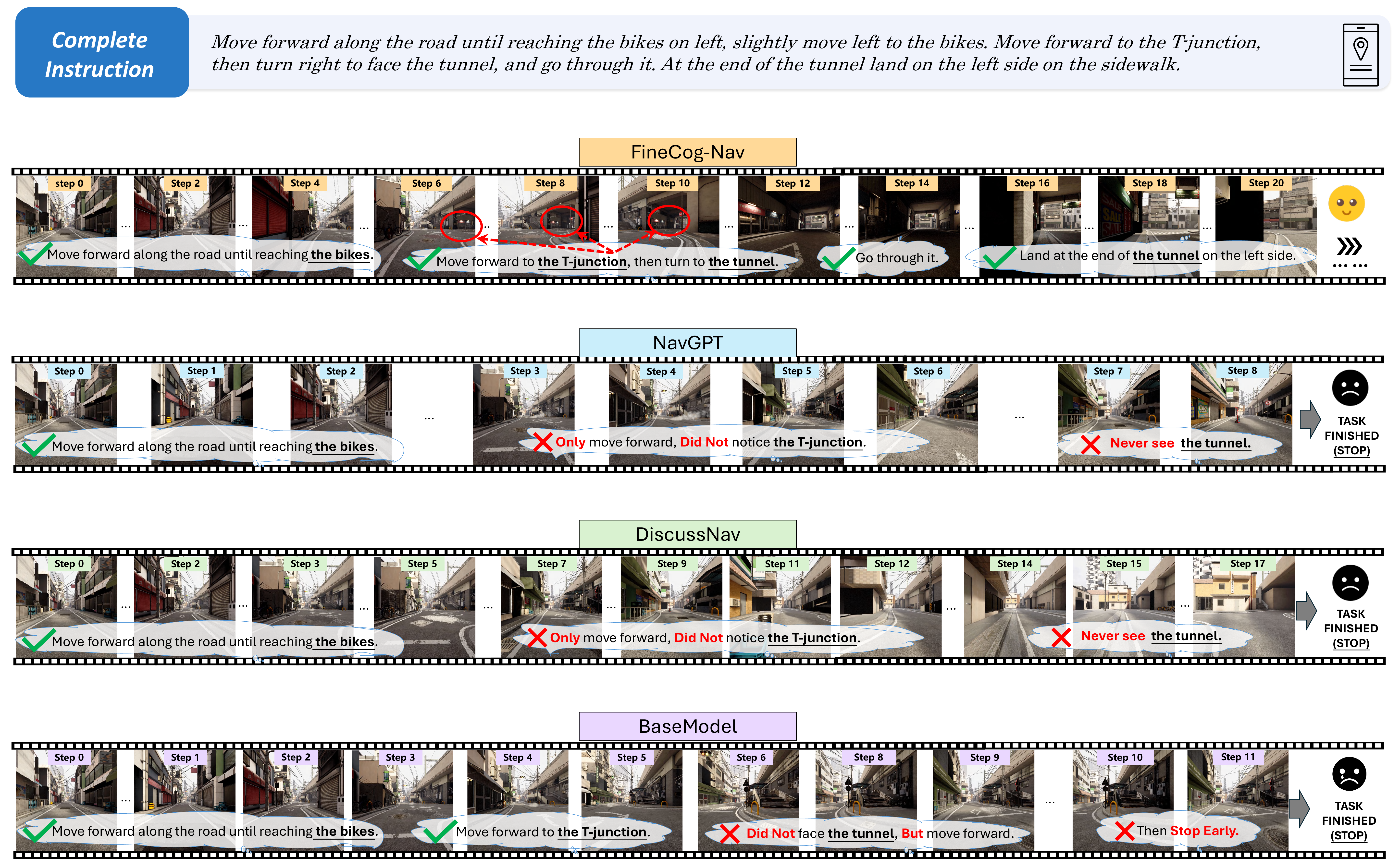}
    \caption{Qualitative comparison of FineCog-Nav and baselines on a challenging UAV VLN episode. FineCog-Nav demonstrates superior instruction-following by correctly identifying the key landmarks (the T-junction, the tunnel), while baseline methods overlook key landmarks, terminate prematurely, or encounter navigation errors. }
    \label{fig:quali_failure}
\end{figure*}

\subsection{Demo Video}
To provide a more comprehensive and intuitive understanding of the navigation process, we additionally include a demonstration video in the supplementary material, to which we kindly refer interested readers for further insight.
In this demo, we showcase not only the full navigation trajectory of our proposed \textbf{FineCog-Nav}, but also compare it against the single-module baseline (\textit{BaseModel}) and two framework-level baselines (\textit{NavGPT} and \textit{DiscussNav}) on representative navigation tasks with natural language instructions, sampled from the AerialVLN-Fine dataset.

As illustrated in the demo, FineCog-Nav successfully accomplishes the navigation task, faithfully following the provided instructions throughout the trajectory. In contrast, the baseline methods demonstrate a variety of failure modes, such as early deviation from the intended route, neglecting instructions, traversing through obstacles, and aimlessly navigating towards irrelevant areas (e.g., the sky or sea), among other issues.

\subsection{Qualitative Comparison with Baselines}

Since navigating UAVs in 3D continuous environments for Vision-and-Language Navigation (VLN) tasks remains highly challenging, existing zero-shot methods based on large language and vision-language models (LLMs and VLMs) generally achieve low success rates. Therefore, we also provide a qualitative analysis of failure cases to demonstrate that, even under these difficult conditions where all methods fail to reach the success rate threshold, our FineCog-Nav model exhibits markedly superior behavior.

As illustrated in \cref{fig:quali_failure}, we examine a challenging episode with the following instruction (ID: 3WJEQKOXAH0G632K1XLQY3JARLMA18):
\begin{quote}
    \textit{``Move forward along the road until reaching the bikes on left, slightly move left to the bikes. Move forward to the T-junction, then turn right to face the tunnel, and go through it. At the end of the tunnel land on the left side on the sidewalk."}
\end{quote}

We can see that FineCog-Nav demonstrates a strong ability to follow complex instructions and to perceive and utilize environmental cues. Specifically, our model correctly detects the bikes as an initial landmark, then successfully navigates toward and recognizes the T-junction, and further identifies and enters the tunnel as instructed. Notably, FineCog-Nav is the only method to correctly complete the critical step of reaching the tunnel entrance, traversing it, and finally landing at the instructed location on the sidewalk. This sequence showcases the model’s precise step-by-step alignment with the provided instructions and robust landmark grounding throughout the navigation episode.

In contrast, the baseline methods show pronounced limitations:

\ding{169} \textbf{NavGPT} fails to notice the T-junction, continuing straight past the intended turning point. Lacking awareness of the tunnel, NavGPT never attempts to enter it and terminates the task after only 8 steps.

\ding{169} \textbf{DiscussNav} also overlooks the T-junction and continues without attempting to locate the tunnel, resulting in premature task termination before reaching the intended goal.

\ding{171} For the single-module baseline (\textbf{BaseModel}), the model similarly fails to ground the instruction to "face the tunnel," proceeds straight ahead without adjustment, and ultimately stops early.

\begin{table*}[h]
    \caption{\textbf{Per-Capability Ablation Analysis on $SR^{3D}$ (\%).} 
    The last row shows absolute Success Rates of the full model; other rows show $\Delta SR^{3D}$ relative to full. 
    For each ablation, the largest and second-largest performance drops are marked with bold and underline, respectively.}
    \small
    \label{tab:capability_ablation_checkmark}
    \centering
    \begin{tabular}{ccc | cccccc}
    \toprule
    \multicolumn{3}{c|}{\textbf{Module}} & 
    \multicolumn{5}{c}{\textbf{Capability Dimension (SR \% or $\Delta$SR \%)}} \\
    Attention & Imagination & Subgoal &
    Spatial & Multi-Target & Trajectory & Temporal & Ordinal/ \\
    & & & Relations & Path Planning & Constraints & Relations & Cardinal \\
    \midrule
    
    \ding{56} & \ding{51} & \ding{51} & 
    \underline{-3.00} & -2.85 & -2.75 & -2.60 & \textbf{-4.08} \\
    
    \ding{51} & \ding{56} & \ding{51} & 
    -2.33 & \underline{-2.49} & \textbf{-2.75} & -0.52 & -2.04 \\
    
    \ding{51} & \ding{51} & \ding{56} & 
    \underline{-4.00} & -3.91 & \textbf{-4.71} & -2.08 & -3.40 \\
    
    \rowcolor{yellow!10}
    \ding{56} & \ding{56} & \ding{51} & 
    -2.33 & -2.49 & \textbf{-3.53} & \underline{-3.13} & -2.72 \\
    
    \rowcolor{yellow!10}
    \ding{51} & \ding{56} & \ding{56} & 
    -2.00 & \underline{-2.14} & \textbf{-2.75} & -1.04 & -2.04 \\

    \rowcolor{cyan!10}
    \ding{51} & \ding{51} & \ding{51} & 
    6.00 & 5.69 & 6.27 & 4.69 & 4.76 \\
    \bottomrule
    \end{tabular}
\end{table*}

In summary, these qualitative observations highlight that, although all methods face substantial challenges in the UAV VLN task, our FineCog-Nav stands out in its ability to consistently ground on-step instructions, accurately perceive and act upon key environmental landmarks, and maintain coherent progress towards the target. In contrast, baseline models frequently miss crucial cues, lose alignment with the instructions, and terminate navigation without fulfilling the task.

\subsection{Quantitative Analysis of Ablations} 
\label{capability_ablation}

\begin{figure*}[h]
    \centering
    \includegraphics[width=1.0\linewidth]{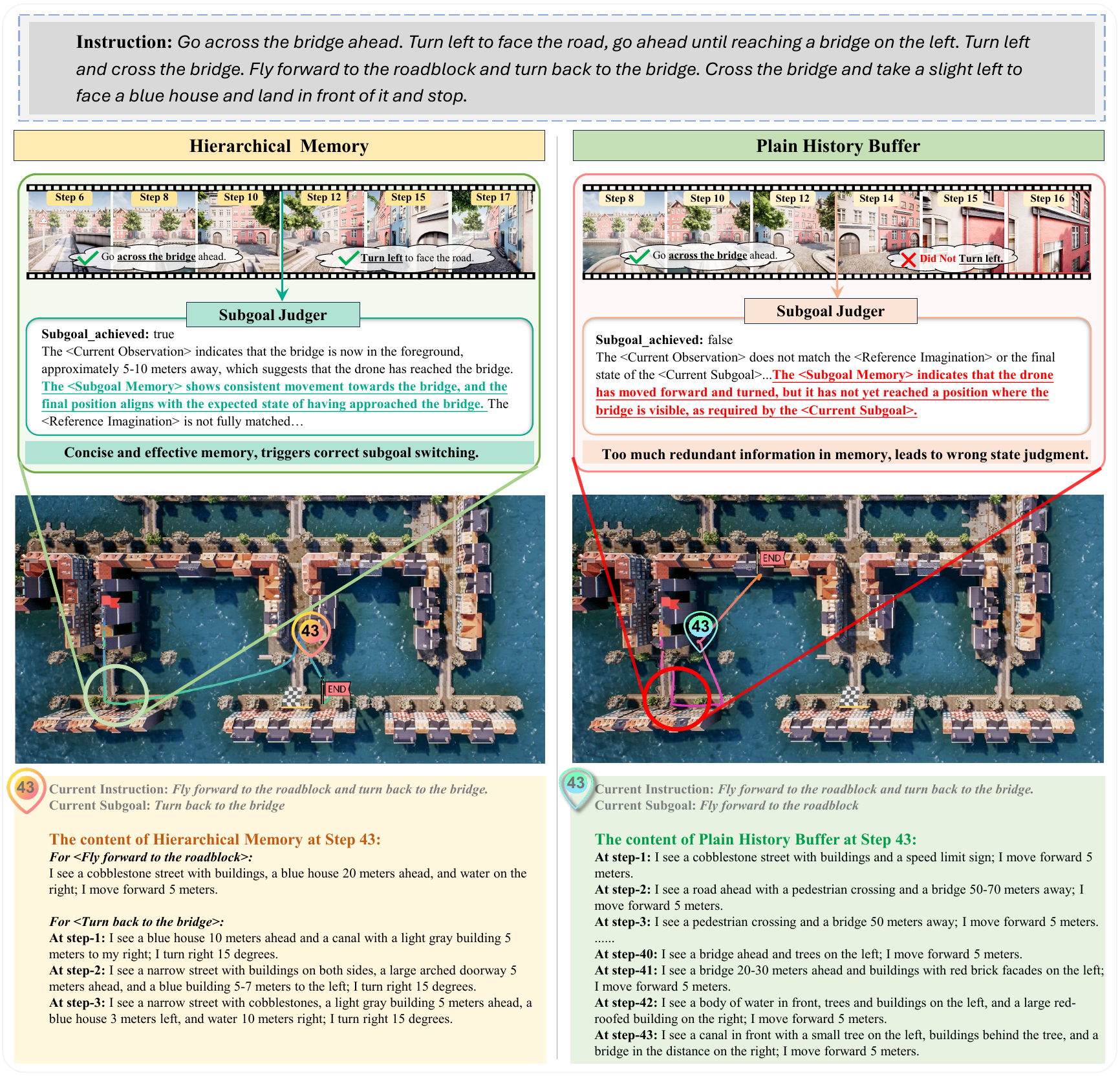}
    \caption{\textbf{Comparison of hierarchical memory and plain history buffer} in a complex navigation scenario. The top panels illustrate the impact on subgoal switching: hierarchical memory enables timely and accurate transitions, while the flat buffer leads to delayed or incorrect switching. The bottom panels compare memory content at Step 43 under the same subgoal, in which Hierarchical memory provides a concise, structured summary of key events, whereas the plain buffer stores a lengthy, unstructured sequence, leading to information overload and ambiguous decisions.}
    \label{fig:quali_ablation_mem}
\end{figure*}

\begin{figure*}[h]
    \centering
    \includegraphics[width=1.0\linewidth]{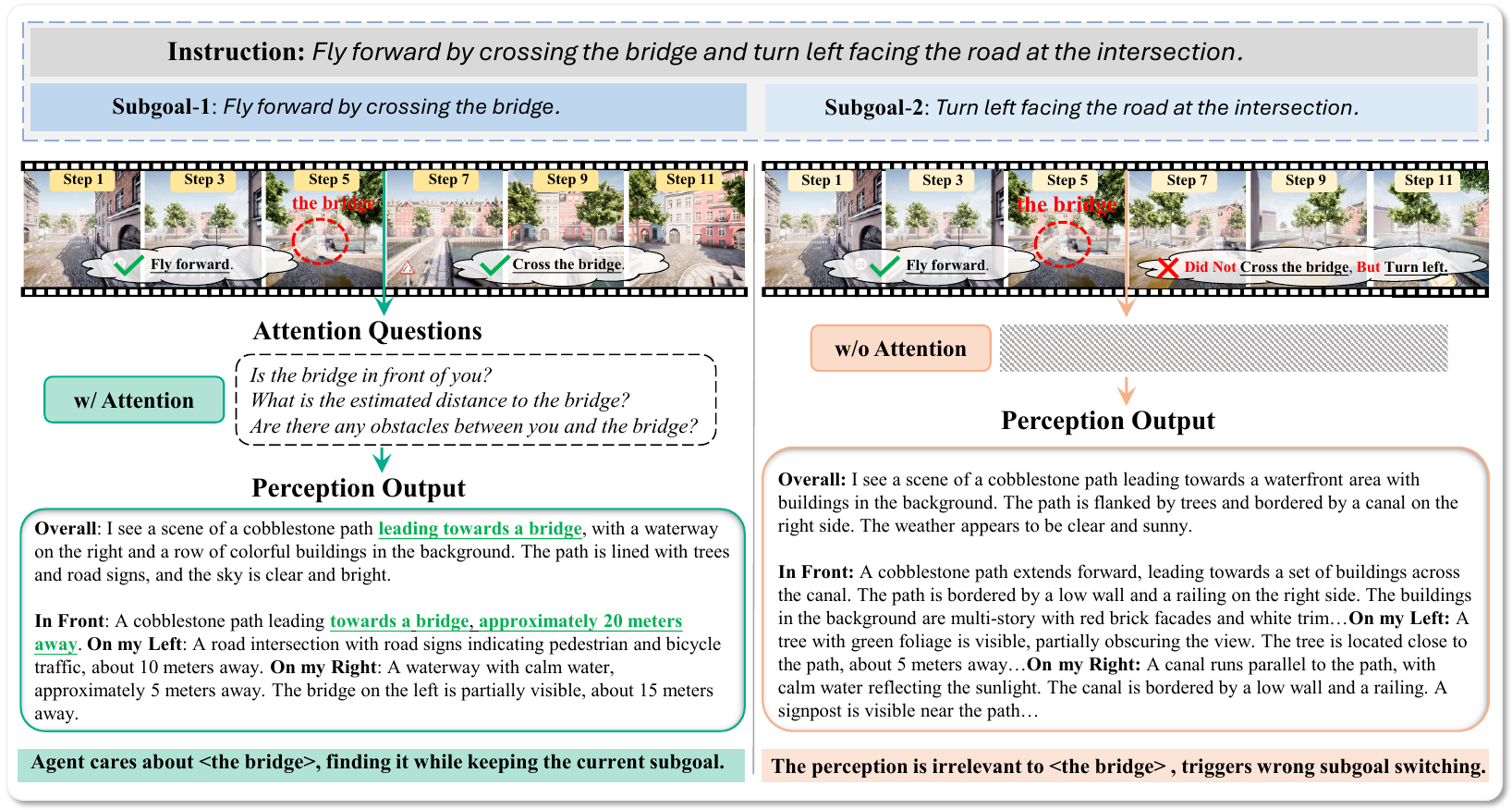}
    \caption{Illustration of a significant difference in Perception outcomes when handling the same scenario with and without the Attention module. This highlights the guiding role of the Attention module in directing the Perception process, as well as its collaborative influence on subsequent instruction transitions and the decision-making modules.}
    \label{fig:quali_ablation_attn}
\end{figure*}

We present further quantitative ablation analysis to elucidate the functional contributions of each module in FineCog-Nav.
To understand how each component contributes to distinct capabilities, we analyze the 3D success rate ($SR^{3D}$) across the five capability dimensions defined in \cref{dataset_statics}: \textit{spatial relations}, \textit{multi-target planning}, \textit{trajectory constraints}, \textit{temporal relations}, and \textit{ordinal/cardinal understanding}.

As summarized in \cref{tab:capability_ablation_checkmark}, removing any single module degrades performance, with the degree and affected dimensions reflecting the specialized function of each component:

\noindent
\ding{56} \textbf{Hierarchical Memory}: Critical Infrastructure.
Consistent with our main results, ablating the \textit{Hierarchical Memory} module (replacing it with a simple flat history buffer) causes the largest and most pervasive performance drops across all capability dimensions (e.g., $-5.33$ for spatial relations, $-5.49$ for trajectory constraints, and $-4.69$ for temporal relations). 
This underscores that memory serves not as an auxiliary module, but as critical infrastructure that underpins and interconnects all reasoning components. As its impact is system-wide and not independent, we omit memory ablation results from the table. To elaborate a little, the hierarchical structure is especially effective for maintaining long-term contextual coherence, enabling robust information integration and planning over extended trajectories. Its absence disrupts both individual module effectiveness and overall system synergy, further validating our design.

\ding{56} \textbf{Attention}:
Ablating the \textit{Attention} leads to the sharpest decline in \textit{ordinal/cardinal understanding} ($-4.08$), indicating that this module is essential for grounding numerical and sequential cues in language; without robust attention, the agent struggles with fine-grained language-visual alignment and stepwise reasoning.

\ding{56} \textbf{Imagination}: 
 Ablating \textit{Imagination} results in a notable drop in \textit{trajectory constraints} ($-2.75$), and also affects \textit{spatial relations} and \textit{multi-target planning} to a lesser extent. This indicates that Imagination is crucial for simulating potential outcomes and evaluating subgoal feasibility, which is indispensable for satisfying path constraints and planning multi-stage routes.

\ding{56} \textbf{Subgoal}:  
Ablating the \textit{Subgoal} mechanism causes the most significant drops in \textit{spatial relations} ($-4.00$), \textit{multi-target planning} ($-3.91$), and \textit{trajectory constraints} ($-4.71$), reflecting its central role in decomposing complex instructions and structuring long-horizon navigation. The inability to break a task into manageable subgoals leads to compounding errors and loss of strategic planning.

Compound ablations further highlight the interdependence among modules:

\ding{56} \textbf{Attention \& Imagination}: 
Ablating both modules leads to the greatest drops in \textit{trajectory constraints} ($-3.53$) and \textit{temporal relations} ($-3.13$), indicating that combining perceptual grounding with prospective scene imagination is critical for satisfying complex spatial constraints and ensuring the correct temporal ordering and execution of navigation subgoals.

\ding{56} \textbf{Imagination \& Subgoal}:  
Ablating these two modules causes severe degradation in \textit{trajectory constraints} ($-2.75$) and \textit{multi-target planning} ($-2.14$), as \textit{Subgoal} provides structured decomposition and \textit{Imagination} evaluates transition feasibility. Removing both disrupts the planning–verification loop that is central to coherent long-horizon navigation.

In summary, the above quantitative analysis demonstrates that the modules in FineCog-Nav fulfill complementary roles within the cognitive workflow, and their integration is crucial for achieving robust performance across the diverse and challenging scenarios in UAV VLN tasks.

\subsection{Qualitative Analysis of Ablations} 
We qualitatively examine how ablating core modules affects agent behavior by visualizing representative examples, focusing specifically on the Hierarchical Memory and Attention modules.

\noindent \ding{43} \textbf{Analysis of Hierarchical Memory.}
\cref{fig:quali_ablation_mem} illustrates both the performance consequences of replacing hierarchical memory with a plain history buffer, and the differences in memory representations at a critical navigation stage.

\ding{172} \textit{Performance Impact:} Ablating hierarchical memory impairs the agent’s ability to execute complex instructions. With only a flat history buffer, the agent struggles to detect subgoal completion, delays subgoal switching, and makes navigation errors. In contrast, hierarchical memory maintains coherent progress tracking and enables timely, accurate subgoal transitions.

\ding{173} \textit{Memory Content Comparison:} At Step 43, under the instruction “Fly forward to the roadblock and turn back to the bridge” with the current subgoal “Turn back to the bridge”,
the hierarchical memory encodes a concise, structured summary of key events and subgoal states. This abstraction supports precise state judgment and efficient instruction execution. Conversely, the plain history buffer accumulates all 43 low-level observations and actions in an unorganized list, resulting in information overload and ambiguous decision-making.

To summarize, Hierarchical memory provides compact and structured representations essential for subgoal tracking and long-horizon planning. Its absence leads to redundant internal states and degraded navigation performance.

\noindent \ding{43} \textbf{Analysis of Attention.}
\cref{fig:quali_ablation_attn} compares the agent’s behavior with and without the Attention module. To elaborate, when executing the instruction \textit{``Fly forward by crossing the bridge and turn left facing the road at the intersection''}, the behaviors diverge clearly:

\ding{52} \textit{With Attention}:
Attention-driven queries keep the agent focused on verifying whether the bridge is ahead, its distance, and potential obstacles. This targeted perceptual grounding enables the agent to locate the bridge, cross it reliably, and maintain the correct subgoal sequence.

\ding{56} \textit{Without Attention}:
Without the attention module, the agent’s perception becomes scattered and shifts away from the bridge. As a result, the agent ignores the bridge and wrongly extracts subgoals from the instruction, failing to cross the bridge and making incorrect decisions.

This qualitative result highlights how Attention guides Perception and how their interaction produces different task outcomes, illustrating the synergistic effect among modules in our framework.

\section{Human Study Details}
In this section, we first present the detailed design of the human study questionnaire, followed by correlation and significance analyses of the human study results.

\subsection{Design of the Human Study Process}
The questionnaire consists of three parts: an introduction page, ten evaluation pages, and a farewell page. Below, we present the content and illustrate its visual appearance on both web and mobile platforms.

\noindent \textbf{\ding{253} Introduction Page:}
\begin{figure*}[h]
    \centering
    \includegraphics[width=0.9\linewidth]{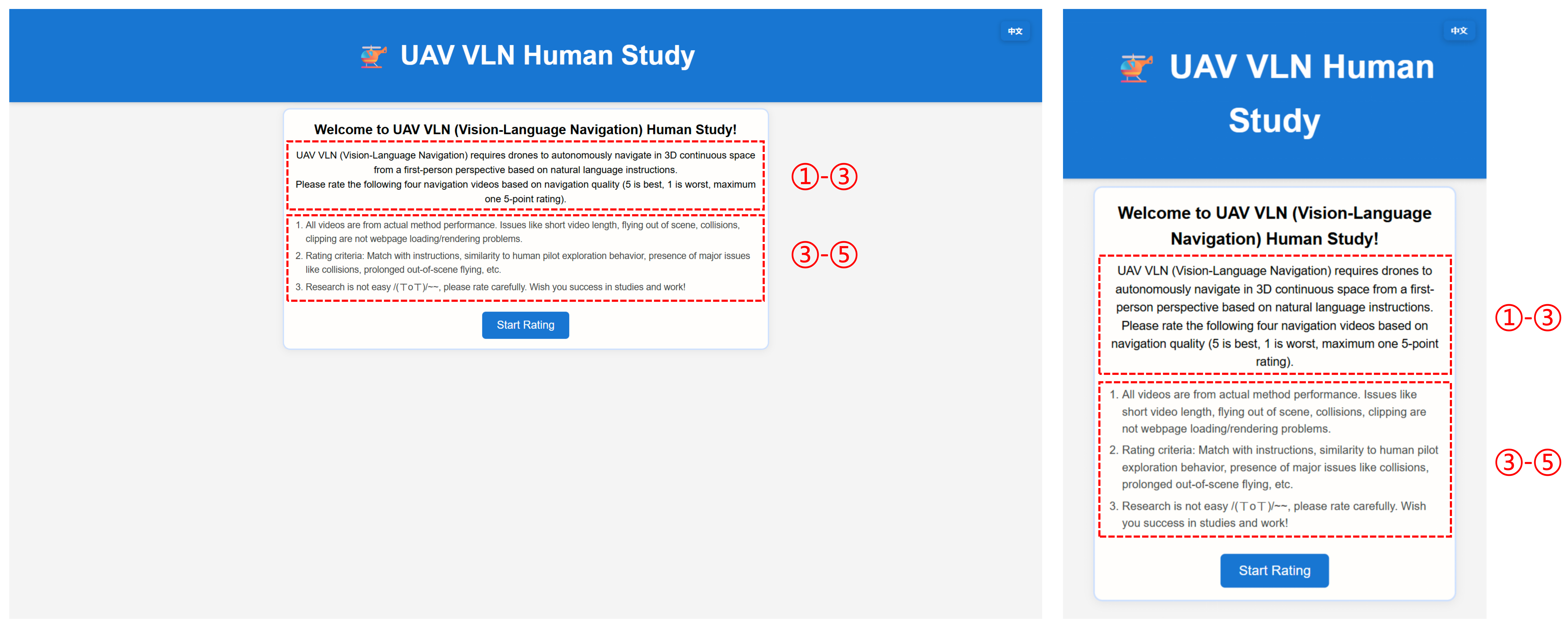}
    \caption{Introduction page of our questionnaire.}
    \label{fig:study_intro}
\end{figure*}

\noindent The introduction page, as shown in \cref{fig:study_intro}, provides an overview of the questionnaire, including:

\ding{172} \textbf{Task definition}: \textit{``UAV VLN (Drone Vision-Language Navigation) requires drones to autonomously navigate in a 3D continuous space from a first-person perspective based on natural language instructions.''};

\ding{173} \textbf{The participant’s task}: \textit{``Please rate the following five navigation videos based on their overall quality''};

\ding{174} \textbf{Scoring Criteria}: \textit{``5 being the best, 1 being the worst, with only one video allowed to receive a score of 5.''}, and \textit{``Reference scoring criteria: whether it matches the instructions, whether it resembles human pilot exploration behavior, whether there are critical issues like collisions, and whether it flies out of the scene for extended periods.''};

 \ding{175} \textbf{Brief Introduction of Trajectories in the Questionnaire}: \textit{``All videos reflect the actual performance of the methods. If issues such as short video length, flying out of the scene, collisions, or clipping occur, these are not due to webpage loading or rendering problems.''};

\ding{176} \textbf{Thank You Note to Participants}. ``Research is challenging/(ToT)/~~,so we hope everyone will rate carefully. We wish you academic success and a smooth career ahead!''.

\noindent \textbf{\ding{253} Evaluation Page:}
\begin{figure*}[h]
    \centering
    \includegraphics[width=0.9\linewidth]{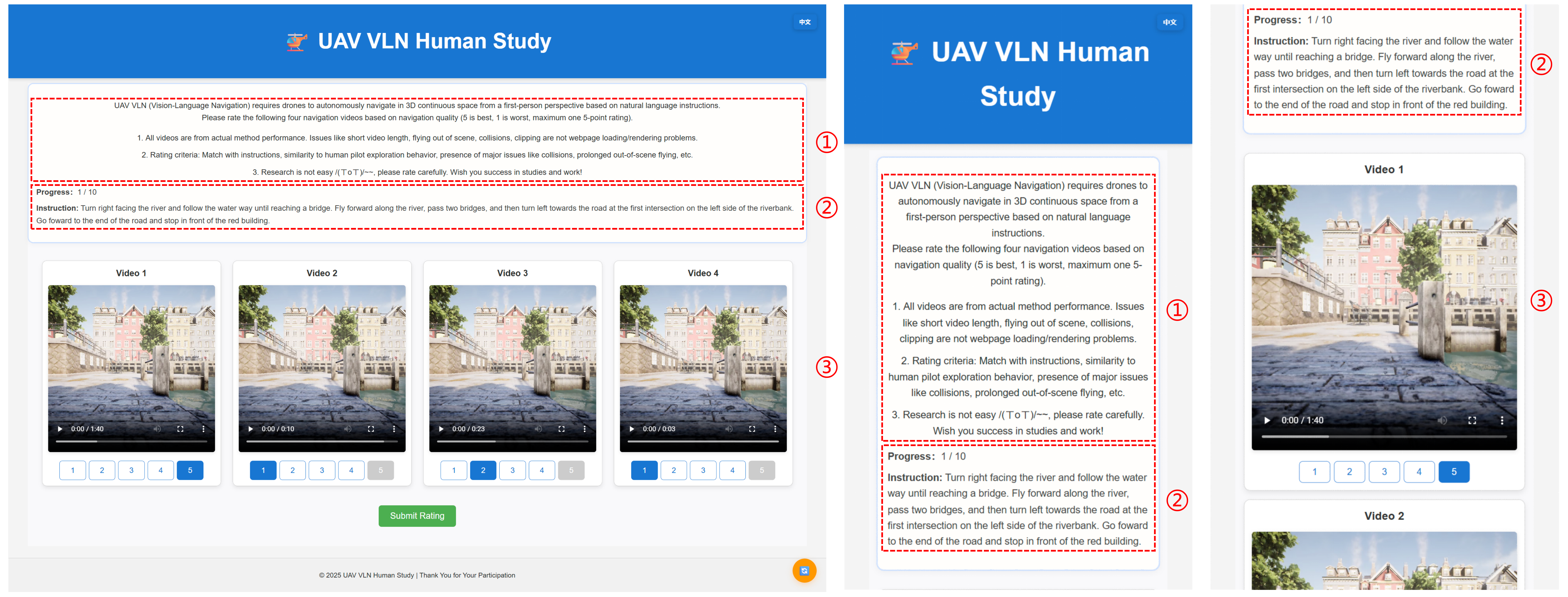}
    \caption{Evaluation page of our questionnaire.}
    \label{fig:study_eval}
\end{figure*}

\noindent As \cref{fig:study_eval}, the evaluation page also consists of three parts:

\ding{172} \textbf{Task Definition}, which is the same as Introduction Page;

\ding{173} \textbf{Current Progress and the Current Instruction}, e.g. \textit{``Review Progress:1/10. Instruction: Turn right facing the river and follow the water way until reaching a bridge. Fly forward along the river, pass two bridges, and then turn left towards the road at the first intersection on the left side of the riverbank. Go forward to the end of the road and stop in front of the red building.''};

\ding{174} \textbf{Video Display and Scoring}. Four embedded video players (Ours, BaseModel, NavGPT, and DiscussNav) are presented, each followed by a scoring interface that allows participants to assign a score from 1 (worst) to 5 (best), with the constraint that only one video per group may receive a score of 5.

\noindent \textbf{\ding{253} Farewell Page:}
\begin{figure*}[h]
    \centering
    \includegraphics[width=0.9\linewidth]{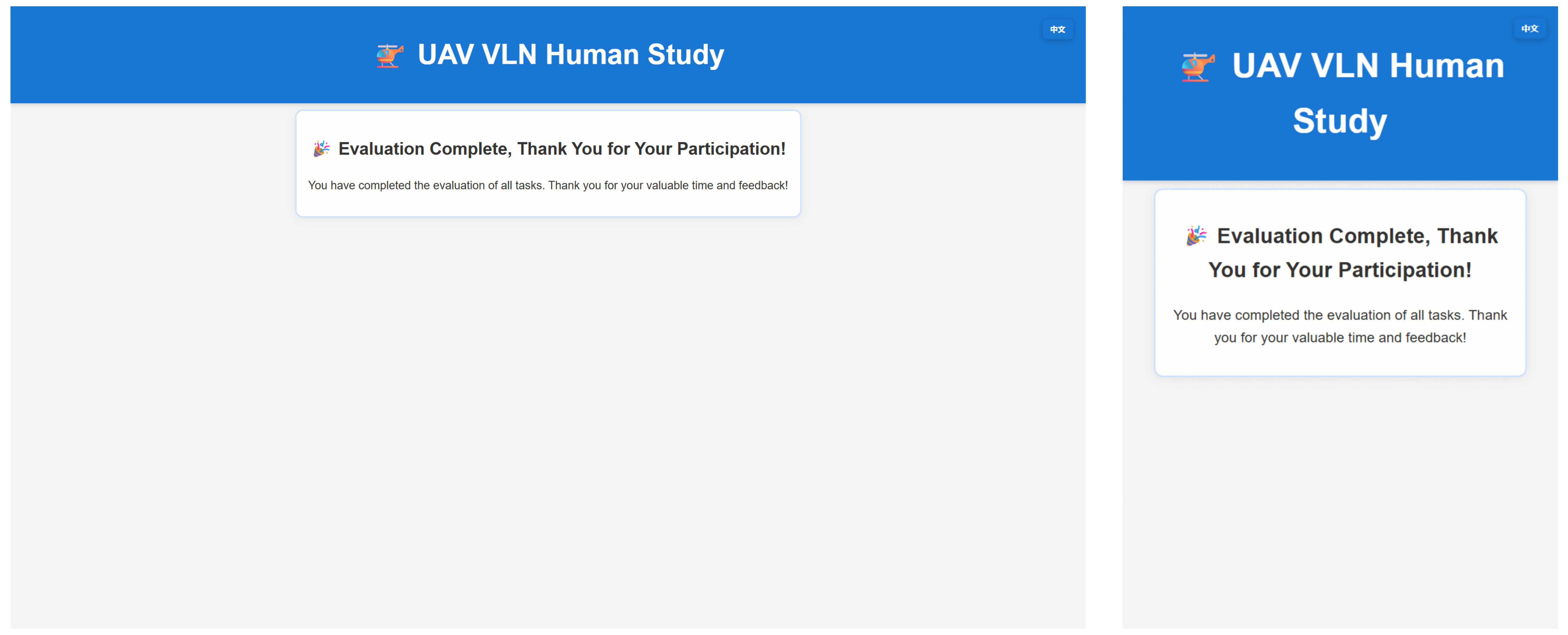}
    \caption{Farewell page of our questionnaire.}
    \label{fig:study_close}
\end{figure*}

\noindent As shown in \cref{fig:study_close}, the farewell page once again expresses our sincere gratitude to all participants for their time and contributions to the study. \textit{``Review Completed. Thank You for Your Participation! You have finished reviewing all tasks. Thank you for your valuable time and feedback!''}

\subsection{Analysis of Human Study Results}

\begin{figure*}[ht]
    \centering
    \includegraphics[width=\linewidth]{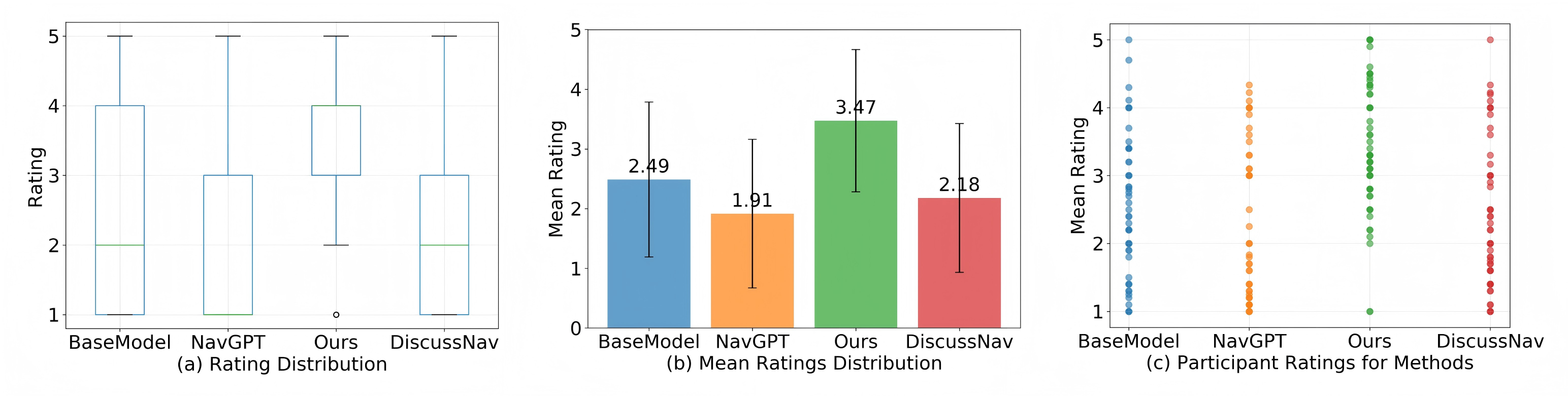}
    \caption{Human study results analysis: (a) Rating Distribution by Method (Box Plot), (b) Mean Ratings by Method (with Standard Deviation), (c) Participant Ratings for Different Methods.}
    \label{fig:human_study_analysis}
\end{figure*}

Our human study aimed to compare the performance of our proposed FineCog-Nav against three baseline methods: BaseModel (a prompt based on the STMR~\cite{STMR} action maker, covering basic instruction understanding, memory, and planning), NavGPT, and DiscussNav. A total of 69 participants contributed ratings across 10 trajectories per method, yielding 447 valid rating instances for analysis after removing duplicate responses.

\ding{172} \textbf{Performance Metrics:} The analysis of mean ratings showed that our method achieved the highest score at $3.474 \pm 1.192$, significantly outperforming BaseModel ($2.488 \pm 1.298$), DiscussNav ($2.179 \pm 1.247$), and NavGPT ($1.915 \pm 1.245$). Notably, our method received the highest proportion of top ratings—24.2\% of its ratings were 5 points—and a majority (52.4\%) were rated 4 or 5. In contrast, NavGPT received 58.8\% of its ratings as 1 point, reflecting poor perceived performance. This trend is further supported by the rating distribution box plot (\cref{fig:human_study_analysis}(a)), where our method exhibits a higher median (4.0) and substantially fewer low ratings compared to all baselines.

\ding{173} \textbf{Significance Testing:} Pairwise comparisons using the Wilcoxon signed-rank test revealed statistically significant differences between all method pairs (all p-values $< 0.0001$). Effect size analysis confirmed the practical relevance of these differences, with large effect sizes observed when comparing our method against BaseModel (Cohen's $d = -0.792$), NavGPT ($d = -1.280$), and DiscussNav ($d = 1.062$).

\ding{174} \textbf{Inter-method Correlation Analysis:} Correlation analysis indicated that participants’ ratings for the baseline methods were moderately to strongly correlated (e.g., BaseModel vs. DiscussNav: $r = 0.517$; NavGPT vs. DiscussNav: $r = 0.663$), suggesting consistent perception patterns among them. In contrast, our method showed very weak correlations with all baselines—BaseModel ($r = 0.057$), NavGPT ($r = 0.132$), and DiscussNav ($r = 0.175$)—highlighting its distinct and superior perceived behavior.

\ding{175} \textbf{Participant Rating Consistency:} Despite notable inter-participant variability in rating tendencies—as illustrated in the participant ratings heatmap (\cref{fig:human_study_analysis}(c))—our method consistently received a higher fraction of favorable ratings (4s and 5s) across nearly all participants compared to the baselines.

Overall, the findings from this human study, comprehensively presented in \cref{fig:human_study_analysis}, provide robust evidence for the effectiveness of our proposed FineCog-Nav over existing baselines.

\section{More Evaluations}

\subsection{Controlled Evaluation}

See, Point, Fly (SPF)\cite{hu2025see} is a zero-shot UAV-VLN framework, which shows high success rates in their paper. However, their evaluation is limited to self-constructed tasks that are nearly saturated (success rates of 90–100\%), offering little room for meaningful comparison. In contrast, our benchmark encompasses a broader spectrum of scenarios, ranging from relatively easy instances to challenging cases. 

\begin{figure}[h]
    \centering
    \includegraphics[width=0.8\linewidth]{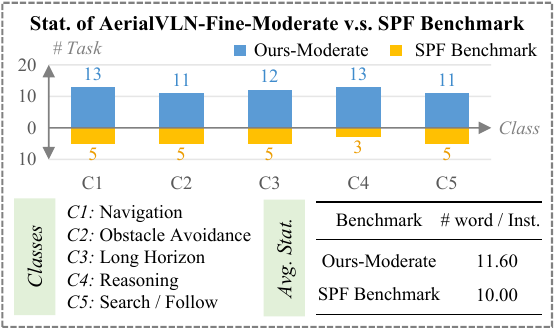}
    \caption{Statistics of AerialVLN-Fine-Moderate}
    \label{fig:spf_stat}
\end{figure}

To ensure a fair comparison, we curate a subset of our dataset—AerialVLN-Fine-Moderate—that closely aligns with the task design of SPF. The statistics of AerialVLN-Fine-Moderate and the SPF benchmark are presented in \cref{fig:spf_stat}. As shown in \cref{tab:spf}, on this comparable subset, our method achieves a success rate (SR) of 53.8\%, significantly outperforming SPF’s 30.8\%, thereby demonstrating that the lower SR observed in our full benchmark primarily stems from increased task difficulty rather than model limitations.

\begin{table}[h]
    \caption{Performance on AerialVLN-Fine-Moderate}
	\centering
    \setlength{\tabcolsep}{3pt}
    \label{tab:spf}
    \begin{tabular}{lcccccc}
    \toprule
    Method & C1 & C2 & C3 & C4 & C5 & All \\
    \midrule
    \textbf{S}ee,\textbf{P}oint,\textbf{F}ly & 30.8 & 9.1 & 16.7 & 15.4 & 9.1 & 16.7 \\
    BaseModel & 7.7 & 9.1 & 8.3 & 7.7 & 9.1 & 8.3 \\
    DiscussNav & 7.7 & 9.1 & 25.0 & 23.1 & 18.2 & 16.7 \\
    NavGPT & 15.4 & 0.0 & 16.7 & 7.7 & 0.0 & 8.3 \\
    \rowcolor{cyan!10}
    FineCog-Nav & 53.8 & 18.2 & 33.3 & 15.4 & 18.2 & 28.3 \\
    \bottomrule
    \end{tabular}
\end{table}

\subsection{Generalization Evaluation}

To further demonstrate the generalization capability of our method across diverse benchmarks, we evaluate it on the OpenFly dataset\cite{openfly}, specifically using the AirSim16 scene, the results are shown in \cref{tab:openfly}. Our approach matches the performance of the trained one and surpasses SPF~\cite{hu2025see}(the zero-shot UAV-VLN baseline), highlighting its generalization ability.

\begin{table}[h]
    \caption{Performance on OpenFly(AirSim 16 subset)}
	\centering
    \setlength{\tabcolsep}{3pt}
    \label{tab:openfly}
	\begin{tabular}{lcrrrr}
	\toprule
    Method & Trained & SR/\% & OSR/\% & SPL/\% & NE/m \\
    \midrule
    OpenFly & \ding{51} & 10.6 & 57.9 & 8.3 & 178.9 \\
    \midrule
    \textbf{S}ee,\textbf{P}oint,\textbf{F}ly & \ding{56} & 3.5 & 5.4 & 2.6 & 88.2 \\
    \rowcolor{cyan!10}
    FineCog-Nav & \ding{56} & 9.9 & 32.7 & 9.1 & 80.8 \\
    \bottomrule
	\end{tabular}
\end{table}

\subsection{Efficiency Evaluation}

\begin{table}[t]
    \caption{Cost Analysis of FineCog-Nav and Baselines on AerialVLN-Fine.}
	\centering
    \renewcommand\arraystretch{0.85}
    \renewcommand{\tabcolsep}{2pt}
    \setlength{\abovecaptionskip}{-0.2cm}
    \setlength{\belowcaptionskip}{-0.2cm}
    \label{tab:efficiency}
	\begin{tabular}{lr|rrr}
	\toprule
    Method & SR/\% & Calls & Tokens & Latency/s \\
    \midrule
    BaseModel & 2.7 & 2.0 & 2.6K & 18.8 \\
    NavGPT & 0.0 & 3.1 & 10.7K & 116.4 \\
    DiscussNav & 2.7 & 6.2 & 18.7K & 80.5 \\
    \rowcolor{cyan!10}
    FineCog-Nav & 6.0 & 4.9 & 3.9K & 23.5 \\
    \bottomrule
	\end{tabular}
\end{table}

We compare the computational efficiency of FineCog-Nav with multiple baselines on AerialVLN-Fine (\cref{tab:efficiency}). We report the average amount of API calls, token consumption per step, and latency per step.

While achieving the highest success rate (6.0\%), FineCog-Nav maintains a favorable efficiency–performance balance. Compared with two framework-based methods (DiscussNav and NavGPT), FineCog-Nav significantly reduces computational overhead, e.g., token usage (3.9K v.s. 10.7K/18.7K) and latency (23.5s v.s. 116.4s/80.5s), while achieving better performance. Compared with the BaseModel, although incurring modest additional cost, FineCog-Nav improves the success rate by more than 2×.

Overall, FineCog-Nav achieves a better balance between reasoning capability and computational efficiency.

\subsection{Ablation of Collision Warning}

\begin{table}[t]
    \caption{Ablation Analysis of Collision Warning Module on AerialVLN-Fine.}
	\centering
    \scriptsize
    \renewcommand\arraystretch{0.9}
    \renewcommand{\tabcolsep}{1.5pt}
    \setlength{\abovecaptionskip}{-0.2cm}
    \setlength{\belowcaptionskip}{-0.2cm}
    \label{tab:ablation_collision}
	\begin{tabular}{lrrrr|rrrr}
	\toprule
    \multirow{2}{*}{Method} & \multicolumn{4}{c}{w/ collision warning} & \multicolumn{4}{c}{w/o collision warning} \\
    \cline{2-9}
    \\[-1.9ex] 
    & SR/\% & OSR/\% & NE/m & nDTW/\% & SR/\% & OSR/\% & NE/m & nDTW/\% \\
    \midrule
    BaseModel & 2.7 & 6.3 & 270.1 & 10.7 & 1.0 & 3.7 & 317.1 & 13.4 \\
    NavGPT & 0.0 & 0.7 & 110.9 & 15.9 & 0.3 & 0.7 & 122.4 & 14.3 \\
    DiscussNav & 2.7 & 3.3 & 98.4 & 19.6 & 1.7 & 1.7 & 111.1 & 17.3 \\
    \rowcolor{cyan!10} 
    FineCog-Nav & 6.0 & 9.0 & 91.4 & 22.8& 2.7 & 5.3 & 100.4 & 19.2 \\
    \bottomrule
	\end{tabular}
\end{table}
We evaluated all methods w/ and w/o the collision warning ($W_t$) to quantify its impact, shown in \cref{tab:ablation_collision}.
Removing the collision warning module leads to consistent performance drops across methods, especially in SR and OSR, indicating that it effectively reduces collision-induced failures and improves navigation success. At the same time, FineCog-Nav outperforms all baselines in both settings, with consistent relative gains. This suggests that the improvements of our method do not rely on this module, but stem from its stronger reasoning and decision-making capabilities.

\definecolor{promptorange}{RGB}{255, 248, 220}
\definecolor{promptwhite}{RGB}{254,248,241}
\definecolor{promptpurple}{RGB}{152,125,154}
\definecolor{promptgolden}{RGB}{255, 185, 15}
\definecolor{promptred}{RGB}{205, 38, 38}

\definecolor{instruct}{RGB}{255,225,225}
\definecolor{instructback}{RGB}{255,247,247}
\definecolor{subgoal}{RGB}{255,193,196}
\definecolor{subgoalback}{RGB}{255,247,247}
\definecolor{percpt}{RGB}{255,247,200}
\definecolor{percptback}{RGB}{255,251,228}
\definecolor{atten}{RGB}{255,234,139}
\definecolor{attenback}{RGB}{255,251,228}
\definecolor{imagin}{RGB}{192,238,228}
\definecolor{imaginback}{RGB}{230,243,231}
\definecolor{subjudger}{RGB}{193,231,178}
\definecolor{subjudgerback}{RGB}{230,243,231}
\definecolor{memory}{RGB}{210,205,253}
\definecolor{memoryback}{RGB}{241,235,247}
\definecolor{decision}{RGB}{165,215,255}
\definecolor{decisionback}{RGB}{232,245,255}

\clearpage
\newpage
\clearpage

\section{Detailed Prompts for BaseModel}
In designing the Basemodel, we retained only the most basic and minimal prompt to minimize performance gains from prompt engineering. Specifically, the Basemodel consists of two modules: Perception, responsible for scene understanding, and Action, responsible for decision-making.

\subsection{Perception}

\begin{tcolorbox}[colback = attenback, colframe = atten, fonttitle = \bfseries, colbacktitle = atten, coltitle=white, title=Perception]
\centering
\centering{
\begin{minipage}{0.98\textwidth}
Describe the image.
\end{minipage}
}
\end{tcolorbox}

We present a demonstration of the perception output generated by Qwen2.5-VL-32B-Instruct for an arbitrary input image.

\begin{tcolorbox}[colback = attenback, colframe = atten, fonttitle = \bfseries, colbacktitle = atten, coltitle=white, title=Perception]
\centering
\textbf{Input Information}
\begin{center}
\includegraphics[width=0.45\linewidth]{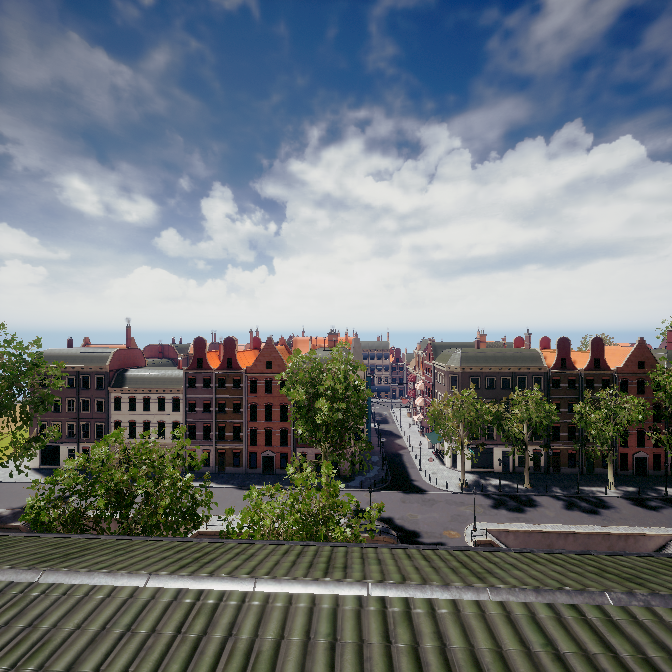}
\label{fig:1}
\end{center}
\centering{
\begin{minipage}{0.98\textwidth}
Describe the image.
\end{minipage}
}
\centering
\textbf{Output Information}\\

\centering{
\begin{minipage}{0.98\textwidth}
The image depicts a picturesque urban scene with a European architectural style. Here are the key details:\\

\#\#\# **Foreground:**\\
- The foreground shows a portion of a roof, likely from a building or structure in the viewer's immediate vicinity. The roof is made of dark, textured tiles, suggesting a traditional or historical design.\\

\#\#\# **Middle Ground:**\\
- The middle ground features a street that runs diagonally through the scene, leading the viewer's eye toward the background. The street is relatively empty, with no visible vehicles or people. \\
- On either side of the street, there are rows of buildings with classic European architecture. The buildings are primarily made of red brick, with 
\end{minipage}
}
\end{tcolorbox}
\begin{tcolorbox}[colback = attenback, colframe = atten, fonttitle = \bfseries, colbacktitle = atten, coltitle=white, title=Perception]
\centering{
\begin{minipage}{0.98\textwidth}
pitched roofs and chimneys, giving them a uniform and historic appearance.
- The buildings have multiple stories, with large windows and symmetrical facades. Some of the buildings have ornate details, such as decorative trim or small balconies, adding to their charm.\\
- Trees are planted along the sidewalks, providing greenery and shade. The trees are lush and well-maintained, contributing to the overall aesthetic of the scene.\\

\#\#\# **Background:**\\
- The background shows a clear blue sky with scattered white clouds, indicating a bright and sunny day.\\
- In the distance, the horizon is visible, suggesting that the city is near a body of water or an open landscape, though the specific details are not clear.\\

\#\#\# **Overall Atmosphere:**\\
- The image conveys a calm and serene urban environment, with a focus on historical architecture and well-maintained greenery. The lighting is bright and natural, enhancing the colors and details of the scene.\\
- The perspective and composition create a sense of depth, drawing the viewer into the street and the buildings beyond.\\

This image likely represents a digitally rendered or computer-generated scene, given the clean lines, lighting, and overall aesthetic quality. It could be from a video game, architectural visualization, or a digital art project. 
\end{minipage}
}
\end{tcolorbox}

\subsubsection{Rule-based Collision Estimator}

The rule-based collision estimator is shared by all baselines and FineCog-Nav. Considering that collisions are rare during TURN LEFT or TURN RIGHT, and that MOVE LEFT and MOVE RIGHT are also infrequently used, we estimate collision risk only for MOVE FORWARD, ASCENT, and DESCENT.

To estimate the risk of MOVE FORWARD, we examine the central region of the depth map.
Because the camera’s field of view is limited, depth values above and below the image are not directly observable. Estimating the risk of ASCENT and DESCENT therefore requires an additional assumption. By treating the unseen upper and lower regions as planar surfaces, we infer their distances using the vertical height gradient within the visible area.\\

\lstset{
language=python,
frame=single,
breaklines=true,
tabsize=2,
basicstyle=\ttfamily,
commentstyle= \color{black!20!blue!50},
columns=fullflexible,
}
\begin{lstlisting}
# Check whether there is collision risk in performing the action
def _check_collision(self, depth_img, action, img_width=672, img_height=672, drone_width=1.0, drone_height=0.1, fov=90, distance=5.1):
	# depth_img.shape # (h, w, c)
	pixel_angle = fov / img_width
	# the coordinates of center pixel
	center_x = img_width // 2
	center_y = img_height // 2
	# angle of deviation of the edge of the central region from center
	half_angle_x=np.arctan(drone_width/(2*distance))*(180/np.pi)
	half_angle_y=np.arctan(drone_height/(2*distance))*(180/np.pi)
	# size of the central region
	half_width = math.ceil(half_angle_x / pixel_angle)
	half_height = math.ceil(half_angle_y / pixel_angle)
	if action == 1: # MOVE_FORWARD
		for dx in range(-half_width, half_width):
			for dy in range(-half_height, half_height):
				x = center_x + dx
				y = center_y + dy
				if x<0 or x>=img_width or y<0 or y>=img_height:
					continue
				if depth_img[y, x] < distance:
					# depth less than threshold -> risky
					return True
		return False
	elif action == 4: # ASCENT
		# Calculate the height of each pixel
		height_map=np.zeros_like(depth_img)
		for y in range(img_height):
			angle_y_tan=np.tan(abs(y-center_y)*pixel_angle*(np.pi/180))
			height_map[y]=angle_y_tan*depth_img[y]
		height=math.ceil(img_height*0.05)
		# Calculate the gradient in the vertical direction
		gradient_y = np.gradient(height_map, axis=0)
		gradient_threshold = 0.02
		for dx in range(-half_width, half_width):
			x = center_x + dx
			for dy in range(0, height):
				y = img_height + dy
				if x<0 or x>=img_width or y<0 or y>=img_height:
					continue
				gradient = abs(gradient_y[y, x])
				if height_map[y,x]<distance and gradient<=gradient_threshold:
					# height and gradient less than threshold -> risky
					return True
		return False
	elif action == 5: # DESCENT
		# Same as ASCENT
	else:
		return False
def check_collision(self, depth_img):
	risks = ""
	if _check_collision(depth_img, 1):
		risks += "MOVE_FORWARD will collide with objects. "
	if _check_collision(depth_img, 4, distance=2.2):
		risks += "ASCEND will collide with objects. "
	if _check_collision(depth_img, 5, distance=2.2):
		risks += "DESCEND will collide with objects. "
	if len(risks) == 0:
		risks = "None"
	return risks
\end{lstlisting}

\subsection{Action}

The Action module, following current modular design trends, incorporates fundamental components for instruction understanding, memory management, and long-term planning, operating with standardized input-output formats. 
\begin{tcolorbox}[colback = decisionback, colframe = decision, fonttitle = \bfseries, colbacktitle = decision, coltitle=white, title=Action]
\centering{
\begin{minipage}{0.98\textwidth}
{[Task Description]} \\
You are an embodied unmanned aerial vehicle that navigates in the real world. You need to explore between some places marked and ultimately find the destination to stop. \\
\\
{[Input Format]}\\
`Instruction' is a global, step-by-step detailed guidance.
\end{minipage}
}
\end{tcolorbox}
\begin{tcolorbox}[colback = decisionback, colframe = decision, fonttitle = \bfseries, colbacktitle = decision, coltitle=white, title=Action]
\centering{
\begin{minipage}{0.98\textwidth}
`Observation': the description of the input image.\\
`History' is your previously executed actions.\\
`Plan' records previous long-term multi-step planning info that you can refer to now. \\
`Collision Warning' is the tips to avoid collisions and infer your relative position with surroundings.\\
\\
{[Output Format]} \\
```json\\
\{\{\\
    ``Thought'': ``your thoughts about the task, which may include your comprehension, surroundings, history, and etc.'',\\
    ``Plan'': ``your updated plan.'',\\
    ``Action\_Number'': ``only your next one action number.''\\
\}\}\\
'''\\

You must follow the Output Format to output, and you must not output any other words. Output with plain text format. Think step by step. First, judge by the input image, give a first `Thought', and depict your orientation. Second, check that if a landmark in the current `plan' is within 5 meters of your current position, then based on `Instruction' and the previous `Plan', update your multi-step `Plan'. Each plan needs to follow a state word (TODO, In Process, Completed). Finally, judge by the input image again, and give the only one specific `action number' according to the [Valid Action List]. \\
\\
{[Valid Action List]} \\
0: TASK\_FINISH\\
1: MOVE\_FORWARD (5 meters)\\
2: TURN\_LEFT (15 degrees)\\
3: TURN\_RIGHT (15 degrees)\\
4: ASCENT (2 meters)\\
5: DESCENT (2 meters) \\
6: MOVE\_LEFT (5 meters) \\
7: MOVE\_RIGHT (5 meters) \\
\\
{[Input]}\\
Instruction: \{instruction\}\\
Observation: \{observation\}\\
History: \{history\}\\
Plan: \{plan\}\\
Collision Warning: \{warning\}
\end{minipage}
}
\end{tcolorbox}

\newpage

Here, we further demonstrate how the action module processes the output from the perception module. After obtaining a scene description from the perception module, the action module parses this information, integrates it with the current navigation context (e.g., instruction history, UAV state), and performs reasoning to determine the next executable action. We use Qwen2.5-72B-Instruct \cite{qwen2.5} as the LLM.
\begin{tcolorbox}[colback = decisionback, colframe = decision, fonttitle = \bfseries, colbacktitle = decision, coltitle=white, title=Action]
\centering
\textbf{Input Information}
\centering{
\begin{minipage}{0.98\textwidth}
{[Input]}\\
Instruction: Reduce to the second floor height, then turn right to face the street, and then proceed straight ahead to the brown wooden door of the light red building at the corner of the road. Turn your back and face the road, then fly straight forward to the intersection beside the bridge. Turn right, face the white building at the end of the bridge and the road, and then fly straight along the road. Turn right and face the tree at the end of the road, then fly forward to the road intersection. Turn right and head towards the sidewalk on the left side of the building with the dog mural. Walk forward to the building with the dog mural, then turn left to face the red building with the green door on the other side of the river, and then fly over the river, then turn right to face the building with the green roof and red walls, then fly forward to the red building and land there.\\
\\
Observation: The image depicts a picturesque urban scene with a European architectural style. Here are the key details: \\
\\
\#\#\# **Foreground:**\\
- The foreground shows a portion of a roof, likely from a building or structure in the viewer's immediate vicinity. The roof is made of dark, textured tiles, suggesting a traditional or historical design. \\
\\
\#\#\# **Middle Ground:**\\
- The middle ground features a street that runs diagonally through the scene, leading the viewer's eye toward the background. The street is relatively empty, with no visible vehicles or people. \\
- On either side of the street, there are rows of buildings with classic European architecture. The buildings are primarily made of red brick, with pitched roofs and chimneys, giving them a uniform and historic appearance. 
\end{minipage}
}
\end{tcolorbox}
\begin{tcolorbox}[colback = decisionback, colframe = decision, fonttitle = \bfseries, colbacktitle = decision, coltitle=white, title=Action]
\centering{
\begin{minipage}{0.98\textwidth}
- The buildings have multiple stories, with large windows and symmetrical facades. Some of the buildings have ornate details, such as decorative trim or small balconies, adding to their charm. \\
- Trees are planted along the sidewalks, providing greenery and shade. The trees are lush and well-maintained, contributing to the overall aesthetic of the scene.\\

\#\#\# **Background:**\\
- The background shows a clear blue sky with scattered white clouds, indicating a bright and sunny day.\\
- In the distance, the horizon is visible, suggesting that the city is near a body of water or an open landscape, though the specific details are not clear.\\

\#\#\# **Overall Atmosphere:**\\
- The image conveys a calm and serene urban environment, with a focus on historical architecture and well-maintained greenery. The lighting is bright and natural, enhancing the colors and details of the scene.\\
- The perspective and composition create a sense of depth, drawing the viewer into the street and the buildings beyond.\\

This image likely represents a digitally rendered or computer-generated scene, given the clean lines, lighting, and overall aesthetic quality. It could be from a video game, architectural visualization, or a digital art project.\\

History: [`DESCENT (2 meters)', `ASCENT (2 meters)'] \\

Plan: Reduce to the second floor height (In Process), Turn right to face the street (TODO), Proceed straight ahead to the brown wooden door of the light red building at the corner of the road (TODO), Turn back and face the road (TODO), Fly straight forward to the intersection beside the bridge (TODO), Turn right and face the white building at the end of the bridge and the road (TODO), Fly straight along the road (TODO), Turn right and face the tree at the end of the road (TODO), Fly forward to the road intersection (TODO), Turn right and head towards the sidewalk on the left side of the building with the dog mural (TODO), Walk forward to the building with the dog mural (TODO), Turn left to face the red building with the green door on the other side of the 
\end{minipage}
}
\end{tcolorbox}

\begin{tcolorbox}[colback = decisionback, colframe = decision, fonttitle = \bfseries, colbacktitle = decision, coltitle=white, title=Action]
\centering{
\begin{minipage}{0.98\textwidth}river (TODO), Fly over the river (TODO), Turn right to face the building with the green roof and red walls (TODO), Fly forward to the red building and land there (TODO). \\
Collision Warning: None
\end{minipage}
}
\centering\textbf{Output Information}
\centering{
\begin{minipage}{0.98\textwidth}
```json\\
\{\\
    ``Thought": ``I am currently at the height of the second floor, as I have just completed an ascent and descent. The image shows a European-style urban scene with a street running diagonally through the middle ground. The street is lined with red brick buildings and trees. I need to turn right to face the street and proceed to the brown wooden door of the light red building at the corner of the road.", \\
    ``Plan": ``Reduce to the second floor height (Completed), Turn right to face the street (In Process), Proceed straight ahead to the brown wooden door of the light red building at the corner of the road (TODO), Turn back and face the road (TODO), Fly straight forward to the intersection beside the bridge (TODO), Turn right and face the white building at the end of the bridge and the road (TODO), Fly straight along the road (TODO), Turn right and face the tree at the end of the road (TODO), Fly forward to the road intersection (TODO), Turn right and head towards the sidewalk on the left side of the building with the dog mural (TODO), Walk forward to the building with the dog mural (TODO), Turn left to face the red building with the green door on the other side of the river (TODO), Fly over the river (TODO), Turn right to face the building with the green roof and red walls (TODO), Fly forward to the red building and land there (TODO).", \\
    ``Action\_Number": 3 \\
\}\\
```
\end{minipage}
}
\end{tcolorbox}

\section{Detailed Prompts for Framework Baselines}

Since most existing zero-shot methods are designed for discrete ground-level navigation, we made specific adjustments to each method to adapt them to the aerial VLN setting. These modifications mainly include:
\begin{itemize}
    \item [1.] Replacing all original outputs designed to select a \textit{waypoint} with \textit{action} selection from a predefined set of UAV-executable commands, including ascend, descend, turn left, turn right, move left, and move right.
    \item[2.] Replacing all terms like \textit{indoor} and \textit{vehicle} in the prompts with \textit{outdoor} and \textit{UAV}, respectively.
\end{itemize}
The modified versions of the baseline codes will be released as open-source. We provide the specific modifications made for each baseline method. All deleted content from the original prompts is highlighted in light gray, while the added modifications are marked in red. The details are as follows.

\subsection{NavGPT}
NavGPT\cite{zhou2024navgpt2unleashingnavigationalreasoning} originally uses BLIP-2 \cite{Li2023BLIP2BL} and Faster R-CNN \cite{Girshick2015FastR} to generate scene descriptions, GPT-4 \cite{Achiam2023GPT4TR} for navigation reasoning, and GPT-3.5-turbo for history management. In our adaptation, we replace GPT-4 with the LLM defined in our setup, BLIP-2 with our selected VLM, and convert its discrete waypoint output to UAV-compatible action outputs. We want to highlight that NavGPT does not use standalone prompts to obtain scene information; instead, it relies on pre-generated JSON files containing visual observations of the environment. To align it with our unified prompt-based framework, We adopt a basic scene understanding module, which offers basic yet clearer detail description compared to the original JSON format. We randomly select one scene from the original R2R Json file and our AerialVLN-Fine dataset to illustrate the prompt here.

\begin{tcolorbox}[colback=promptwhite, colframe=lightgray, boxrule=0mm, arc=1mm,colbacktitle=promptpurple]
\textbf{Original JSON File}\\
A series of hallways, rooms, and bathrooms with wooden floors, white walls, and various light fixtures and furniture.
\tcblower
\textbf{Generated Scene Description}\\
The image depicts an outdoor scene with a relatively simple and open layout. Here is a detailed description of the scene:\\
\\
\#\#\# **Objects and Layout:**\\
1. **Foreground:**\\
   - The ground is a flat, smooth, concrete surface, likely a rooftop or a paved area.\\
   - There is a small, cylindrical object (possibly a barrel or a container) placed near the center-right of the foreground. It is dark in color and appears to be made of metal or a similar material.\\
2. **Left Side:**\\
   - A tall, red brick wall runs vertically along the left side of the scene. The wall is solid and appears to be part of a building or structure.\\
   - The wall extends from the foreground into the background, creating a boundary on the left side of the scene.\\
3. **Right Side:**\\
   - On the right side, there is a partial view of another structure or wall, which appears to be made of a different material (possibly metal or concrete) and is 
\end{tcolorbox}

\begin{tcolorbox}[colback=promptwhite, colframe=lightgray, boxrule=0mm, arc=1mm,colbacktitle=promptpurple]
painted in a dark color.\\
   - The right side is less prominent compared to the left, and the structure is not fully visible.\\
4. **Background:**\\
   - Beyond the wall on the left, there is a tree with green foliage. The tree is relatively large and occupies a significant portion of the background on the right side.\\
   - In the far background, there is a power line structure (a tall electrical tower) visible, with multiple power lines running horizontally across the scene. The lines stretch from the left side of the image toward the right.\\
   - The sky is clear with a few scattered clouds, indicating a daytime scene with good visibility.\\
\\
\#\#\# **Relative Positions:**\\
- The cylindrical object is positioned closer to the right side of the scene and slightly in front of the viewer.\\
- The red brick wall is on the left, running vertically and parallel to the edge of the scene.\\
- The tree is located behind the wall, slightly to the right of the center.\\
- The power lines and tower are in the far background, extending horizontally across the scene. \\
\#\#\# **Possible Directions for Navigation:**\\
1. **Forward (Straight Ahead):**\\
   - Moving forward would take you along the concrete surface, away from the viewer's perspective. The path appears clear, but the cylindrical object might be an obstacle if you need to navigate around it.\\
2. **Left:**\\
   - Moving left would bring you closer to the red brick wall. The wall is a solid barrier, so navigating to the left would require staying close to the wall or finding a way around it. \\
3. **Right:**\\
   - Moving right would take you toward the dark structure on the right side. The area appears open, but the partial structure might limit movement in that direction.\\
4. **Backward:**\\
   - Moving backward would take you away from the scene, back toward the origin point of the viewer.\\
\#\#\# **Obstacles:**\\
- The **cylindrical object** in the foreground could be an obstacle if precise navigation is required.\\
- The **red brick wall** on the left side is a solid barrier that would need to be navigated around if moving left is necessary. \\
- The **power lines** in the background are high 
\end{tcolorbox}
\begin{tcolorbox}[colback=promptwhite, colframe=lightgray, boxrule=0mm, arc=1mm,colbacktitle=promptpurple]
and do not pose a direct obstacle, but they are a notable feature in the scene.\\
\#\#\# **Landmarks:**\\
- The **red brick wall** is a prominent landmark on the left side, providing a clear boundary.\\
- The **tree** in the background is a natural landmark, helping to orient the scene.\\
- The **power lines and tower** are a distinct feature in the background, useful for navigation or orientation.\\

\#\#\# **Overall Impression:**\\
The scene is relatively open and flat, with clear boundaries provided by the wall and the tree. The cylindrical object and the wall are the primary obstacles, while the power lines and tree serve as notable landmarks. The concrete surface suggests a rooftop or an open area, and the overall layout is simple and navigable with some attention to the obstacles.
\end{tcolorbox}

For details, please refer to the prompt design details. \par
\begin{tcolorbox}[colback=promptwhite, colframe=lightgray, boxrule=0mm, arc=1mm,colbacktitle=promptpurple, fonttitle=\Large, coltitle=white, title=\textbf{NavGPT Prompt}, toptitle = 1mm, bottomtitle = 0.5mm]
\centering
\textbf{Instruction Split Prompt}\\
\centering{
\begin{minipage}{0.98\textwidth}
Given the long instruction: \{instruction\}\\
Divide the long instruction into action steps with detailed descriptions in the following format: \\
Action plan:\\
1. action$\_$step$\_$1\\
2. action$\_$step$\_$2\\
...\\
Action plan:
\end{minipage}
}
\centering{
\textbf{Get Scene Prompt}
\begin{minipage}{0.98\textwidth}
\textcolor{promptred}{
Please describe the current outdoor scene as detailed as possible, including the objects you can see, their relative positions, the layout of the scene, and the possible directions that can be navigated. Also, point out any obstacles and landmarks that might be helpful for navigation.}
\end{minipage}
}
\textbf{Action Make Prompt}\\
\centering{
\begin{minipage}{0.98\textwidth}
You are an agent following an action plan to navigation in \textcolor{lightgray}{indoor} \textcolor{promptred}{outdoor} environment.\\
Action plan: \{action$\_$plan\}\\
\textcolor{promptred}{
Action list: [`TASK$\_$FINISH', `MOVE$\_$FORW ARD (5 meters)', `TURN$\_$LEFT (15 degrees)', `TURN$\_$RIGHT (15 degrees)', `ASCENT (2 meters)', `DESCENT (2 meters)', `MOVE$\_$LEFT (5 meters)', `MOVE$\_$RIGHT (5 meters)']}
\end{minipage}
}
\end{tcolorbox}
\begin{tcolorbox}[colback=promptwhite, colframe=lightgray, boxrule=0mm, arc=1mm,colbacktitle=promptpurple, fonttitle=\Large, coltitle=white, title=\textbf{NavGPT Prompt}, toptitle = 1mm, bottomtitle = 0.5mm]
\centering
\centering{
\begin{minipage}{0.98\textwidth}
\textcolor{promptred}{
You are currently at one of the steps in the plan. You will be given the history of previous steps you have taken, the current observation of the \textcolor{lightgray}{environment, and the navigable viewpoints for the next step} \textcolor{promptred}{current position}.
}
You should:\\
1) evaluate the history and observation to decide which step of action plan you are at.\\
2) choose \textcolor{lightgray}{one viewpoint from the navigable viewpoints} \textcolor{promptred}{the next action from the action list}.\\
\textcolor{lightgray}{Each navigable viewpoint has a unique ID,} you should only answer the \textcolor{lightgray}{ID} \textcolor{promptred}{action} in the Final Answer.\\
----\\
Starting below, you should strictly follow this format:\\
History: the history of previous steps you have taken\\
Observation: the current observation of the \textcolor{lightgray}{environment} \textcolor{promptred}{current position}\\
\textcolor{lightgray}{Navigable viewpoints: the navigable viewpoints for the next step}\\
Thought: your thought on the next step\\
Collision Warning: Tips to avoid collisions and infer your relative position with surroundings\\
Final Answer: \textcolor{lightgray}{`viewpointID'} \textcolor{promptred}{`action'} \\
----\\
Begin!\\
History: \{history\}\\
Observation: \{observation\}\\
\textcolor{lightgray}{Navigable viewpoints: \{navigable$\_$viewpoints\}}\\
\textcolor{promptred}{Collision Warning: \{warning\}}\\
Thought:
\end{minipage}
}
\textbf{History Manage Prompt}\\
\centering{
\begin{minipage}{0.98\textwidth}
You are an agent navigating in \textcolor{lightgray}{indoor} \textcolor{promptred}{an outdoor} environment.\\
You have reached a new \textcolor{lightgray}{viewpoint} \textcolor{promptred}{position} after taking the previous action. You will be given the navigation history, the current observation of the \textcolor{lightgray}{environment} \textcolor{promptred}{current position}, and the previous action you took. \\
You should:\\
1) Evaluate the new observation and history.\\
2) Update the history with the previous action and the new observation.\\
History: \{history\}\\
Previous action: \{previous\_action\}\\
Observation: \{observation\}\\
Update history with the new observation:
\end{minipage}
}
\textbf{Back Trace Prompt}\\
\centering{
\begin{minipage}{0.98\textwidth}
You are an agent following an action plan to navigation in \textcolor{lightgray}{indoor} \textcolor{promptred}{outdoor} environment.
\end{minipage}
}
\end{tcolorbox}

\begin{tcolorbox}[colback=promptwhite, colframe=lightgray, boxrule=0mm, arc=1mm,colbacktitle=promptpurple, fonttitle=\Large, coltitle=white, title=\textbf{NavGPT Prompt}, toptitle = 1mm, bottomtitle = 0.5mm]
\centering
\centering{
\begin{minipage}{0.98\textwidth}
You are currently at an intermediate step of the trajectory but seems going off the track. You will be given the action plan describing the whole trajectory, the history of previous steps you have taken, the observations of the \textcolor{lightgray}{viewpoints along the trajectory} \textcolor{promptred}{current position}.\\
You should evaluate the history, the action plan and the observations along the way to decide the \textcolor{lightgray}{viewpoints} \textcolor{promptred}{next action} to go back to.\\
\textcolor{lightgray}{Each navigable viewpoint has a unique ID, you should only answer the ID in the Final Answer.}
\textcolor{promptred}{Action list: [`TASK$\_$FINISH', `MOVE$\_$FORW ARD (5 meters)', `TURN$\_$LEFT (15 degrees)', `TURN\_RIGHT (15 degrees)', `ASCENT (2 meters)', `DESCENT (2 meters)', `MOVE\_LEFT (5 meters)', `MOVE\_RIGHT (5 meters)']}\\
You must choose one from the \textcolor{lightgray}{navigable viewpoints} \textcolor{promptred}{action list}, DO NOT answer None of the above.\\
----\\
Starting below, you should follow this format:\\
Action plan: the action plan describing the whole trajectory\\
History: the history of previous steps you have taken\\
Observation: the observations of \textcolor{lightgray}{each viewpoint along the trajectory} \textcolor{promptred}{the current position}

Thought: your thought about the next step\\
Final Answer: `\textcolor{lightgray}{viewpointID} \textcolor{promptred}{action}'\\
----\\
Begin!\\
Action plan: \{action\_plan\}\\
History: \{history\}\\
Observation: \{observation\}\\
Thought:
\end{minipage}
}
\end{tcolorbox}

\subsection{DiscussNav}
DiscussNav\cite{discussnav} employs BLIP-2 and RAM \cite{Zhang2023RecognizeAA} for visual understanding, and GPT-4 for reasoning across four expert modules. Like NavGPT, DiscussNav was designed for indoor, discrete VLN tasks. We apply the same modifications: replacing LLM/VLM modules and adjusting the input/output formats to suit UAV navigation.\par

\begin{tcolorbox}[colback=promptwhite, colframe=lightgray, boxrule=0mm, arc=1mm,colbacktitle=promptpurple, fonttitle=\Large, coltitle=white, title=\textbf{DiscussNav Prompt}, toptitle = 1mm, bottomtitle = 0.5mm]
\centering
\textbf{Instruction Analysis Experts}\\
\centering{
\begin{minipage}{0.98\textwidth}
\textbf{System Prompt: }\\
You are \textcolor{lightgray}{an action} \textcolor{promptred}{a sub-instructions} decomposition expert. Your task is to detect all \textcolor{lightgray}{actions} \textcolor{promptred}{sub-instructions} in the given navigation instruction. You need to ensure the integrity of each \textcolor{lightgray}{action} \textcolor{promptred}{sub-instruction}.\\
\textbf{User Prompt: }\\
Can you decompose \textcolor{lightgray}{actions} \textcolor{promptred}{sub-instructions} in the instruction `\{instruction\}'? \textcolor{lightgray}{Actions} \textcolor{promptred}{Sub-instructions}: \\
\textbf{System Prompt: }\\
You are a landmark extraction expert. Your task is to detect all landmarks in the given navigation instruction. You need to ensure the integrity of each landmark.\\
\textbf{User Prompt: }\\
Can you extract landmarks in the instruction `\{\textcolor{lightgray}{actions} \textcolor{promptred}{sub-instructions}\}'? Landmarks: 
\end{minipage}
}
\textbf{Vision Perception Experts}\\
\centering{
\begin{minipage}{0.98\textwidth}
Describe this \textcolor{lightgray}{indoor} scene in details
\end{minipage}
}
\textbf{Completion Estimation Experts}\\
\centering{
\begin{minipage}{0.98\textwidth}
\textbf{System Prompt: }\\
You are a trajectory summary expert. Your task is to simplify navigation thought process as short and clear as possible.\\
\textbf{User Prompt: }\\
Given Thought Process `\{thought\}', Summarization:\\
\textbf{System Prompt: }\\
You are a completion estimation expert. Your task is to estimate what \textcolor{lightgray}{actions} \textcolor{promptred}{sub-instructions} in the instruction have been executed based on navigation history and landmarks. \\
All \textcolor{lightgray}{actions} \textcolor{promptred}{sub-instructions} in the instruction are given following the temporal order. Your answer includes two parts: `Thought' and `Executed \textcolor{lightgray}{Actions} \textcolor{promptred}{Sub-instructions}'. \\
In the `Thought', you must follow procedures to analyze as detailed as possible what \textcolor{lightgray}{actions} \textcolor{promptred}{sub-instructions} have been executed: \\
(1) What given landmarks of \textcolor{lightgray}{actions} \textcolor{promptred}{sub-instructions} have appeared in the navigation history? \\
(2) Analyze the direction change at each step in the navigation history. \\
(3) Estimate each \textcolor{lightgray}{action} \textcolor{promptred}{sub-instruction} in the instruction based on each step in the navigation history to check their completion. \\
In the `Executed \textcolor{lightgray}{Actions} \textcolor{promptred}{Sub-instructions}', you must only write down \textcolor{lightgray}{actions} \textcolor{promptred}{sub-instructions} that have been executed without other words. You must 
\end{minipage}
}
\end{tcolorbox}
\begin{tcolorbox}[colback=promptwhite, colframe=lightgray, boxrule=0mm, arc=1mm,colbacktitle=promptpurple, fonttitle=\Large, coltitle=white, title=\textbf{DiscussNav Prompt}, toptitle = 1mm, bottomtitle = 0.5mm]
\centering{
\begin{minipage}{0.98\textwidth}
strictly refer original \textcolor{lightgray}{actions} \textcolor{promptred}{sub-instructions} in the given instruction to estimate.\\
\textbf{User Prompt: }\\
Given Navigation History `\{history$\_$traj\}' and Landmarks `\{landmarks\}', estimate what \textcolor{lightgray}{actions} \textcolor{promptred}{sub-instructions} in instruction `{\textcolor{lightgray}{actions} \textcolor{promptred}{sub-instructions}}' have been executed.\\  
\end{minipage}
}
\textbf{Decision Testing Experts}\\
\centering{
\begin{minipage}{0.98\textwidth}
\textbf{System Prompt: }\\
You are a thought fusion expert. Your task is to fuse given thought processes into one thought. You need to reserve key information related to actions, landmarks, direction changes. You should only answer fused thought without other words.\\
\textbf{User Prompt: }\\
Can you help me fuse the thoughts leading to the same movement \textcolor{lightgray}{direction} \textcolor{promptred}{action}? The thoughts are :\{multiple$\_$thoughts\}, Fused thought: \\
\textbf{System Prompt: }\\
You are a decision testing expert. Your task is to evaluate the feasibility of each movement prediction based on thought process and environment. Then, you will make a final decision about \textcolor{lightgray}{navigation viewpoint ID} \textcolor{promptred}{movement action} without other words.\\
\textbf{User Prompt: }\\
Can you help me make a final decision \textcolor{promptred}{about the next action}? \textcolor{promptred}{Please choose only one executable action from the options available.} The Observation: \{observation\}, Navigation Instruction: \{instruction\}, \{fused$\_$pred$\_$thought$\_$\}, Final Decision: \\
\end{minipage}
}
\centering
\textbf{Planner Prompt(DiscussNav agent)}\\
\centering{
\begin{minipage}{0.98\textwidth}
You are a navigation agent who follows instruction to move in an \textcolor{lightgray}{indoor} \textcolor{promptred}{outdoor} environment with the least action steps. \textcolor{promptred}{you can only choose one action from the given list:\\
\{description\}}\\
I will give you one instruction and tell you landmarks. I will also give you navigation history and estimation of executed actions for reference. \\
You can observe current environment by scene descriptions, scene objects and possible existing landmarks in different directions around you. \\ 
\textcolor{lightgray}{Each direction contains navigable viewpoints you can move to. Your task is to predict moving to which navigable viewpoint.} \\
Note that environment direction that contains more landmarks mentioned in the instruction is usually the better choice for you. 
\end{minipage}
}
\end{tcolorbox}

\begin{tcolorbox}[colback=promptwhite, colframe=lightgray, boxrule=0mm, arc=1mm,colbacktitle=promptpurple, fonttitle=\Large, coltitle=white, title=\textbf{DiscussNav Prompt}, toptitle = 1mm, bottomtitle = 0.5mm]
\centering{
\begin{minipage}{0.98\textwidth}
\textcolor{promptred}{The Collision Warning is the tips to avoid collisions and infer your relative position with surroundings. }\\
\textcolor{lightgray}{If you are required to go up stairs, you need to move to direction with higher position. If you are required to go down stairs, you need to move to direction with lower position.\\
You are encouraged to move to new viewpoints to explore environment while avoid revisiting accessed viewpoints in non-essential situations. } \\
If you feel struggling to find the landmark or execute the \textcolor{lightgray}{action} \textcolor{promptred}{sub-instruction}, you can try to execute the subsequent \textcolor{lightgray}{action} \textcolor{promptred}{sub-instruction} and find the subsequent landmark. \\
Your answer includes two parts: `Thought' and `Prediction'. In the `Thought', you should think as detailed as possible following procedures: \\
(1) Check whether the latest executed action has been completed by comparing current environment and landmark in the latest executed action. \\
(2) Determine the action you should execute and landmark you should reach now. If the latest executed action have not been completed, you should continue to execute it. Otherwise, you should execute the next action in the given instruction. \\
(3) Analyze which direction in the current environment is most suitable to execute the action you decide and explain your reason. \\
(4) \textcolor{lightgray}{Predict moving to which navigable viewpoint based on your thought process.} \textcolor{promptred}{Predict promptred}{which action to execute based on your thought process.}\\
Then, please make decision on the next \textcolor{lightgray}{viewpoint} \textcolor{promptred}{action} in the `Prediction'. You must only answer next \textcolor{lightgray}{viewpoint ID} \textcolor{promptred}{action} in the `Prediction' without other words. \\
Your decision is very important, must make it very carefully."
\end{minipage}
}
\end{tcolorbox}

\section{Detailed Prompts for FineCog-Nav}
We follow the order of module descriptions in Figure 2 of the main text to present our detailed prompt designs. For clarity, each module is highlighted using the corresponding background color from the pipeline diagram.

\subsection{Instruction Parser}
Instruction Parser decomposes the long, complex instruction into a sequence of concise, actionable instruction sentences. This segmentation ensures that the agent can focus on localized goals rather than handling the entire instruction at once.

\begin{tcolorbox}[colback = instructback, colframe = instruct, fonttitle = \bfseries, colbacktitle = instruct, coltitle=black, title=Instruction Parser]

\centering{
\begin{minipage}{0.98\textwidth}
$\left[ \right.$General TASK$\left. \right]$: You are an Instruction Manager for UAV Navigation. You have the following two tasks:\\

1. SENTENCE SEGMENTATION\\
- Split input text into individual sentences using periods as separators\\
- Preserve original wording including leading conjunctions (e.g., ``and...")\\
- Maintain original capitalization and spacing\\

2. LANDMARK EXTRACTION\\
- Identify ALL navigational landmarks (physical objects/locations)\\
- Capture full noun phrases following prepositions: to/at/near/above/before\\
- Pay special attention to landmarks after temporal prepositions (e.g., ``after the building")\\
- Retain modifiers: ``small building", ``shop entrance", etc.\\
\\
$\left[ \right.$NOTE$\left. \right]$:\\
- Verify period placement for accurate segmentation\\
- Strictly detect temporal clauses and adjust execution order\\
- Include ALL landmarks per sentence (1-3 typical)\\
- Ensure landmarks are bound to corresponding actions in complex sentences\\
- NEVER modify original wording in sub-instructions\\
- Eliminate all JSON syntax errors\\

$\left[ \right.$OUTPUT CONSTRAINTS$\left. \right]$:\\
- Strictly formatted as a JSON code block without any explanatory text\\
- Always use arrays even for single items\\

$\left[ \right.$EXPECTED OUTPUT$\left. \right]$:\\
```json\\
$\left[ \right.$ $\{\{$\\
``sub-instruction": ``...",\\
``landmark": $\left[ \right.$``LANDMARK1",...$\left. \right]$
  $\}\}$,\\
  $\{\{$\\
 ``sub-instruction": ``...",\\
    ``landmark": $\left[ \right.$``LANDMARK1",...$\left. \right]$\\
  $\}\}$ $\left. \right]$\\
```\\
\#\#\# Input \#\#\#\\
$\langle$Instructions$\rangle$:\\
$\{$navigation\_instruction$\}$
\end{minipage}
}
\end{tcolorbox}

\subsection{Subgoal Extractor}
During each navigation step, Subgoal Extractor identifies the immediate subgoal based on the current instruction sentence and the perceived scene context. This dynamic subgoal extraction bridges high-level instructions and low-level UAV actions, enabling precise, step-wise planning.

\begin{tcolorbox}[colback = subgoalback, colframe = subgoal, fonttitle = \bfseries, colbacktitle = subgoal, coltitle=black, title=Subgoal Extractor]
\centering{
\begin{minipage}{0.98\textwidth}
[General TASK]: You are a UAV Navigation Strategist. Perform means-end analysis to decompose the $\langle$current instruction$\rangle$ into sequential subgoals.\\
\\
{[Input]}:\\
$\langle$Current Instruction$\rangle$\\
$\langle$Current Observation$\rangle$\\
\\
{[Detailed Task]}: SUBGOAL EXTRACTION \\
  - Identify all embedded subgoals/intermediate objectives within the $\langle$Current Instruction$\rangle$\\
  - Extract complete phrases denoting actions/steps (e.g., ``turn left at intersection")\\
  - Sequence subgoals by PHYSICAL EXECUTION ORDER, prioritizing temporal logic over sentence structure (Example: ``Do A when B" $\rightarrow$ {[``B", ``A"]})\\
  - Replace pronouns with explicit referents (e.g., ``... A and ... it" $\rightarrow$ {[``... A", ``... A"]})\\
  - Preserve descriptive modifiers/contextual details.\\
  - Cross-reference $\langle$Current Observation$\rangle$ to ensure executable subgoal generation\\

{[WARNING]}:\\
  - Carefully follow $\langle$Current Instruction$\rangle$, DO NOT involve extra descriptive words before $\langle$landmarks$\rangle$, DO NOT add extra information not mentioned.\\

{[OUTPUT CONSTRAINTS]}:\\
- Strictly formatted as a JSON code block without any explanatory text\\
- Always use arrays even for single items\\
\\
{[EXPECTED OUTPUT]}: \\
```json\\
{[``SUBGOAL1", ``SUBGOAL2"]}\\
```\\
\#\#\# Input \#\#\#\\
$\langle$Current Instruction$\rangle$:
\{instruction\}\\
\\
$\langle$Current Observation$\rangle$:
\{observation\}
\end{minipage}
}
\end{tcolorbox}

\subsection{Attention}
The Attention Module is tightly coupled with the Perception module and is responsible for generating targeted questions for the perception module based on the current and upcoming instruction sentences.

\begin{tcolorbox}[colback = attenback, colframe = atten, fonttitle = \bfseries, colbacktitle = atten, coltitle=black, title=Attention]
\centering{
\begin{minipage}{0.98\textwidth}
$\left[ \right.$TASK$\left. \right]$: You are an attention guider for a drone's vision module. Generate questions considering distinct landmarks in $\left[ \right.$Instructions$\left. \right]$ to direct visual attention of Perception Module. \\
\\
$\left[ \right.$INPUT$\left. \right]$: \\
$\langle$Current Instruction$\rangle$\\
$\langle$Next Instruction$\rangle$\\
\\
$\left[ \right.$OUTPUT CONSTRAINTS$\left. \right]$: \\
1. Strictly formatted as a JSON code block without any explanatory text.\\
2. The OUTPUT JSON must include the following information:\\
A list (array) where each element is an object with the following contents:\\
  - ``landmark": The name of the landmark.\\
  - ``question": A question regarding the landmark's CURRENT status(e.g. position or distance, etc.), guiding the attention of the perception module.\\
\\
$\left[ \right.$NOTE$\left. \right]$:\\
1. Question examples:\\
  - ``Is the $\langle$landmark$\rangle$ in front of you / on your left / on your right / above / below?"\\
  - ``Is the $\langle$landmark$\rangle$ far / near?"\\
  - ``What are the distinct attributes(color, shape, size, etc.) of the $\langle$landmark$\rangle$?"\\
2. The questions should be specific to the CURRENT status of the landmark and should not reference any actions or future states.\\
\\
$\left[ \right.$EXPECTED OUTPUT$\left. \right]$:\\
```json\\
$\left[ \right.$ \{\{\\
  ``landmark": ``landmark 1",\\
  ``question": ``...?"\\
  \}\},\\
  ...$\left. \right]$\\
```\\
\#\#\# Input \#\#\#\\
$\langle$Current Instruction$\rangle$:\{current\_instruction\}\\
\\
$\langle$Next Instruction$\rangle$:\{next\_instruction\}
\end{minipage}
}
\end{tcolorbox}

\subsection{Perception}
The Perception Module generates scene descriptions based on the targeted questions provided by the attention module, serving as the foundational component of scene understanding.

\begin{tcolorbox}[colback = percptback, colframe = percpt, fonttitle = \bfseries, colbacktitle = percpt, coltitle=black, title=Perception]
\centering{
\begin{minipage}{0.98\textwidth}
$\left[ \right.$TASK$\left. \right]$: You are an advanced multimodal perception system for a drone that navigates in the real world. Your task is to analyze first-person view(the drone's camera is facing forward) image and generate environmental semantics.\\
\\
$\left[ \right.$INPUT$\left. \right]$:\\
$\langle$Attention$\rangle$: The landmarks you should focus on\\
\\
$\left[ \right.$NOTE$\left. \right]$:\\
1. ONLY include CURRENT VISIBLE landmarks in the $\langle$Attention$\rangle$.\\
\\
$\left[ \right.$OUTPUT CONSTRAINTS$\left. \right]$:\\
1. Strictly formatted as a JSON code block without any explanatory text.\\
2. The $\left[ \right.$OUTPUT$\left. \right]$ JSON must include the following information:\\
- ``overall": A string describe the scene according to the image input.\\
- ``details": A string describe the VISIBLE landmarks in the $\langle$Attention$\rangle$, estimate their the location and distance(in meters). In Front: describe the middle part of the image, On my Left: describe the left part of the image,On my Right: describe the right part of the image, etc. \\
\\
$\left[ \right.$EXPECTED OUTPUT$\left. \right]$:\\
```json\\
\{\{\\
  ``overall": ``Overall: I see a scene of ...", \\
  ``details": ``In Front: ... On my Left: ... On my Right: ... etc."\\
\}\}\\
```\\
\\
\#\#\# Input \#\#\#\\
$\langle$Attention$\rangle$:
\{suggestion\}
\end{minipage}
}
\end{tcolorbox}

\subsection{Imagination}
Based on the current and next subgoals, along with the landmark list, the Imagination Module imagines the anticipated state after completing the current subgoal.
\begin{tcolorbox}[colback = imaginback, colframe = imagin, fonttitle = \bfseries, colbacktitle = imagin, coltitle=black, title=Imagination]
\centering{
\begin{minipage}{0.98\textwidth}
$\left[ \right.$TASK$\left. \right]$: You are a drone performing navigation task, trying to achieve $\langle$subgoal$\rangle$. Now please imagine the final VISUAL state when you have achieved the given $\langle$subgoal$\rangle$. First-person View (The drone's camera is facing forward).\\
\\
$\left[ \right.$INPUT$\left. \right]$: \\
$\langle$Current Subgoal$\rangle$\\
$\langle$Landmarks$\rangle$\\
$\langle$Next Subgoal$\rangle$\\
\\
$\left[ \right.$NOTE$\left. \right]$:\\
1. Your imagination should be a very concise sentence ONLY describing the visual state of $\langle$Landmarks$\rangle$ after you achieved the $\langle$Current Subgoal$\rangle$, including:\\
  - The location and distance(in meters) of $\langle$Landmarks$\rangle$. Prefer SPATIAL terms: LEFT/RIGHT/CENTER, etc.\\
2. Consider the $\langle$Next Subgoal$\rangle$ to infer the final state of $\langle$Current Subgoal$\rangle$.\\
3. You can only see a small window: a little to the left, right, up, and down (about 45° each way).\\
   - If something ends up underneath you, behind you, or too high/low (outside that window), do NOT mention it.\\
   - Example (general): After moving above a landmark, it is now below you and out of view.\\
\\
$\left[ \right.$OUTPUT CONSTRAINTS$\left. \right]$:\\
Strictly formatted as a JSON code block without any explanatory text.\\
\\
$\left[ \right.$EXPECTED OUTPUT$\left. \right]$:\\
```json\\
\{\{
  ``state": ``When I have finished the $\langle$subgoal$\rangle$, I might see ...(the location and distance of the $\langle$Landmarks$\rangle$)."
\}\}\\
```\\
\\
\#\#\# Input \#\#\#\\
$\langle$Current Subgoal$\rangle$:
\{subgoal\}\\
\\
$\langle$Landmarks$\rangle$:
\{landmark\}\\
\\
$\langle$Next Subgoal$\rangle$:
\{next\_subgoal\}
\end{minipage}
}
\end{tcolorbox}

\subsection{Subgoal Judger}
The Subgoal Judger determines whether the current subgoal has been successfully completed.

\begin{tcolorbox}[colback = subjudgerback, colframe = subjudger, fonttitle = \bfseries, colbacktitle = subjudger, coltitle=black, title=Subgoal Judger]
\centering{
\begin{minipage}{0.98\textwidth}
$\left[ \right.$TASK$\left. \right]$: You are a drone navigation analysis expert. Your task is to estimate whether the subgoal has been achieved or not. \\
\\
$\left[ \right.$INPUT$\left. \right]$:\\
$\langle$Current Subgoal$\rangle$\\
$\langle$Subgoal Memory$\rangle$: What you have ALREADY DONE and OBSERVED for $\langle$Current Subgoal$\rangle$.\\
$\langle$Current Observation$\rangle$\\
$\langle$Reference Imagination$\rangle$: The blind imagined reference VISUAL state when you have achieved the $\langle$Subgoal$\rangle$.\\
$\langle$Next Subgoal$\rangle$: The next subgoal you are trying to transition to\\
\\
$\left[ \right.$NOTE$\left. \right]$:\\
1. To make your decision, analyze the inputs as follows:\\
- Check the $\langle$Current Observation$\rangle$, mainly focus on the status of Landmarks, and then decide if it is time to switch from the $\langle$Current Subgoal$\rangle$ to $\langle$Next Subgoal$\rangle$.\\
- Review the $\langle$Subgoal Memory$\rangle$ to check whether the actions taken align with the $\langle$Current Subgoal$\rangle$. \\
- $\langle$Reference Imagination$\rangle$ is a blind guess, take it for reference occasionally. \\
2. If the subgoal says **turn left/right** without a specified degree or reference, it means a large turn, usually $>$ 45 degrees( $>$ 3 times).\\
3. If the $\langle$Current Observation$\rangle$ is not matched with the $\langle$Reference Imagination$\rangle$, but still make sense as both initial state of the $\langle$Next Subgoal$\rangle$ and final state of the $\langle$Current Subgoal$\rangle$, then it might also achieved.\\
4. If there are multiple($>$15) steps in the $\langle$Subgoal Memory$\rangle$, you should check whether you have achieved the $\langle$Current Subgoal$\rangle$ before the last step. \\
\\
$\left[ \right.$ADDITIONAL REQUIREMENTS$\left. \right]$:\\
1. $\left[ \right.$Action Compliance Verification$\left. \right]$ For each subgoal evaluation, you MUST verify: \\
  - Every action SPECIFIED in the subgoal description has been executed (cross-validate with $\langle$Subgoal Memory$\rangle$)\\
\\
$\left[ \right.$OUTPUT CONSTRAINTS$\left. \right]$:\\
1. Strictly formatted as a JSON code block without any explanatory text.\\
2. Please also provide a brief explanation involving your evidence from $\left[ \right.$INPUT$\left. \right]$.
\end{minipage}
}
\end{tcolorbox}
\begin{tcolorbox}[colback = subjudgerback, colframe = subjudger, fonttitle = \bfseries, colbacktitle = subjudger, coltitle=black, title=Subgoal Judger]
\centering{
\begin{minipage}{0.98\textwidth}
$\left[ \right.$EXPECTED OUTPUT$\left. \right]$:\\
```json\\
\{\{\\
  ``subgoal": ``$\langle$subgoal$\rangle$",\\
  ``achieved": $\langle$true$|$false$\rangle$,\\
  ``reason": ``...(brief explanation)"\\
\}\}\\
```\\
\#\#\# Input \#\#\#\\
$\langle$Current Subgoal$\rangle$:
\{subgoal\}\\
\\
$\langle$Subgoal Memory$\rangle$:
\{history\}\\
\\
$\langle$Current Observation$\rangle$:
\{scene\}\\
\\
$\langle$Reference Imagination$\rangle$:
\{state\}\\
\\
$\langle$Next Subgoal$\rangle$:
\{next\_subgoal\}
\end{minipage}
}
\end{tcolorbox}

\subsection{Multi-Level Memory}
The Multi-Level Memory module implements a hierarchical memory structure that integrates both short-term and long-term memory. In detail, this module consists of Step Memory, Subgoal Memory, and Instruction Memory. The corresponding prompts for each component are described.

\begin{tcolorbox}[colback = memoryback, colframe = memory, fonttitle = \bfseries, colbacktitle = memory, coltitle=black, title=Multi-Level Memory(Step Memory)]
\centering{
\begin{minipage}{0.98\textwidth}
$\left[ \right.$TASK$\left. \right]$: You are a **Memory Manager** for a drone that navigates in the real world. Generate a memory statement using the provided $\left[ \right.$INPUT$\left. \right]$ data.\\
\\
$\left[ \right.$INPUT$\left. \right]$:\\
$\langle$Current Observation$\rangle$: The scene you are *SEEING*\\
$\langle$Current Action$\rangle$: The action you are *DOING*\\
$\langle$Landmarks$\rangle$\\
\\
$\left[ \right.$NOTE$\left. \right]$:\\
1. Keep accurate DEGREE/DISTANCE in $\langle$Action$\rangle$\\
2. ONLY includes VISIBLE $\langle$Landmarks$\rangle$ in the $\langle$Current Observation$\rangle$\\
3. Do NOT infer or include any planning steps or actions not explicitly mentioned in the input.\\
\\
$\left[ \right.$OUTPUT CONSTRAINTS$\left. \right]$:\\
1. Strictly formatted as a JSON code block without any explanatory text
\end{minipage}
}
\end{tcolorbox}
\begin{tcolorbox}[colback = memoryback, colframe = memory, fonttitle = \bfseries, colbacktitle = memory, coltitle=black, title=Multi-Level Memory(Step Memory)]
\centering{
\begin{minipage}{0.98\textwidth}
2. **Word limit**: $<$ 30 words\\
\\
$\left[ \right.$EXPECTED OUTPUT$\left. \right]$:\\
```json\\
\{\{  ``step\_memory": the memory in the form of ``I see ...(observation); I ...(action)". \}\}\\
'''\\
\#\#\# Input \#\#\#\\
$\langle$Current Observation$\rangle$:
\{observation\}\\
\\
$\langle$Current Action$\rangle$:
\{action\}\\
\\
$\langle$Landmarks$\rangle$:
\{landmarks\}
\end{minipage}
}
\end{tcolorbox}

For each subgoal, we introduce a Subgoal Memory that operates during its execution phase.

\begin{tcolorbox}[colback = memoryback, colframe = memory, fonttitle = \bfseries, colbacktitle = memory, coltitle=black, title=Multi-Level Memory(Subgoal Memory)]
\centering{
\begin{minipage}{0.98\textwidth}
$\left[ \right.$TASK$\left. \right]$: You are a **Memory Manager** for a drone that navigates in the real world. **Consolidate** $\langle$Raw Subgoal Memory$\rangle$ containing a list of actions and observations in temporal order.\\
\\
$\left[ \right.$INPUT$\left. \right]$:\\
$\langle$Raw Subgoal Memory$\rangle$\\
$\langle$Landmarks$\rangle$:\\
\\
$\left[ \right.$NOTE$\left. \right]$:\\
1. **Merge the Information:**\\
  - Merge the same actions(note that turn left and turn right are different actions) and accumulate the numerical values.\\
  - Only merge *consecutive* actions of the same type (for example, consecutive ``move forward" steps).\\
    - Do not merge actions if they are interrupted by a different action type.\\
    - Preserve the temporal order of actions and observations.\\
  - Only include the landmarks that are VISIBLE in the $\langle$Raw Subgoal Memory$\rangle$, and pay special attention to changes in landmarks..\\
2. **Output Requirements:**\\
  - The output memory should be concise, clear, and logically organized. \\
\\
$\left[ \right.$OUTPUT CONSTRAINTS$\left. \right]$:\\
1. Strictly formatted as a JSON code block without any explanatory text
\end{minipage}
}
\end{tcolorbox}

\begin{tcolorbox}[colback = memoryback, colframe = memory, fonttitle = \bfseries, colbacktitle = memory, coltitle=black, title=Multi-Level Memory(Subgoal Memory)]
\centering{
\begin{minipage}{0.98\textwidth}
2. **Word limit**: $<$ 40 words\\
\\
$\left[ \right.$EXPECTED OUTPUT$\left. \right]$:\\
```json\\
\{\{  ``subgoal\_memory": ``I have seen ...; I ...(action)". \}\}\\
```\\
\#\#\# Input \#\#\#\\
$\langle$Raw Subgoal Memory$\rangle$:
\{raw\_memory\}\\
\\
$\langle$Landmarks$\rangle$:
\{landmarks\}
\end{minipage}
}
\end{tcolorbox}

The Instruction Memory is simply a collection of completed subgoal records, without any associated prompts. 

\subsection{Decision Making}
The Decision Making module serves as the final stage of the entire framework. Its primary function is to determine the UAV’s next action, such as move forward, turn left/right, ascend, descend, or stop.

\begin{tcolorbox}[colback = decisionback, colframe = decision, fonttitle = \bfseries, colbacktitle = decision, coltitle=black, title=Decision-Making]
\centering{
\begin{minipage}{0.98\textwidth}
$\left[ \right.$TASK$\left. \right]$: You are an embodied drone that navigates in the real world. Your task is to CHOOSE the most reasonable action to be executed next from the VALID ACTIONS, so that you could achieve the subgoal.\\
\\
$\left[ \right.$INPUT$\left. \right]$: \\
$\langle$Instruction Memory$\rangle$: What you have ALREADY DONE and OBSERVED for the current instruction.\\
$\langle$Current Observation$\rangle$\\
$\langle$Collision Warning$\rangle$\\
$\langle$Current Instruction$\rangle$\\
$\langle$Current Subgoal$\rangle$: The subgoal of the current instruction you are working on.\\
$\langle$Valid Actions$\rangle$\\
\\
$\left[ \right.$NOTE$\left. \right]$:\\
1. To make your decision, analyze the inputs as follows: \\
  - Understand the $\langle$Current Subgoal$\rangle$, find out the actions aligned with it.\\
  - Consider the $\langle$Collision Warning$\rangle$ to avoid collisions.\\
2. **Probability Rules**:\\
  - Output probabilities for **ALL Valid Actions**
\end{minipage}
}
\end{tcolorbox}
\begin{tcolorbox}[colback = decisionback, colframe = decision, fonttitle = \bfseries, colbacktitle = decision, coltitle=black, title=Decision-Making]
\centering{
\begin{minipage}{0.98\textwidth}
  - Higher probability = stronger preference\\
3. Carefully check the $\langle$Instruction Memory$\rangle$: If over multiple steps you consistently observe scenes that no visible landmarks - immediately select the action 0\\
\\
$\left[ \right.$OUTPUT CONSTRAINTS$\left. \right]$:\\
1. Strictly formatted as a JSON code block without any explanatory text.\\
2. The $\left[ \right.$OUTPUT$\left. \right]$ JSON must include the following information:\\
  - ``thought": why you choose this action, considering the $\langle$Current Subgoal$\rangle$, $\langle$Current Observation$\rangle$, $\langle$Instruction Memory$\rangle$, $\langle$Current Instruction$\rangle$ and $\langle$Collision Warning$\rangle$.\\
  - ``probabilities": probability distribution over the valid action list.\\
 - ``selected\_action": Explicitly select the action with highest probability. Note that only output the action number.\\
\#\#\# INPUT\\
$\langle$Instruction Memory$\rangle$:
\{memory\}\\
\\
$\langle$Current Observation$\rangle$:
\{observation\}\\
\\
$\langle$Collision Warning$\rangle$:
\{collisions\_warning\}\\
\\
$\langle$Current Instruction$\rangle$:
\{current\_instruction\}\\
\\
$\langle$Current Subgoal$\rangle$:
\{subgoal\}\\
\\
$\langle$Valid Actions$\rangle$:
\{actions\}
\end{minipage}
}
\end{tcolorbox}

\section{Execution Details}
Since the instructions in our AerialVLN-Fine dataset are refined, to more comprehensively and realistically demonstrate the performance of our method, we select trajectory 3570Y55XZYHE60QQZMK98C5MBKWYG9 from the original AerialVLN-S validation set for detailed analysis. In this example, we apply our FineCog-Nav framework, utilizing Qwen2.5-72B-Instruct \cite{qwen2.5} as the LLM and Qwen2.5-VL-32B \cite{qwen2.5} as the VLM. Specifically, our method traverses 75 meters and terminates at step 31. We present the complete information for several representative steps, including the inputs and outputs of each module at every stage of execution.

To simplify the description, we omit the repeated content from \textbf{$\left[ \right.$TASK$\left. \right]$} to \textbf{\#\#\#Input\#\#\#} as detailed in the previous section, and only present the specific inputs and outputs here.

\subsection{Step 0}

\begin{tcolorbox}[colback = instructback, colframe = instruct, fonttitle = \bfseries, colbacktitle = instruct, coltitle=black, title=Instruction Parser]
\centering
\textbf{Input Information}
\centering{
\begin{minipage}{0.98\textwidth}
$\langle$Instructions$\rangle$:
turn left and fly over the yellow truck. turn left and take the next left at another truck. land in front of the back tire of the truck. stop on the sidewalk.\\
\end{minipage}
}
\centering
\textbf{Output Information}
\centering{
\begin{minipage}{0.98\textwidth}
$\left[ \right.$
  \{
    ``sub-instruction": ``turn left and fly over the yellow truck.",\\
    ``landmark": $\left[ \right.$``the yellow truck"$\left. \right]$
  \},\\
  \{
    ``sub-instruction": ``turn left and take the next left at another truck.",\\
    ``landmark": $\left[ \right.$``another truck"$\left. \right]$
  \}, \\
  \{
    ``sub-instruction": ``land in front of the back tire of the truck.",\\
    ``landmark": $\left[ \right.$``the back tire of the truck"$\left. \right]$
  \},\\
  \{
    ``sub-instruction": ``stop on the sidewalk.",\\
    ``landmark": $\left[ \right.$``the sidewalk"$\left. \right]$
  \}
$\left. \right]$
\end{minipage}
}
\end{tcolorbox}

\begin{tcolorbox}[colback = attenback, colframe = atten, fonttitle = \bfseries, colbacktitle = atten, coltitle=black, title=Attention]
\centering
\textbf{Input Information}
\centering{
\begin{minipage}{0.98\textwidth}
$\langle$Current Instruction$\rangle$:\\
turn left and fly over the yellow truck.\\
Landmark: the yellow truck\\
$\langle$Next Instruction$\rangle$:\\
turn left and take the next left at another truck.\\
Landmark: another truck\\
\end{minipage}
}
\centering
\textbf{Output Information}
\centering{
\begin{minipage}{0.98\textwidth}
$\left[ \right.$
  \{`landmark': `the yellow truck', `question': `Is the yellow truck in front of you or on your left?'\}, \\
  \{`landmark': `another truck', `question': `Can you see another truck in your current field of view?'\}
$\left. \right]$
\end{minipage}
}
\end{tcolorbox}

\begin{tcolorbox}[colback = percptback, colframe = percpt, fonttitle = \bfseries, colbacktitle = percpt, coltitle=black, title=Perception]
\centering
\textbf{Input Information}
\centering{
\begin{minipage}{0.9\textwidth}
\begin{center}
\includegraphics[width=0.9\linewidth]{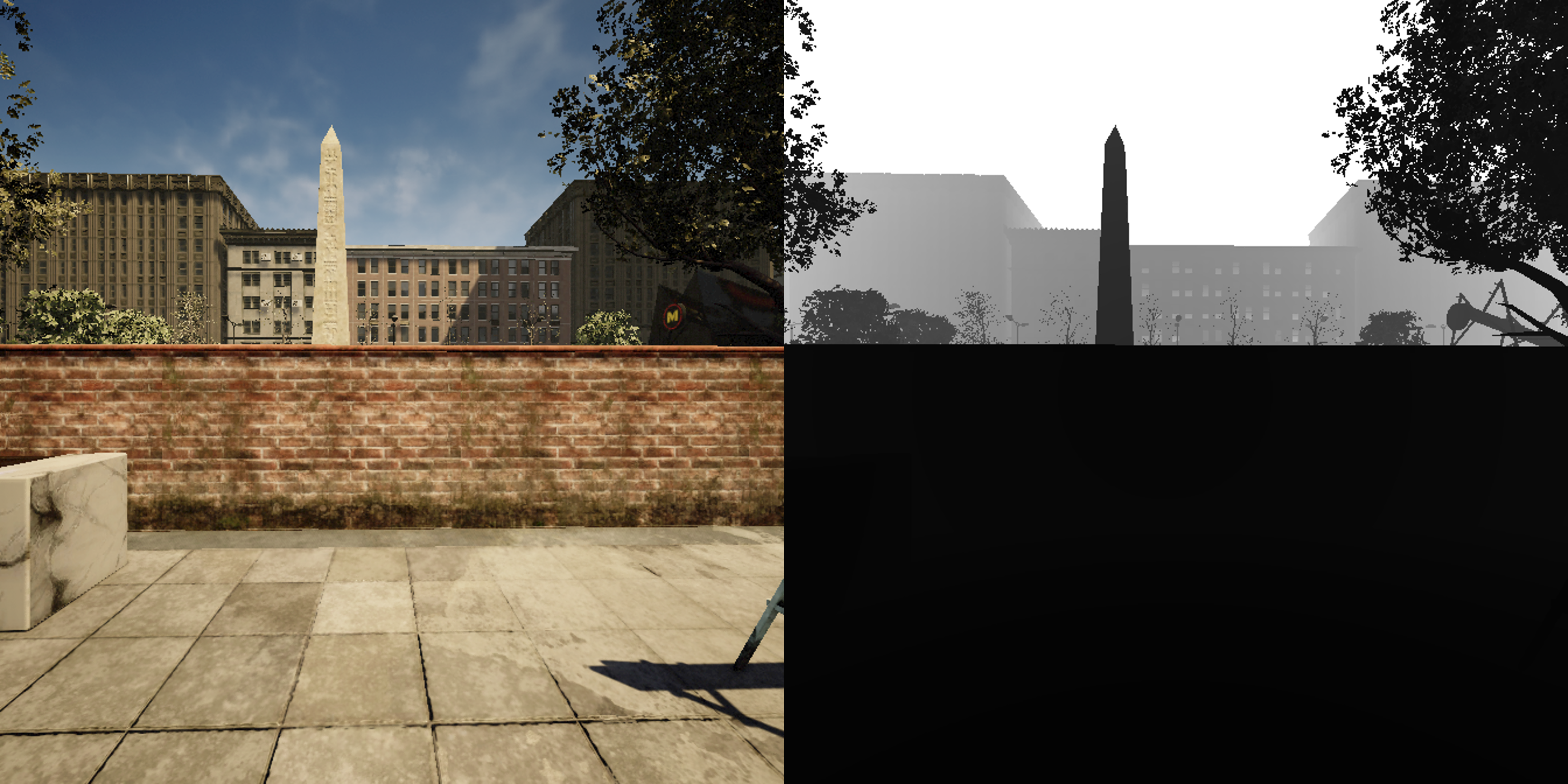}
\label{fig:2_2}
\end{center}
\end{minipage}
}
\centering{
\begin{minipage}{0.98\textwidth}
$\langle$Attention$\rangle$:\\
Is the yellow truck in front of you or on your left?\\
Can you see another truck in your current field of view?
\end{minipage}
}
\end{tcolorbox}
\begin{tcolorbox}[colback = percptback, colframe = percpt, fonttitle = \bfseries, colbacktitle = percpt, coltitle=black, title=Perception]
\centering
\textbf{Output Information}
\centering{
\begin{minipage}{0.98\textwidth}
Overall: I see a scene of an urban environment with a clear sky, a tall obelisk in the background, and a brick wall in the foreground. The area appears to be a plaza or open space with paved ground. \\
In Front: A paved plaza with a brick wall running horizontally across the middle of the image. The wall is about 5 meters away. On my Left: A white marble structure, possibly a monument or sculpture, is visible near the left edge of the image, approximately 3 meters away. On my Right: A dark-colored truck is partially visible behind the brick wall, about 10 meters away. Another truck is not visible in the current field of view.
\end{minipage}
}
\end{tcolorbox}

\begin{tcolorbox}[colback = subgoalback, colframe = subgoal, fonttitle = \bfseries, colbacktitle = subgoal, coltitle=black, title=Subgoal Extractor]
\centering
\textbf{Input Information}
\centering{
\begin{minipage}{0.98\textwidth}
$\langle$Current Instruction$\rangle$:\\
turn left and fly over the yellow truck.\\
$\langle$Current Observation$\rangle$: \\
Overall: I see a scene of an urban environment with a clear sky, a tall obelisk in the background, and a brick wall in the foreground. The area appears to be a plaza or open space with paved ground. \\
In Front: A paved plaza with a brick wall running horizontally across the middle of the image. The wall is about 5 meters away. On my Left: A white marble structure, possibly a monument or sculpture, is visible near the left edge of the image, approximately 3 meters away. On my Right: A dark-colored truck is partially visible behind the brick wall, about 10 meters away. Another truck is not visible in the current field of view.\\
\end{minipage}
}
\centering
\textbf{Output Information}
\centering{
\begin{minipage}{0.98\textwidth}
$\left[ \right.$
  ``turn left", ``fly over the yellow truck"
$\left. \right]$
\end{minipage}
}
\end{tcolorbox}

\begin{tcolorbox}[colback = imaginback, colframe = imagin, fonttitle = \bfseries, colbacktitle = imagin, coltitle=black, title=Imagination]
\centering
\textbf{Input Information}
\centering{
\begin{minipage}{0.98\textwidth}
$\langle$Current Subgoal$\rangle$:\\
turn left\\
$\langle$Landmarks$\rangle$:\\
the yellow truck\\
$\langle$Next Subgoal$\rangle$:\\
fly over the yellow truck
\end{minipage}
}
\end{tcolorbox}

\begin{tcolorbox}[colback = imaginback, colframe = imagin, fonttitle = \bfseries, colbacktitle = imagin, coltitle=black, title=Imagination]
\centering
\textbf{Output Information}
\centering{
\begin{minipage}{0.98\textwidth}
When I have finished the turn left, I might see the yellow truck at the CENTER, about 10 meters ahead.
\end{minipage}
}
\end{tcolorbox}
\begin{tcolorbox}[colback = subjudgerback, colframe = subjudger, fonttitle = \bfseries, colbacktitle = subjudger, coltitle=black, title=Subgoal Judger]
\centering
\textbf{Input Information}
\centering{
\begin{minipage}{0.98\textwidth}
$\langle$Current Subgoal$\rangle$:\\
turn left\\
$\langle$Subgoal Memory$\rangle$:\\
None\\
$\langle$Current Observation$\rangle$:\\
Overall: I see a scene of an urban environment with a clear sky, a tall obelisk in the background, and a brick wall in the foreground. The area appears to be a plaza or open space with paved ground.\\
In Front: A paved plaza with a brick wall running horizontally across the middle of the image. The wall is about 5 meters away. On my Left: A white marble structure, possibly a monument or sculpture, is visible near the left edge of the image, approximately 3 meters away. On my Right: A dark-colored truck is partially visible behind the brick wall, about 10 meters away. Another truck is not visible in the current field of view.\\
$\langle$Reference Imagination$\rangle$:\\
When I have finished the turn left, I might see the yellow truck at the CENTER, about 10 meters ahead.\\
$\langle$Next Subgoal$\rangle$:\\
fly over the yellow truck
\end{minipage}
}
\centering
\textbf{Output Information}
\centering{
\begin{minipage}{0.98\textwidth}
subgoal: turn left \\
achieved: false\\
reason: The $\langle$Current Observation$\rangle$ does not match the $\langle$Reference Imagination$\rangle$. The yellow truck is not visible in the center of the field of view, and the $\langle$Subgoal Memory$\rangle$ indicates no actions have been taken to turn left. The current view is still aligned with the initial state before the turn.
\end{minipage}
}
\end{tcolorbox}

\begin{tcolorbox}[colback = decisionback, colframe = decision, fonttitle = \bfseries, colbacktitle = decision, coltitle=black, title=Decision-Making]
\centering
\textbf{Input Information}
\centering{
\begin{minipage}{0.98\textwidth}
$\langle$Instruction Memory$\rangle$:\\
For $\langle$turn left$\rangle$:\\
None\\
$\langle$Current Observation$\rangle$:\\
Overall: I see a scene of an urban environment 
\end{minipage}
}
\end{tcolorbox}

\begin{tcolorbox}[colback = decisionback, colframe = decision, fonttitle = \bfseries, colbacktitle = decision, coltitle=black, title=Decision-Making]
\centering{
\begin{minipage}{0.98\textwidth}
with a clear sky, a tall obelisk in the background, and a brick wall in the foreground. The area appears to be a plaza or open space with paved ground.\\
In Front: A paved plaza with a brick wall running horizontally across the middle of the image. The wall is about 5 meters away. On my Left: A white marble structure, possibly a monument or sculpture, is visible near the left edge of the image, approximately 3 meters away. On my Right: A dark-colored truck is partially visible behind the brick wall, about 10 meters away. Another truck is not visible in the current field of view. \\
$\langle$Collision Warning$\rangle$:\\
**MOVE\_FORWARD will collide with objects. DESCEND will collide with objects. ** \\
$\langle$Current Instruction$\rangle$:\\
turn left and fly over the yellow truck.\\
$\langle$Current Subgoal$\rangle$:\\
turn left\\
$\langle$Valid Actions$\rangle$:\\
1: MOVE\_FORWARD (5 meters)\\
2: TURN\_LEFT (15 degrees)\\
3: TURN\_RIGHT (15 degrees)\\
4: ASCEND (2 meters)\\
5: DESCEND (2 meters)\\
6: MOVE\_LEFT (5 meters)\\
7: MOVE\_RIGHT (5 meters)\\
\end{minipage}
}
\centering
\textbf{Output Information}
\centering{
\begin{minipage}{0.98\textwidth}
thought: The current subgoal is to turn left. The current observation shows a brick wall in front and a white marble structure on the left, which is approximately 3 meters away. The collision warning indicates that moving forward or descending will result in a collision. Therefore, the most reasonable action is to turn left to align with the subgoal and avoid any obstacles.\\
probabilities: \{``1": 0.05, ``2": 0.85, ``3": 0.05, ``4": 0.02, ``5": 0.01, ``6": 0.01, ``7": 0.01\}\\
selected\_action": 2
\end{minipage}
}
\end{tcolorbox}

\begin{tcolorbox}[colback = memoryback, colframe = memory, fonttitle = \bfseries, colbacktitle = memory, coltitle=black, title=Multi-Level Memory(Step Memory)]
\centering
\textbf{Input Information}
\centering{
\begin{minipage}{0.98\textwidth}
$\langle$Current Observation$\rangle$:\\
Overall: I see a scene of an urban environment with a clear sky, a tall obelisk in the background, and a brick wall in the foreground. The area appears to be a plaza or open space with paved ground.
\end{minipage}
}
\end{tcolorbox}

\begin{tcolorbox}[colback = memoryback, colframe = memory, fonttitle = \bfseries, colbacktitle = memory, coltitle=black, title=Multi-Level Memory(Step Memory)]
\centering{
\begin{minipage}{0.98\textwidth}
In Front: A paved plaza with a brick wall running horizontally across the middle of the image. The wall is about 5 meters away. On my Left: A white marble structure, possibly a monument or sculpture, is visible near the left edge of the image, approximately 3 meters away. On my Right: A dark-colored truck is partially visible behind the brick wall, about 10 meters away. Another truck is not visible in the current field of view. \\
$\langle$Current Action$\rangle$:\\
TURN\_LEFT (15 degrees)\\
$\langle$Landmarks$\rangle$:\\
the yellow truck $|$ another truck\\
\end{minipage}
}
\centering
\textbf{Output Information}
\centering{
\begin{minipage}{0.98\textwidth}
I see a brick wall 5 meters away and a white marble structure 3 meters to my left; I turn left 15 degrees.
\end{minipage}
}
\end{tcolorbox}

\begin{tcolorbox}[colback = memoryback, colframe = memory, fonttitle = \bfseries, colbacktitle = memory, coltitle=black, title=Multi-Level Memory(Subgoal Memory)]
\centering{
\begin{minipage}{0.98\textwidth}
At step-1: I see a brick wall 5 meters away and a white marble structure 3 meters to my left; I turn left 15 degrees.
\end{minipage}
}
\end{tcolorbox}

\subsection{Step 13}
To simplify the process, we directly move to Step 13, where a subgoal is completed, to illustrate how the prompt focus shifts before and after the subgoal transition.

Firstly, the Subgoal Judger determines that the current task is complete. Note that the Attention questions and Imagination states have been cached.

\begin{tcolorbox}[colback = percptback, colframe = percpt, fonttitle = \bfseries, colbacktitle = percpt, coltitle=black, title=Perception]
\centering
\textbf{Input Information}
\centering{
\begin{minipage}{0.9\textwidth}
\begin{center}
\includegraphics[width=0.9\linewidth]{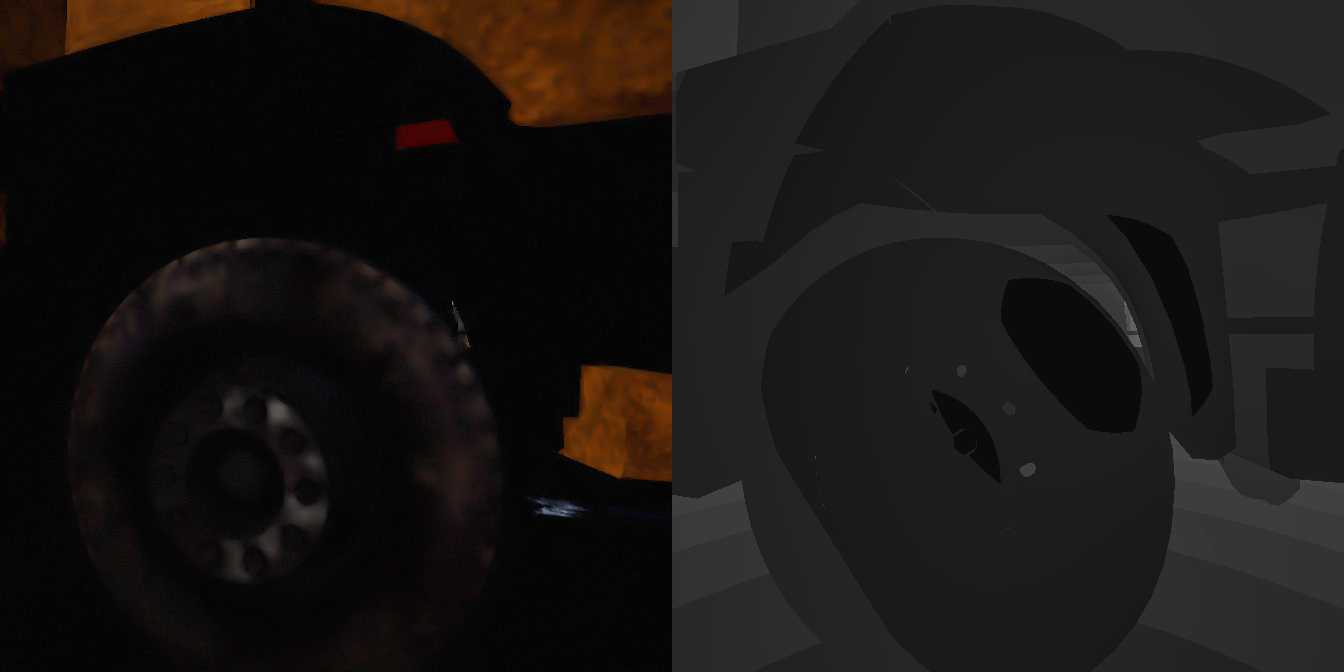}
\label{fig:step13}
\end{center}
\end{minipage}
}
\centering{
\begin{minipage}{0.98\textwidth}
$\langle$Attention$\rangle$:\\
Is the yellow truck in front of you or on your left?\\
Can you see another truck in your current field of view?
\end{minipage}
}\end{tcolorbox}

\begin{tcolorbox}[colback = percptback, colframe = percpt, fonttitle = \bfseries, colbacktitle = percpt, coltitle=black, title=Perception]
\centering
\textbf{Output Information}
\centering{
\begin{minipage}{0.98\textwidth}
Overall: I see a scene with a close-up view of a vehicle's wheel and part of its body, with a dark and somewhat obscured environment. \\
In Front: The wheel of a vehicle is prominently visible, occupying the central part of the image. On my Left: The left side of the image shows part of the vehicle's body, which appears to be dark and metallic. On my Right: The right side of the image shows a portion of the vehicle's body and a small red object, possibly a light or reflector. The yellow truck is not visible in the current field of view. Another truck is also not visible in the current field of view.
\end{minipage}
}
\end{tcolorbox}

\begin{tcolorbox}[colback = subjudgerback, colframe = subjudger, fonttitle = \bfseries, colbacktitle = subjudger, coltitle=black, title=Subgoal Judger]
\centering
\textbf{Input Information}
\centering{
\begin{minipage}{0.98\textwidth}
$\langle$Current Subgoal$\rangle$:\\
fly over the yellow truck\\
$\langle$Subgoal Memory$\rangle$:\\
At step-1: I see a paved area with a bench and a marble-like structure in front, a green dumpster on my left, and a tall obelisk-like structure in the background; I turn left 15 degrees.\\
At step-2: I see a paved plaza with a bench, a black trash bin, and a tree; I turn left 15 degrees.\\
At step-3: I see a wide sidewalk, a blue trash bin, a streetlight, a brick wall, a bench, and a marble structure; I turn left 15 degrees. \\
At step-4: I see a red truck 10 meters left, a brick wall, bench, and trash bin 5 meters right; I turn left 15 degrees. \\
At step-5: I see a yellow truck 20 meters left, a red truck 15 meters right, and trash bins, bench 5-10 meters right; I move left 5 meters.\\
At step-6: I see a yellow truck 50 meters ahead, tall buildings on both sides, and a red car 30 meters to the right; I move forward 5 meters.\\
At step-7: I see a red car 10m right, a yellow truck 30m left, and a blue trash bin 15m right; I turn left 15 degrees.\\
At step-8: I see a yellow truck 20 meters ahead and a red car 10 meters to the right; I move forward 5 meters.\\
At step-9: I see a yellow truck 20 meters ahead, tall buildings 30 meters on both sides; I move forward 5 meters.\\
At step-10: I see a yellow truck 50 meters ahead, a tall building with a `PARKING' sign 30 meters to 
\end{minipage}
}
\end{tcolorbox}
\begin{tcolorbox}[colback = subjudgerback, colframe = subjudger, fonttitle = \bfseries, colbacktitle = subjudger, coltitle=black, title=Subgoal Judger]
\centering{
\begin{minipage}{0.98\textwidth}
the left, and another tall building 30 meters to the right; I move forward 5 meters. \\
At step-11: I see a yellow truck 10 meters ahead and tall buildings on both sides; I move forward 5 meters.\\
$\langle$Current Observation$\rangle$:\\
Overall: I see a scene with a close-up view of a vehicle's wheel and part of its body, with a dark and somewhat obscured environment. \\
In Front: The wheel of a vehicle is prominently visible, occupying the central part of the image. On my Left: The left side of the image shows part of the vehicle's body, which appears to be dark and metallic. On my Right: The right side of the image shows a portion of the vehicle's body and a small red object, possibly a light or reflector. The yellow truck is not visible in the current field of view. Another truck is also not visible in the current field of view.\\
$\langle$Reference Imagination$\rangle$:\\
When I have finished the $\langle$subgoal$\rangle$, I might see the yellow truck 10 meters below me and out of view.\\
$\langle$Next Subgoal$\rangle$:\\
turn left and take the next left at another truck. 
\end{minipage}
}
\centering
\textbf{Output Information}
\centering{
\begin{minipage}{0.98\textwidth}
subgoal: fly over the yellow truck,\\
achieved: true,\\
reason: The $\langle$Current Observation$\rangle$ shows a close-up view of a vehicle's wheel and part of its body, which is consistent with being directly above the yellow truck. The $\langle$Subgoal Memory$\rangle$ indicates that the drone has moved forward and is now very close to the yellow truck, which aligns with the $\langle$Reference Imagination$\rangle$ of the yellow truck being 10 meters below and out of view. This suggests that the drone has successfully flown over the yellow truck.
\end{minipage}
}
\end{tcolorbox}

After completing the subgoal, the Subgoal Memory is compressed, and a new empty subgoal memory is initialized. The compressed memory is referred to as raw memory.

\begin{tcolorbox}[colback = memoryback, colframe = memory, fonttitle = \bfseries, colbacktitle = memory, coltitle=black, title=Multi-Level Memory(Subgoal Memory)]
\centering
\textbf{Input Information}
\centering{
\begin{minipage}{0.98\textwidth}
$\langle$Raw Subgoal Memory$\rangle$:\\
At step-1: I see a paved area with a bench and a marble-like structure in front, a green dumpster on my left, and a tall obelisk-like structure in the background; I turn left 15 degrees.
\end{minipage}
}
\end{tcolorbox}

\begin{tcolorbox}[colback = memoryback, colframe = memory, fonttitle = \bfseries, colbacktitle = memory, coltitle=black, title=Multi-Level Memory(Subgoal Memory)]
\centering{
\begin{minipage}{0.98\textwidth}
At step-2: I see a paved plaza with a bench, a black trash bin, and a tree; I turn left 15 degrees.\\
At step-3: I see a wide sidewalk, a blue trash bin, a streetlight, a brick wall, a bench, and a marble structure; I turn left 15 degrees.\\
At step-4: I see a red truck 10 meters left, a brick wall, bench, and trash bin 5 meters right; I turn left 15 degrees.\\
At step-5: I see a yellow truck 20 meters left, a red truck 15 meters right, and trash bins, bench 5-10 meters right; I move left 5 meters.\\
At step-6: I see a yellow truck 50 meters ahead, tall buildings on both sides, and a red car 30 meters to the right; I move forward 5 meters.\\
At step-7: I see a red car 10m right, a yellow truck 30m left, and a blue trash bin 15m right; I turn left 15 degrees.\\
At step-8: I see a yellow truck 20 meters ahead and a red car 10 meters to the right; I move forward 5 meters.\\
At step-9: I see a yellow truck 20 meters ahead, tall buildings 30 meters on both sides; I move forward 5 meters.\\
At step-10: I see a yellow truck 50 meters ahead, a tall building with a `PARKING' sign 30 meters to the left, and another tall building 30 meters to the right; I move forward 5 meters.\\
At step-11: I see a yellow truck 10 meters ahead and tall buildings on both sides; I move forward 5 meters.\\
$\langle$Landmarks$\rangle$:\\
the yellow truck $|$ another truck\\
\end{minipage}
}
\centering
\textbf{Output Information}
\centering{
\begin{minipage}{0.98\textwidth}
I see a paved area, green dumpster, obelisk; I turn left 60 degrees; I see a red truck, yellow truck, trash bins, bench; I move left 5 meters, forward 20 meters; I see a yellow truck, tall buildings, `PARKING' sign; I move forward 5 meters.
\end{minipage}
}
\end{tcolorbox}

Since this is the final subgoal of the instruction, the Attention questions, Scene description, Subgoal list, and Imagination states must be regenerated.

\begin{tcolorbox}[colback = attenback, colframe = atten, fonttitle = \bfseries, colbacktitle = atten, coltitle=black, title=Attention]
\centering
\textbf{Input Information}
\centering{
\begin{minipage}{0.98\textwidth}
$\langle$Current Instruction$\rangle$:\\
turn left and take the next left at another truck.\\
Landmark: another truck\\
$\langle$Next Instruction$\rangle$:
\end{minipage}
}
\end{tcolorbox}

\begin{tcolorbox}[colback = attenback, colframe = atten, fonttitle = \bfseries, colbacktitle = atten, coltitle=black, title=Attention]
\centering{
\begin{minipage}{0.98\textwidth}
land in front of the back tire of the truck.\\
Landmark: the back tire of the truck \\
\end{minipage}
}
\centering
\textbf{Output Information}
\centering{
\begin{minipage}{0.98\textwidth}
$\left[ \right.$ \{`landmark': `another truck', `question': `Is the truck on your left?'\}, \\
  \{`landmark': `the back tire of the truck', `question': `Is the back tire of the truck in front of you?'\} $\left. \right]$
\end{minipage}
}
\end{tcolorbox}
\begin{tcolorbox}[colback = subgoalback, colframe = subgoal, fonttitle = \bfseries, colbacktitle = subgoal, coltitle=black, title=Subgoal Extractor]
\centering
\textbf{Input Information}
\centering{
\begin{minipage}{0.98\textwidth}
$\langle$Current Instruction$\rangle$:\\
turn left and fly over the yellow truck.\\
$\langle$Current Observation$\rangle$:\\
Overall: I see a scene with a close-up view of a vehicle's wheel and part of its body, with a dark and somewhat obscured environment. \\
In Front: The wheel of a vehicle is prominently visible, occupying the central part of the image. On my Left: The left side of the image shows part of the vehicle's body, which appears to be dark and metallic. On my Right: The right side of the image shows a portion of the vehicle's body and a small red object, possibly a light or reflector. The yellow truck is not visible in the current field of view. Another truck is also not visible in the current field of view.
\end{minipage}
}
\centering
\textbf{Output Information}
\centering{
\begin{minipage}{0.98\textwidth}
$\left[ \right.$
``turn left", ``take the next left at another truck"
$\left. \right]$
\end{minipage}
}
\end{tcolorbox}
\begin{tcolorbox}[colback = percptback, colframe = percpt, fonttitle = \bfseries, colbacktitle = percpt, coltitle=black, title=Perception]
\centering
\textbf{Input Information}
\centering{
\begin{minipage}{0.9\textwidth}
\begin{center}
\includegraphics[width=0.9\linewidth]{sec/figs_supp/step13.png}
\label{fig:2_1}
\end{center}
\end{minipage}
}
\centering{
\begin{minipage}{0.98\textwidth}
$\langle$Attention$\rangle$:\\
Is the truck on your left?\\
Is the back tire of the truck in front of you?\\
\end{minipage}
}
\centering
\textbf{Output Information}
\centering{
\begin{minipage}{0.98\textwidth}
Overall: I see a scene with a close-up view of a vehicle's tire, likely a truck, in a dimly lit environment.
\end{minipage}
}
\end{tcolorbox}
\begin{tcolorbox}[colback = percptback, colframe = percpt, fonttitle = \bfseries, colbacktitle = percpt, coltitle=black, title=Perception]
\centering{
\begin{minipage}{0.98\textwidth}
In Front: A large tire is prominently visible, occupying the central part of the image. The tire appears to be part of a truck, and its back tire is directly in front of me. Distance: Approximately 1 meter. On my Left: The left side of the image shows part of the truck's body, which is dark and blends into the background. Distance: Approximately 1-2 meters. On my Right: The right side of the image shows a portion of the truck's body and a small red light or reflector. Distance: Approximately 1-2 meters.
\end{minipage}
}
\end{tcolorbox}

\begin{tcolorbox}[colback = decisionback, colframe = decision, fonttitle = \bfseries, colbacktitle = decision, coltitle=black, title=Decisioin-Making]
\centering
\textbf{Input Information}
\centering{
\begin{minipage}{0.98\textwidth}
$\langle$Instruction Memory$\rangle$:\\
None\\
$\langle$Current Observation$\rangle$:\\
Overall: I see a scene with a close-up view of a vehicle's tire, likely a truck, in a dimly lit environment. \\
In Front: A large tire is prominently visible, occupying the central part of the image. The tire appears to be part of a truck, and its back tire is directly in front of me. Distance: Approximately 1 meter. On my Left: The left side of the image shows part of the truck's body, which is dark and blends into the background. Distance: Approximately 1-2 meters. On my Right: The right side of the image shows a portion of the truck's body and a small red light or reflector. Distance: Approximately 1-2 meters.\\
$\langle$Collision Warning$\rangle$:\\
**MOVE\_FORWARD will collide with objects. ASCEND will collide with objects. DESCEND will collide with objects. **
$\langle$Current Instruction$\rangle$:\\
turn left and take the next left at another truck.\\
$\langle$Current Subgoal$\rangle$:\\
turn left\\
$\langle$Valid Actions$\rangle$:\\
1: MOVE\_FORWARD (5 meters)\\
2: TURN\_LEFT (15 degrees)\\
3: TURN\_RIGHT (15 degrees)\\
4: ASCEND (2 meters)\\
5: DESCEND (2 meters)\\
6: MOVE\_LEFT (5 meters)\\
7: MOVE\_RIGHT (5 meters)
\end{minipage}
}
\end{tcolorbox}
\begin{tcolorbox}[colback = decisionback, colframe = decision, fonttitle = \bfseries, colbacktitle = decision, coltitle=black, title=Decisioin-Making]
\centering
\textbf{Output Information}
\centering{
\begin{minipage}{0.98\textwidth}
thought: Given the current subgoal to turn left, and considering the collision warnings for moving forward, ascending, and descending, the most reasonable action is to turn left. The current observation shows a large tire and truck body on the left, which aligns with the subgoal of turning left.\\
probabilities: \{``1": 0.05, ``2": 0.85, ``3": 0.02, ``4": 0.01, ``5": 0.01, ``6": 0.03, ``7": 0.01\}\\
selected\_action: 2
\end{minipage}
}
\end{tcolorbox}

\begin{tcolorbox}[colback = memoryback, colframe = memory, fonttitle = \bfseries, colbacktitle = memory, coltitle=black, title=Multi-Level Memory(Step Memory)]
\centering
\textbf{Input Information}
\centering{
\begin{minipage}{0.98\textwidth}
$\langle$Current Observation$\rangle$:\\
Overall: I see a scene with a close-up view of a vehicle's tire, likely a truck, in a dimly lit environment. \\
In Front: A large tire is prominently visible, occupying the central part of the image. The tire appears to be part of a truck, and its back tire is directly in front of me. Distance: Approximately 1 meter. On my Left: The left side of the image shows part of the truck's body, which is dark and blends into the background. Distance: Approximately 1-2 meters. On my Right: The right side of the image shows a portion of the truck's body and a small red light or reflector. Distance: Approximately 1-2 meters. \\
$\langle$Current Action$\rangle$:\\
TURN\_LEFT (15 degrees)\\
$\langle$Landmarks$\rangle$:\\
another truck | the back tire of the truck \\
\end{minipage}
}
\centering
\textbf{Output Information}
\centering{
\begin{minipage}{0.98\textwidth}
I see a large tire and part of a truck's body; I turn left 15 degrees.
\end{minipage}
}
\end{tcolorbox}

\begin{tcolorbox}[colback = memoryback, colframe = memory, fonttitle = \bfseries, colbacktitle = memory, coltitle=black, title=Multi-Level Memory(Subgoal Memory)]
\centering{
\begin{minipage}{0.98\textwidth}
At step-1: I see a large tire and part of a truck's body; I turn left 15 degrees.
\end{minipage}
}
\end{tcolorbox}

\subsection{Step 14}
Our agent finishes a subgoal at step 13. Therefore, at step 14, it starts with a new subgoal and changes in the imagination module.

\begin{tcolorbox}[colback = percptback, colframe = percpt, fonttitle = \bfseries, colbacktitle = percpt, coltitle=black, title=Perception]
\centering
\textbf{Input Information}
\centering{
\begin{minipage}{0.9\textwidth}
\begin{center}
\includegraphics[width=0.9\linewidth]{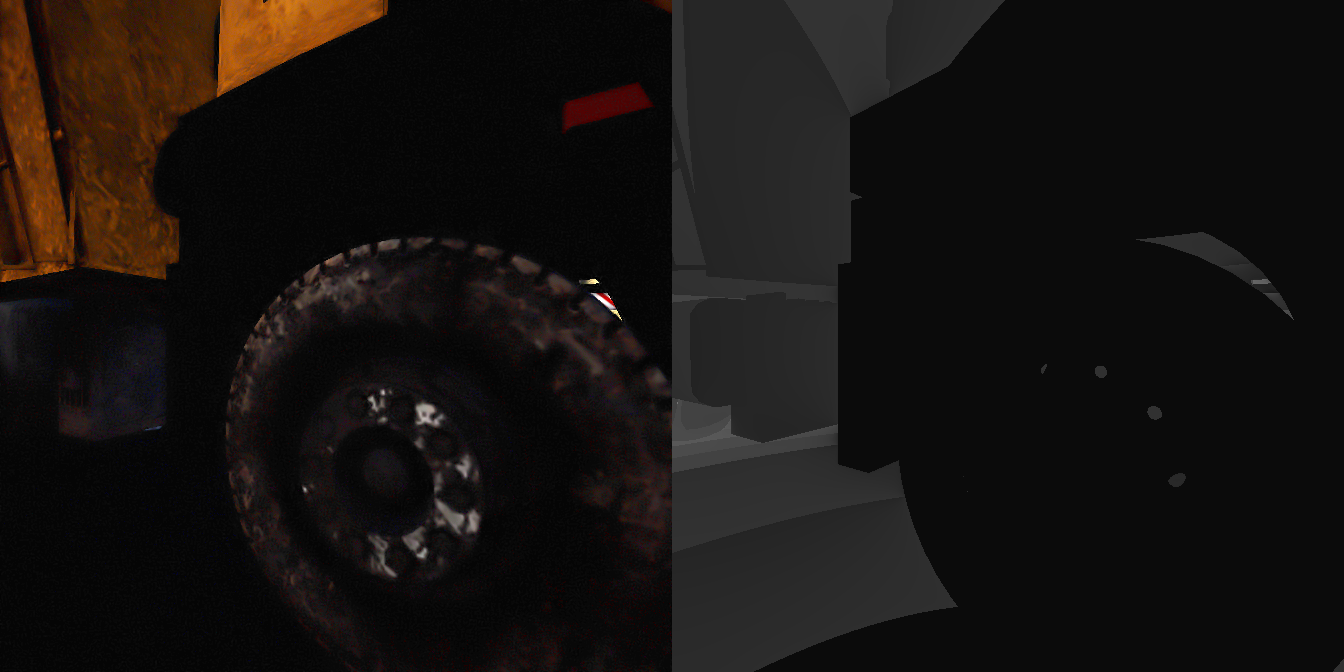}
\label{fig:step14}
\end{center}
\end{minipage}
}
\centering{
\begin{minipage}{0.98\textwidth}
$\langle$Attention$\rangle$:\\
Is the truck on your left?\\
Is the back tire of the truck in front of you?\\
\end{minipage}
}
\centering
\textbf{Output Information}
\centering{
\begin{minipage}{0.98\textwidth}
Overall: I see a close-up view of a truck's rear section, including its tire and part of its body.\\
In Front: The back tire of the truck is directly in front of me, approximately 1 meter away. On my Left: The side of the truck is visible, extending to the left, with the truck's body occupying most of the left side of the image. On my Right: The truck's rear section, including part of its tailgate and body, is visible, occupying the right side of the image.
\end{minipage}
}
\end{tcolorbox}

The Imagination states are regenerated when needed. 

\begin{tcolorbox}[colback = imaginback, colframe = imagin, fonttitle = \bfseries, colbacktitle = imagin, coltitle=black, title=Imagination]
\centering
\textbf{Input Information}
\centering{
\begin{minipage}{0.98\textwidth}
$\langle$Current Subgoal$\rangle$:\\
turn left\\
$\langle$Landmarks$\rangle$:\\
another truck\\
$\langle$Next Subgoal$\rangle$:\\
take the next left at another truck\\
\end{minipage}
}
\centering
\textbf{Output Information}
\centering{
\begin{minipage}{0.98\textwidth}
When I have finished the turn left, I might see another truck 20 meters to the LEFT.
\end{minipage}
}
\end{tcolorbox}

\begin{tcolorbox}[colback = subjudgerback, colframe = subjudger, fonttitle = \bfseries, colbacktitle = subjudger, coltitle=black, title=Subgoal Judger]
\centering
\textbf{Input Information}
\centering{
\begin{minipage}{0.98\textwidth}
$\langle$Current Subgoal$\rangle$:\\
turn left\\
$\langle$Subgoal Memory$\rangle$:\\
At step-1: I see a large tire and part of a truck's body; I turn left 15 degrees.\\
$\langle$Current Observation$\rangle$:\\
Overall: I see a close-up view of a truck's rear section, including its tire and part of its body.
\end{minipage}
}
\end{tcolorbox}
\begin{tcolorbox}[colback = subjudgerback, colframe = subjudger, fonttitle = \bfseries, colbacktitle = subjudger, coltitle=black, title=Subgoal Judger]
\centering{
\begin{minipage}{0.98\textwidth}
In Front: The back tire of the truck is directly in front of me, approximately 1 meter away. On my Left: The side of the truck is visible, extending to the left, with the truck's body occupying most of the left side of the image. On my Right: The truck's rear section, including part of its tailgate and body, is visible, occupying the right side of the image.\\
$\langle$Reference Imagination$\rangle$:\\
When I have finished the turn left, I might see another truck 20 meters to the LEFT.\\
$\langle$Next Subgoal$\rangle$:\\
take the next left at another truck \\
\end{minipage}
}
\centering
\textbf{Output Information}
\centering{
\begin{minipage}{0.98\textwidth}
subgoal: turn left\\
achieved: False\\
reason: The $\langle$Current Observation$\rangle$ indicates that the drone is still facing the rear section of the truck, with the truck's body occupying both the left and right sides of the image. The $\langle$Subgoal Memory$\rangle$ shows only a 15-degree turn, which is not sufficient to achieve a large turn ($>$ 45 degrees). The $\langle$Reference Imagination$\rangle$ suggests that another truck should be visible 20 meters to the left, which is not observed in the current state.
\end{minipage}
}
\end{tcolorbox}

\begin{tcolorbox}[colback = decisionback, colframe = decision, fonttitle = \bfseries, colbacktitle = decision, coltitle=black, title=Decision-Making]
\centering
\textbf{Input Information}
\centering{
\begin{minipage}{0.98\textwidth}
$\langle$Instruction Memory$\rangle$:\\
For $\langle$turn left$\rangle$:\\
At step-1: I see a large tire and part of a truck's body; I turn left 15 degrees.\\
$\langle$Current Observation$\rangle$:\\
Overall: I see a close-up view of a truck's rear section, including its tire and part of its body.\\
In Front: The back tire of the truck is directly in front of me, approximately 1 meter away. On my Left: The side of the truck is visible, extending to the left, with the truck's body occupying most of the left side of the image. On my Right: The truck's rear section, including part of its tailgate and body, is visible, occupying the right side of the image.\\
$\langle$Collision Warning$\rangle$:\\
**MOVE\_FORWARD will collide with objects. ASCEND will collide with objects. DESCEND will collide with objects. **\\
$\langle$Current Instruction$\rangle$:\\
turn left and take the next left at another truck.
\end{minipage}
}
\end{tcolorbox}

\begin{tcolorbox}[colback = decisionback, colframe = decision, fonttitle = \bfseries, colbacktitle = decision, coltitle=black, title=Decision-Making]
\centering{
\begin{minipage}{0.98\textwidth}
$\langle$Current Subgoal$\rangle$:\\
turn left\\
$\langle$Valid Actions$\rangle$:\\
1: MOVE\_FORWARD (5 meters)\\
2: TURN\_LEFT (15 degrees)\\
3: TURN\_RIGHT (15 degrees)\\
4: ASCEND (2 meters)\\
5: DESCEND (2 meters)\\
6: MOVE\_LEFT (5 meters)\\
7: MOVE\_RIGHT (5 meters)\\
\end{minipage}
}
\centering
\textbf{Output Information}
\centering{
\begin{minipage}{0.98\textwidth}
thought: Given the current subgoal to turn left and the observation that the truck's rear section is directly in front and to the sides, the most reasonable action is to turn left. Moving forward, ascending, or descending will result in a collision. Moving left or right will not align with the subgoal of turning left.\\
probabilities: \{``1": 0.05, ``2": 0.85, ``3": 0.02, ``4": 0.01, ``5": 0.01, ``6": 0.02, ``7": 0.04\}\\
selected\_action: 2 
\end{minipage}
}
\end{tcolorbox}

\begin{tcolorbox}[colback = memoryback, colframe = memory, fonttitle = \bfseries, colbacktitle = memory, coltitle=black, title=Multi-Level Memory(Step Memory)]
\centering
\textbf{Input Information}
\centering{
\begin{minipage}{0.98\textwidth}
$\langle$Current Observation$\rangle$:\\
Overall: I see a close-up view of a truck's rear section, including its tire and part of its body.
In Front: The back tire of the truck is directly in front of me, approximately 1 meter away. On my Left: The side of the truck is visible, extending to the left, with the truck's body occupying most of the left side of the image. On my Right: The truck's rear section, including part of its tailgate and body, is visible, occupying the right side of the image.\\
$\langle$Current Action$\rangle$:\\
TURN\_LEFT (15 degrees)\\
$\langle$Landmarks$\rangle$:\\
another truck $|$ the back tire of the truck\\
\end{minipage}
}
\centering
\textbf{Output Information}
\centering{
\begin{minipage}{0.98\textwidth}
I see the back tire of the truck 1 meter in front; I turn left 15 degrees.
\end{minipage}
}
\end{tcolorbox}

\begin{tcolorbox}[colback = memoryback, colframe = memory, fonttitle = \bfseries, colbacktitle = memory, coltitle=black, title=Multi-Level Memory(Subgoal Memory)]
\centering{
\begin{minipage}{0.98\textwidth}
At step-1: I see a large tire and part of a truck's body; I turn left 15 degrees.\\
At step-2: I see the back tire of the truck 1 meter in front; I turn left 15 degrees.
\end{minipage}
}
\end{tcolorbox}

\end{document}